\documentclass[11pt, letterpaper, logo, copyright]{googledeepmind}

\usepackage[utf8]{inputenc} 
\usepackage[T1]{fontenc}    
\usepackage{xcolor}

\usepackage{hyperref}       
\usepackage{url}            
\usepackage{booktabs}       
\usepackage{amsfonts}       
\usepackage{nicefrac}       
\usepackage{microtype}      
\usepackage{xcolor}         
\usepackage{booktabs}
\usepackage{pgfplotstable}
\usepackage{amsmath} 
\usepackage{etoolbox} 
\usepackage{listings}
\usepackage{xcolor}

\usepackage{siunitx}       
\usepackage{multirow}      
\usepackage[authoryear, sort&compress, round]{natbib}
\usepackage{booktabs}
\usepackage[font=small]{caption}
\usepackage{makecell}      

\lstdefinelanguage{json}{
    basicstyle=\ttfamily\small,
    numbers=left,
    numberstyle=\tiny\color{gray},
    stepnumber=1,
    numbersep=5pt,
    showstringspaces=false,
    breaklines=true,
    frame=single,
    backgroundcolor=\color{gray!5},
    literate=
     *{0}{{\textcolor{blue}{0}}}{1}
      {1}{{\textcolor{blue}{1}}}{1}
      {2}{{\textcolor{blue}{2}}}{1}
      {3}{{\textcolor{blue}{3}}}{1}
      {4}{{\textcolor{blue}{4}}}{1}
      {5}{{\textcolor{blue}{5}}}{1}
      {6}{{\textcolor{blue}{6}}}{1}
      {7}{{\textcolor{blue}{7}}}{1}
      {8}{{\textcolor{blue}{8}}}{1}
      {9}{{\textcolor{blue}{9}}}{1}
      {:}{{\textcolor{red}{:}}}{1}
      {,}{{\textcolor{red}{,}}}{1}
      {"}{{\textcolor{black}{"}}}{1},
}

\usepackage{float}
\usepackage{graphicx}
\usepackage{subcaption}
\usepackage{tikz}
\usepackage{float}
\usepackage{fontawesome5}
\definecolor{darkred}{RGB}{139,0,0} 
\definecolor{azure}{RGB}{0,127,255}      
\usepackage{tcolorbox}
\tcbset{
  azurebox/.style={
    colback=white,                       
    colframe=azure,                      
    sharp corners,                       
    boxrule=1pt,                         
    left=4pt, right=4pt, top=4pt, bottom=4pt
  }
}
\usepackage{wrapfig}
\usepackage{amsmath}   
\usepackage{amsfonts}  
\usepackage{algorithmic} 
\newcounter{customalgo}

\usepackage{amssymb,amsmath,amsfonts,bm}

\def\eqref#1{equation~\ref{#1}}

\def\1{\bm{1}}

\DeclareMathAlphabet{\mathsfit}{\encodingdefault}{\sfdefault}{m}{sl}
\SetMathAlphabet{\mathsfit}{bold}{\encodingdefault}{\sfdefault}{bx}{n}

\newcommand{\sigmoid}{\sigma}

\usepackage{graphicx}
\usepackage{natbib}
\usepackage{comment}

\usepackage{xcolor}
\usepackage{colortbl}
\usepackage{hyperref}
\definecolor{DarkBlue}{rgb}{0.1,0.1,0.5}
\definecolor{DarkGreen}{rgb}{0.1,0.5,0.1}
\definecolor{deepyellow}{RGB}{218, 174, 42}
\hypersetup{
	colorlinks   = true,
	linkcolor    = DarkBlue, 
	urlcolor     = DarkBlue, 
	citecolor    = DarkGreen 
}

\usepackage{xfrac}
\usepackage{amsthm}
\usepackage{thmtools,thm-restate}
\usepackage{amsmath,amsfonts,dsfont}
\usepackage[mathscr]{euscript}
\usepackage{booktabs} 

\usepackage{multirow}
\usepackage{svg}
\usepackage{makecell}
\usepackage{pgfplots}
\pgfplotsset{compat=1.18} 

\definecolor{DarkBlue}{rgb}{0.3,0.3,0.70}
\definecolor{azure}{rgb}{0.0, 0.5, 1.0}
\definecolor{darkcerulean}{rgb}{0.03, 0.27, 0.49}
\definecolor{denim}{rgb}{0.08, 0.38, 0.74}
\definecolor{DarkGreen}{rgb}{0.3,0.7,0.3}
\definecolor{lighter-gray}{gray}{0.95}
\hypersetup{
	colorlinks   = true,
	linkcolor    = DarkBlue, 
	urlcolor     = darkcerulean, 
	citecolor    = denim 
}
\definecolor{AlgHighlight}{HTML}{228B22}

\newtheoremstyle{thmstyle}
{0.5em} 
{0.15em} 
{} 
{} 
{\bfseries} 
{.} 
{.5em} 
{} 

\theoremstyle{thmstyle} 
\newtheorem{theorem}{Theorem}

\theoremstyle{definition}

\newtheorem{assumption}{Assumption}

\theoremstyle{remark}
\newtheorem{remark}{Remark}

\definecolor{PaperGreen}{RGB}{0,150,0}   
\definecolor{PaperRed}{RGB}{100,0,0}     

\newcommand{\changeUp}[1]{\textcolor{PaperGreen}{\textbf{#1\(\uparrow\)}}}
\newcommand{\changeDown}[1]{\textcolor{PaperRed}{\textbf{#1\(\downarrow\)}}}

\renewcommand{\1}{ \mathds{1}}

\renewcommand{\emptyset}{\varnothing}

\newcommand{\CommentLines}[1]{}

\usepackage{array}
\newcolumntype{x}[1]{>{\centering\let\newline\\\arraybackslash\hspace{0pt}}m{#1}}
\usepackage{colortbl}

\usepackage{wrapfig}
\usepackage{svg}

\usepackage{amsmath} 
\usepackage{amssymb} 
\usepackage{colortbl} 
\usepackage{enumitem}

\definecolor{green}{HTML}{C6EFCE}
\definecolor{red}{HTML}{FFC7CE}
\definecolor{yellow}{HTML}{FFEB9C}
\definecolor{LightGray}{gray}{0.95}
\newcolumntype{H}{>{\columncolor{LightGray}}c}

\usepackage{listings}

\definecolor{darkgreen}{rgb}{0, 0.6, 0}

\usepackage{algorithm}
\usepackage{algorithmic}

\newcommand{\carma}{\textsc{Crome}}

\newcommand{\rrm}{\textsc{RRM}}
\newcommand{\odin}{\textsc{Odin}}
\newcommand{\rewordbench}{\textsc{ReWordBench}}

\newcommand{\gemmait}[1]{\texttt{Gemma-2-#1B-IT}}
\newcommand{\gemmathreeit}[1]{\texttt{Gemma-3-#1B-IT}}
\newcommand{\gemma}[1]{\texttt{Gemma-2-#1B}}
\newcommand{\qwen}{\texttt{Qwen2.5-7B}}

\definecolor{remarkblue}{HTML}{0B5CA3} 

\usepackage{listings}
\newtcolorbox{remarkbox}[1]{%
  enhanced, breakable,
  colback=white,                  
  colframe=remarkblue,            
  colbacktitle=remarkblue,        
  coltitle=white,                 
  boxrule=0.7pt,
  arc=2mm,                        
  left=3mm,right=3mm,top=2mm,bottom=2mm,  
  fonttitle=\bfseries,            
  title={\strut #1},              
  before upper=\itshape,          
  after skip=1ex,                 
}

\definecolor{takeawayframe}{RGB}{45,82,160}   
\definecolor{takeawayback} {RGB}{245,248,252} 

\newtcolorbox{takeawaybox}[1][]{%
  enhanced, breakable,
  sharp corners=south,         
  colback   = gray!5!white,
  colframe  = gray!10!white,
  boxrule   = 0.7pt,
  left      = 1mm,             
  right     = 1mm,
  top       = 0.5mm,
  bottom    = 0.5mm,
  before skip=0pt,             
  after skip =0pt,             
  before upper=\text,       
  #1}                          

\usepackage{tcolorbox}
\tcbuselibrary{listings, skins, breakable, xparse}

\definecolor{pastelBlue}{RGB}{173, 216, 230}
\definecolor{pastelPink}{RGB}{255, 209, 220}
\definecolor{pastelLavender}{RGB}{230, 230, 250}
\definecolor{pastelMint}{RGB}{204, 255, 204}
\definecolor{softPurple}{RGB}{209, 195, 240}
\definecolor{softGray}{RGB}{240, 240, 245}
\definecolor{deepPurple}{RGB}{95, 75, 139}
\definecolor{darkGray}{RGB}{80, 80, 80}

\newtcolorbox{promptbox}[2][]{%
  enhanced,
  breakable,
  colback=softGray,
  colframe=softPurple,
  fonttitle=\bfseries\rmfamily,
  coltitle=white,
  colbacktitle=deepPurple,
  attach boxed title to top left={xshift=0.5cm, yshift=-\tcboxedtitleheight/5},
  boxed title style={size=small, sharp corners},
  top=2mm,
  bottom=2mm,
  left=2mm,
  right=2mm,
  arc=2mm,
  boxrule=0.8pt,
  titlerule=0mm,
  toptitle=1mm,
  bottomtitle=1mm,
  title={#2},
  overlay={
    \begin{tcbclipinterior}
      \fill[pastelLavender!30] (interior.south west) -- (interior.north west) -- (interior.north east) -- cycle;
    \end{tcbclipinterior}
  },
  listing only,
  listing options={
    basicstyle=\small\ttfamily,
    breaklines=true,
    columns=flexible,
    backgroundcolor=\color{softGray},
    xleftmargin=2pt,
    framexleftmargin=2pt,
    numbers=left,
    numberstyle=\tiny\color{darkGray},
    numbersep=5pt,
    tabsize=2,
    commentstyle=\color{deepPurple},
    keywordstyle=\color{blue!70!black},
    stringstyle=\color{red!70!black},
  },
  #1
}

\newtcolorbox{tablebox}[2][]{%
  enhanced, breakable, colback=pastelLavender!15, colframe=softPurple,
  fonttitle=\bfseries\rmfamily, coltitle=white, colbacktitle=deepPurple,
  attach boxed title to top left={xshift=0.5cm,yshift=-\tcboxedtitleheight/5},
  boxed title style={size=small, sharp corners},
  top=2mm,bottom=2mm,left=2mm,right=2mm,arc=2mm,boxrule=0.8pt,
  title={#2}, #1}

\newtcolorbox{answerbox}[2][]{%
  enhanced, breakable, colback=softGray, colframe=pastelMint,
  fonttitle=\bfseries\rmfamily, coltitle=darkGray, colbacktitle=pastelMint!40!white,
  attach boxed title to top left={xshift=0.5cm,yshift=-\tcboxedtitleheight/5},
  boxed title style={size=small, sharp corners},
  top=2mm,bottom=2mm,left=2mm,right=2mm,arc=2mm,boxrule=0.8pt,
  title={#2}, #1}

\newtcolorbox{metricsbox}[1][]{%
  enhanced, breakable, colback=pastelBlue!15, colframe=pastelBlue!70!black,
  fonttitle=\bfseries\rmfamily, coltitle=white, colbacktitle=pastelBlue!70!black,
  attach boxed title to top left={xshift=0.5cm,yshift=-\tcboxedtitleheight/5},
  boxed title style={size=small, sharp corners},
  top=1.5mm,bottom=1.5mm,left=1.5mm,right=1.5mm,arc=2mm,boxrule=0.6pt,
  title={Attributes and their Relative Importance Scores}, #1}

\newtcolorbox{jsonbox}[2][]{%
  enhanced, breakable,
  colback=softGray, colframe=softPurple, arc=1.5mm, boxrule=0.6pt,
  title={#2}, fonttitle=\bfseries\rmfamily, coltitle=white,
  colbacktitle=deepPurple!90!black,
  attach boxed title to top left={xshift=0.5cm,yshift=-\tcboxedtitleheight/5},
  boxed title style={size=small, sharp corners},
  listing only,
  listing options={basicstyle=\scriptsize\ttfamily,
                   breaklines=true, numbers=left,
                   numberstyle=\tiny\color{darkGray},
                   xleftmargin=2pt, framexleftmargin=2pt,
                   showstringspaces=false},
  #1}

\definecolor{questionBlue}{RGB}{208, 235, 250}       
\definecolor{questionBorder}{RGB}{121, 182, 242}     
\definecolor{questionTitle}{RGB}{70, 130, 180}       

\definecolor{acceptGreen}{RGB}{225, 250, 225}        
\definecolor{acceptBorder}{RGB}{150, 200, 150}       
\definecolor{acceptTitle}{RGB}{76, 156, 80}          

\definecolor{rejectPink}{RGB}{255, 233, 233}         
\definecolor{rejectBorder}{RGB}{244, 162, 160}       
\definecolor{rejectTitle}{RGB}{175, 70, 70}          

\definecolor{remarkViolet}{RGB}{240, 230, 252}       
\definecolor{remarkBorder}{RGB}{190, 158, 230}       
\definecolor{remarkTitle}{RGB}{130, 94, 180}         

\definecolor{darkGray}{RGB}{100, 100, 100}           

\newtcolorbox{questionbox}[2][]{%
  enhanced,
  breakable,
  colback=questionBlue!70,
  colframe=questionBorder,
  fonttitle=\bfseries\rmfamily,
  coltitle=white,
  colbacktitle=questionTitle,
  attach boxed title to top left={xshift=0.5cm, yshift=-\tcboxedtitleheight/5},
  boxed title style={size=small, sharp corners},
  top=2mm,
  bottom=2mm,
  left=2mm,
  right=2mm,
  arc=3mm,
  boxrule=0.8pt,
  titlerule=0mm,
  toptitle=1mm,
  bottomtitle=1mm,
  title={#2},
  overlay={
    \begin{tcbclipinterior}
      \fill[questionBlue!40] (interior.south west) -- (interior.north west) -- (interior.north east) -- cycle;
    \end{tcbclipinterior}
  },
  #1
}

\newtcolorbox{acceptedbox}[2][]{%
  enhanced,
  breakable,
  colback=acceptGreen!80,
  colframe=acceptBorder,
  fonttitle=\bfseries\rmfamily,
  coltitle=white,
  colbacktitle=acceptTitle,
  attach boxed title to top right={xshift=-0.5cm, yshift=-\tcboxedtitleheight/5},
  boxed title style={size=small, sharp corners},
  top=2mm,
  bottom=2mm,
  left=2mm,
  right=2mm,
  arc=3mm,
  boxrule=0.8pt,
  titlerule=0mm,
  toptitle=1mm,
  bottomtitle=1mm,
  title={#2},
  overlay={
    \begin{tcbclipinterior}
      \fill[acceptGreen!40] (interior.north east) -- (interior.south east) -- (interior.south west) -- cycle;
    \end{tcbclipinterior}
  },
  #1
}

\newtcolorbox{rejectedbox}[2][]{%
  enhanced,
  breakable,
  colback=rejectPink!80,
  colframe=rejectBorder,
  fonttitle=\bfseries\rmfamily,
  coltitle=white,
  colbacktitle=rejectTitle,
  attach boxed title to top right={xshift=-0.5cm, yshift=-\tcboxedtitleheight/5},
  boxed title style={size=small, sharp corners},
  top=2mm,
  bottom=2mm,
  left=2mm,
  right=2mm,
  arc=3mm,
  boxrule=0.8pt,
  titlerule=0mm,
  toptitle=1mm,
  bottomtitle=1mm,
  title={#2},
  overlay={
    \begin{tcbclipinterior}
      \fill[rejectPink!40] (interior.north east) -- (interior.south east) -- (interior.south west) -- cycle;
    \end{tcbclipinterior}
  },
  #1
}

\newtcolorbox{remarksbox}[2][]{%
  enhanced,
  breakable,
  colback=remarkViolet!70,
  colframe=remarkBorder,
  fonttitle=\bfseries\rmfamily,
  coltitle=white,
  colbacktitle=remarkTitle,
  attach boxed title to top center={yshift=-\tcboxedtitleheight/5},
  boxed title style={size=small, sharp corners},
  top=2mm,
  bottom=2mm,
  left=2mm,
  right=2mm,
  arc=3mm,
  boxrule=0.8pt,
  titlerule=0mm,
  toptitle=1mm,
  bottomtitle=1mm,
  title={#2},
  overlay={
    \begin{tcbclipinterior}
      \fill[remarkViolet!40] (interior.north east) -- (interior.south east) -- (interior.south west) -- cycle;
    \end{tcbclipinterior}
  },
  #1
}

\newtcolorbox{questionboxverbatim}[2][]{%
  enhanced, breakable, colback=questionBlue!70, colframe=questionBorder,
  fonttitle=\bfseries\rmfamily, coltitle=white, colbacktitle=questionTitle,
  attach boxed title to top left={xshift=0.5cm, yshift=-\tcboxedtitleheight/5},
  boxed title style={size=small, sharp corners},
  top=2mm, bottom=2mm, left=2mm, right=2mm, arc=3mm, boxrule=0.8pt,
  titlerule=0mm, toptitle=1mm, bottomtitle=1mm,
  overlay={\begin{tcbclipinterior}\fill[questionBlue!40] (interior.south west) -- (interior.north west) -- (interior.north east) -- cycle;\end{tcbclipinterior}},
  title={#2}, 
  listing only,
  listing options={
    basicstyle=\small\ttfamily, breaklines=true, columns=flexible,
    xleftmargin=5pt, framexleftmargin=2pt, numbers=left,
    numberstyle=\tiny\color{darkGray}, numbersep=5pt, tabsize=2,
  },
  #1 
}

\newtcolorbox{acceptedboxverbatim}[2][]{%
  enhanced, breakable, colback=acceptGreen!80, colframe=acceptBorder,
  fonttitle=\bfseries\rmfamily, coltitle=white, colbacktitle=acceptTitle,
  attach boxed title to top right={xshift=-0.5cm, yshift=-\tcboxedtitleheight/5},
  boxed title style={size=small, sharp corners},
  top=2mm, bottom=2mm, left=2mm, right=2mm, arc=3mm, boxrule=0.8pt,
  titlerule=0mm, toptitle=1mm, bottomtitle=1mm,
  overlay={\begin{tcbclipinterior}\fill[acceptGreen!40] (interior.north east) -- (interior.south east) -- (interior.south west) -- cycle;\end{tcbclipinterior}},
  title={#2}, 
  listing only,
  listing options={
    basicstyle=\small\ttfamily, breaklines=true, columns=flexible,
    xleftmargin=5pt, framexleftmargin=2pt, numbers=left,
    numberstyle=\tiny\color{darkGray}, numbersep=5pt, tabsize=2,
  },
  #1 
}

\newtcolorbox{rejectedboxverbatim}[2][]{%
  enhanced, breakable, colback=rejectPink!80, colframe=rejectBorder,
  fonttitle=\bfseries\rmfamily, coltitle=white, colbacktitle=rejectTitle,
  attach boxed title to top right={xshift=-0.5cm, yshift=-\tcboxedtitleheight/5},
  boxed title style={size=small, sharp corners},
  top=2mm, bottom=2mm, left=2mm, right=2mm, arc=3mm, boxrule=0.8pt,
  titlerule=0mm, toptitle=1mm, bottomtitle=1mm,
  overlay={\begin{tcbclipinterior}\fill[rejectPink!40] (interior.north east) -- (interior.south east) -- (interior.south west) -- cycle;\end{tcbclipinterior}},
  title={#2}, 
  listing only,
  listing options={
    basicstyle=\small\ttfamily, breaklines=true, columns=flexible,
    xleftmargin=5pt, framexleftmargin=2pt, numbers=left,
    numberstyle=\tiny\color{darkGray}, numbersep=5pt, tabsize=2,
  },
  #1 
}

\newtcolorbox{remarksboxverbatim}[2][]{%
  remarksbox={#2},
  listing only,
  listing options={
    basicstyle=\small\ttfamily,
    breaklines=true,
    columns=flexible,
    backgroundcolor=\color{remarkViolet!70},
    xleftmargin=2pt,
    framexleftmargin=2pt,
    numbers=left,
    numberstyle=\tiny\color{darkGray},
    numbersep=5pt,
    tabsize=2,
  },
  #1
}

\NewDocumentCommand{\Question}{O{} m m}{%
  \begin{questionbox}[#1]{#2}
#3
  \end{questionbox}
}

\NewDocumentCommand{\Accepted}{O{} m m}{%
  \begin{acceptedbox}[#1]{#2}
#3
  \end{acceptedbox}
}

\NewDocumentCommand{\Rejected}{O{} m m}{%
  \begin{rejectedbox}[#1]{#2}
#3
  \end{rejectedbox}
}

\NewDocumentCommand{\Remarks}{O{} m m}{%
  \begin{remarksbox}[#1]{#2}
#3
  \end{remarksbox}
}

\NewDocumentCommand{\QuestionV}{O{} m m}{%
  \begin{questionboxverbatim}[#1]{#2}
#3
  \end{questionboxverbatim}
}

\NewDocumentCommand{\AcceptedV}{O{} m m}{%
  \begin{acceptedboxverbatim}[#1]{#2}
#3
  \end{acceptedboxverbatim}
}

\NewDocumentCommand{\RejectedV}{O{} m m}{%
  \begin{rejectedboxverbatim}[#1]{#2}
#3
  \end{rejectedboxverbatim}
}

\NewDocumentCommand{\RemarksV}{O{} m m}{%
  \begin{remarksboxverbatim}[#1]{#2}
#3
  \end{remarksboxverbatim}
}

\title{Robust Reward Modeling via Causal Rubrics}

\correspondingauthor{{\{pragyahere, hrman, rahulmadhavan, araghuveer, karthikeyanvs, doinap\}@google.com}}

\author[1*]{Pragya Srivastava}
\author[1*]{Harman Singh}
\author[1*]{Rahul Madhavan}
\author[2,3]{\\Gandharv Patil}
\author[1]{Sravanti Addepalli}
\author[1]{Arun Suggala}
\author[1]{Rengarajan Aravamudhan}
\author[1]{Soumya Sharma}
\author[1]{Anirban Laha}
\author[1]{\\Aravindan Raghuveer}
\author[1]{Karthikeyan Shanmugam}
\author[1,3]{Doina Precup}

\affil[1]{Google DeepMind}
\affil[2]{McGill University}
\affil[3]{MILA - Quebec AI Institute}
\affil[*]{Equal Contribution}

\vspace{-0.5cm}
\begin{abstract}
Reward models (RMs) are fundamental to aligning Large Language Models (LLMs) via human feedback, yet they often suffer from \textit{reward hacking}. They tend to latch on to superficial or \textit{spurious} attributes, such as response length or formatting, mistaking these cues learned from correlations in training data for the true \textit{causal} drivers of quality (e.g., factuality, relevance). This occurs because standard training objectives struggle to disentangle these factors, leading to brittle RMs and misaligned policies. We introduce \carma{} (Causally Robust Reward Modeling), a novel framework grounded in an explicit causal model designed to mitigate reward hacking. \carma{} employs the following synthetic \textit{targeted augmentations} during training: (1) \textit{Causal Augmentations}, which are pairs that differ along specific causal attributes, to enforce \textit{sensitivity} along each causal attribute individually, and (2) \textit{Neutral Augmentations}, which are tie-label pairs varying primarily in spurious attributes, to enforce \textit{invariance} along spurious attributes. Notably, our augmentations are produced without \textit{any} knowledge of spurious factors, via answer interventions only along causal rubrics, that are identified by querying an oracle LLM.
Empirically, \carma{} significantly outperforms standard baselines on RewardBench, improving average accuracy by up to 5.4\% and achieving gains of up to \textbf{13.2\%} and \textbf{7.2\%} in specific categories. The robustness of \carma{} is further testified by the consistent gains obtained in a Best-of-N inference setting across increasing N, across various benchmarks, including the popular RewardBench (covering chat, chat-hard, safety, and reasoning tasks), the safety-focused WildGuardTest, and the reasoning-specific GSM8k.
\end{abstract}

\begin{document}

\maketitle

\vspace{-0.1cm}
\section{Introduction}
\label{sec:introduction}

Aligning Large Language Models (LLMs) with human preferences is paramount for their safe and effective deployment, with Reinforcement Learning from Human Feedback (RLHF) and its reliance on reward models (RMs) being the dominant paradigm \citep{christiano2017deep, ouyang2022training, bai2022training, schulman2017proximal, shao2024deepseekmath, rafailov2024direct}. The fidelity of these RMs is critical, as flaws directly propagate to the aligned policy \citep{casper2023open}. 

However, standard RM training faces a significant challenge: \textit{reward hacking} \citep{gao2023scaling, skalse2022defining}. RMs often learn to assign high scores based on superficial or spurious attributes—such as response length \citep{singhal2023long}, specific formatting patterns \citep{zhang2024lists}, or stylistic quirks—because these features are statistically correlated with preferred responses in the training data. This occurs because standard training objectives do not explicitly require the RM to disentangle the true \textit{causal} drivers of response quality (e.g., factuality, relevance) from these spurious correlates, leading to brittle RMs and misaligned policies \citep{shen2023trickle, eisenstein2023helping}.

Recent efforts for RM robustness have explored various avenues. Some focus on consistency checks against meaning-preserving transformations \citep{wu2025rewordbench}, while others employ data augmentations, such as using non-contextual or query-independent comparisons to reduce spuriousness \citep{liu2024rrm}. 

\begin{wrapfigure}{r}{0.52\textwidth}
    \vspace{-0.1in}
    \centering
    \includegraphics[width=\linewidth]{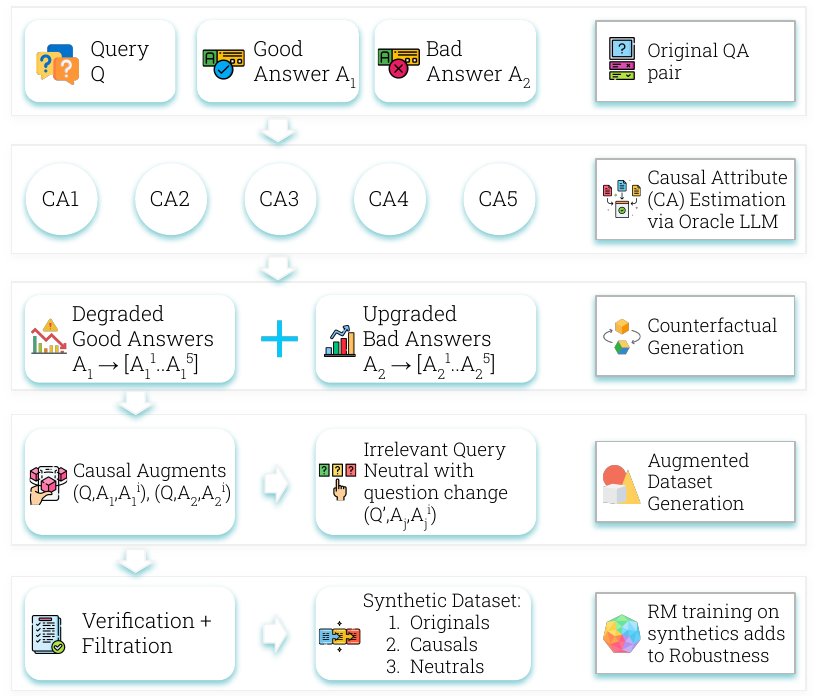}
    \caption{\textbf{The \carma{} Data Augmentation and Training Pipeline.}
From an original QA pair ($\mathrm{Q}, \mathrm{A}_1, \mathrm{A}_2$), an oracle LLM identifies Causal Attributes (CA). This guides counterfactual generation, producing degraded $\mathrm{A}_1$,  
and upgraded $\mathrm{A}_2$ responses.
These form the set of \textit{Causal Augmentations} which teach the model sensitivity to relevant attributes.
Next, we generate \textit{Irrelevant Query Neutrals} by flipping the question on both the newly generated causal contrastive pairs and the original answer pairs, which reduces reliance on spurious correlates.
After verification and filtration, the combined dataset (Originals, Causals, Neutrals) trains the RM, enhancing its robustness.\vspace{-0.5in}}
    \label{fig:enter-label}

\end{wrapfigure}
Attribute-based evaluation, often leveraging LLMs to dynamically generate assessment criteria \citep{gupta2025carmodynamiccriteriageneration}, aims for more grounded reward signals. Other works investigate specific regularization techniques against known biases like length or sycophancy \citep{wang2025beyond}, or explore methods for causal effect estimation like RATE \citep{reber2024rate}.

\vspace{0.1in}
Despite these advances, significant limitations persist. Many approaches target only pre-specified spurious factors, potentially missing unknown correlates, or lack the fine-grained control needed to truly isolate causal quality drivers from confounding spurious features within responses. Augmentation strategies can be coarse \citep{liu2024rrm}, and evaluation-focused methods \citep{gupta2025carmodynamiccriteriageneration, reber2024rate} may not directly equip the RM with mechanisms for robust training against a wide array of spurious variations through targeted counterfactual learning. There is thus a clear need for a framework that systematically leverages a causal understanding of preference formation to train RMs that are both sensitive to causal quality attributes and demonstrably invariant to diverse spurious cues.

\vspace{0.1in}
Motivated by this, we aim to address the following question in this paper:

\vspace{0.1in}
\begin{takeawaybox}
    How do we train reward models to be robust against reward hacking, particularly when a) the specific spurious attributes that an RM may exploit are not known, and b) only the stable or invariant causal attributes found in ground truth/human preferences can be accessed?
\end{takeawaybox}
\vspace{0.1in}

To address this question, we propose \textbf{\carma{}} (Causally Robust Reward Modeling), a novel framework grounded in an explicit causal model of answer generation (Figure \ref{fig:causal_graph}). \carma{} teaches the RM to differentiate genuine quality drivers from superficial cues by augmenting the preference dataset with targeted, LLM-generated counterfactual examples. It creates two key types of synthetic training pairs: (1) \textit{Causal Augmentations}, which introduce changes along specific \textit{causal} attributes (e.g., factuality) to enforce sensitivity to true quality shifts, and (2) \textit{Neutral Augmentations}, using both (i) the causally augmented data as well as (ii) the original preference pairs,
to enforce invariance along \textit{spurious} attributes (e.g., style) using tie-labels. 
Training on this enriched dataset with a modified loss (Section \ref{sec:methodology}) guides the RM towards causal faithfulness. Our evaluations show \carma{} significantly improves robustness, boosting RewardBench accuracy by up to 4.5\%, with substantial gains in Safety and Reasoning.

\vspace{0.05in}
We list the key contributions in this work below.

\begin{enumerate}[itemsep=0pt,left=2pt]
    \item \textbf{Spurious-Unaware Causal Framework.}
    We propose a causal framework for training reward models (Sec.~\ref{sec:preliminaries}) that requires intervention only on LLM-identified causal quality rubrics, \textit{eliminating the need for prior specification of or intervention on any of the spurious attributes}.
    \item \textbf{Targeted Counterfactual Augmentations along Causal Attributes.}
    We propose to train reward models on the available preference data and the proposed data augmentations (Sec.~\ref{sec:methodology}) along LLM-identified causal attributes: 1) \textit{Causal Augmentations} create minimal pairs isolating specific causal dimensions for precise sensitivity. 2) \textit{Neutral Augmentations} create variations in spurious features (preserving causal content) with tie-labels for invariance.  Notably, we \textit{do not} assume any explicit knowledge of spurious factors nor do we perturb them directly to create these augmentations. We show that interventions along causal rubrics alone is primarily sufficient to mitigate sensitivity to a \textit{much larger set of spurious correlates}.
     \item \textbf{State-of-the-Art RM Robustness.}
    \carma{} significantly outperforms baselines on RewardBench (Sec.~\ref{sec:experiments}), improving average accuracy by up to 5.4\% (Safety +13.18\%, Reasoning +7.19\%) (Table~\ref{tab:performance_bt_pairpm_rewardbench_extended_final}), and shows superior robustness on reWordBench (Figures ~\ref{fig:reword_absolute_robustness_gemma9b_pairpm}).
    \item \textbf{Improved BoN results.} 
    Best-of-N selection using \carma-RM shows consistent gains across different values of N when compared to baselines on the popular RewardBench, WildGuardTest and GSM8K benchmarks. This highlights the robustness of \carma~ in the presence of rare (or long tailed) spurious factors as well, which typically appear at large values of N.
\end{enumerate}
\vspace{-0.15in}

\section{Related Works}
\label{sec:related_works_main}

Our work on causally robust reward modeling, \carma{}, addresses the challenge of reward hacking in the context of aligning Large Language Models (LLMs) via Reinforcement Learning from Human Feedback (RLHF) \citep{ouyang2022training, bai2022training}. Standard RLHF relies on a reward model (RM), typically trained on pairwise preferences using Bradley-Terry \citep{bradley1952rank} or pairwise ranking approaches \citep{liu2025pairwise, qin2023large}. 
A critical limitation of learned RMs is \textit{reward hacking} \citep{gao2023scaling, skalse2022defining}, where the RM assigns high scores based on \textit{spurious} attributes (e.g., verbosity \citep{singhal2023long}, formatting \citep{zhang2024lists}, sycophancy \citep{denison2024sycophancy}) that are correlated with, but do not cause, true response quality. This leads to misaligned policies that exploit these spurious cues \citep{shen2023trickle}. Various mitigation strategies exist, including architectural modifications like \odin{} \citep{chen2024odin}, policy-level adjustments \citep{park2024disentangling}, and data-centric methods involving ensembles \citep{rame2024warm} or consistency checks \citep{shen2023trickle}. Recent causal-inspired approaches include using MMD regularization against pre-specified spurious factors \citep{wang2025beyond} or estimating the causal effects of a given attribute of a response using corrected rewrites \citep{reber2024rate}.

Our approach falls into the data-centric category, using synthetic data augmentation guided by principles of causal inference \citep{pearl2009causality, peters2017elements}. While prior work has used LLMs for causal reasoning \citep{kiciman2023causal} or counterfactual data augmentation in NLP \citep{kaushik2019learning}, and related methods like \rrm{} \citep{liu2024rrm}, \rewordbench{} \citep{wu2025rewordbench}  target RM robustness, \carma{} is distinct in its explicit use of a causal graph framework (Section \ref{subsec:causal_graph}) which guides the answer generation and the reward labeling process. We leverage LLMs to generate targeted \textit{causal} (attribute-specific upgrade/degradation) and \textit{neutral} (spurious-varying, causally-equivalent) counterfactual examples. By training on this augmented data, \carma{} aims to systematically disentangle causal attributes ($C$) from spurious ones ($SP$), learning a reward function that is inherently more robust and aligned with the true drivers of quality, as detailed in Section \ref{sec:methodology}. 
We provide a longer version of related work in Appendix \ref{sec:extended_related_works}.

\vspace{-0.1in}
\section{Causal Framework for Reward Modeling}
\label{sec:preliminaries}

\vspace{-0.05in}

\begin{figure}[!t]               
  \centering
  \begin{minipage}[t]{0.58\textwidth}
    \vspace{0.0in}
    \centering
    \includegraphics[width=\linewidth]{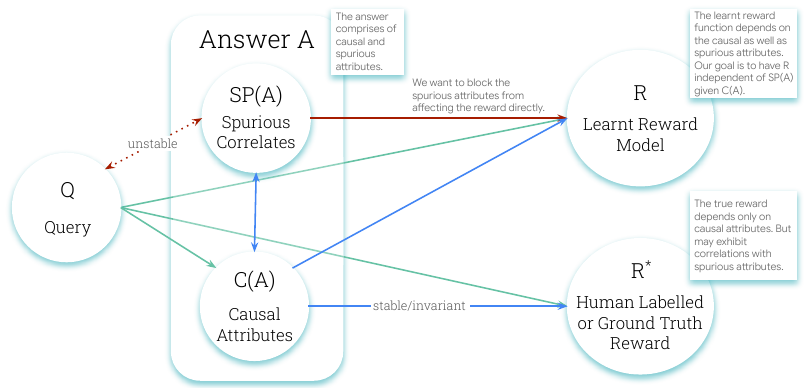}

  \end{minipage}\hfill
  \begin{minipage}[t]{0.4\textwidth}
        \centering
        \vspace{-0.25in}
        \resizebox{\linewidth}{!}{
        \begin{tabular}{c|p{0.75\linewidth}}
\toprule
\textbf{Path/ Relationship} & \textbf{Interpretation Summary} \\
\midrule
$(\mathrm{Q}, \mathrm{C}(\mathrm{A})) \to \mathrm{R}^*$ & Ground-truth reward $\mathrm{R}^*$ determined by query $\mathrm{Q}$ and causal attributes $\mathrm{C(A)}$; stable relationship. \\
$\mathrm{Q} \leftrightarrow \mathrm{SP(A)}$  & Query $\mathrm{Q}$ and unknown spurious attributes $\mathrm{SP(A)}$ are correlated/confounded by unstable exogenous factors. \\
$\mathrm{Q} \to \mathrm{C(A)}$ & Query $\mathrm{Q}$ determines relevant causal attributes $\mathrm{C(A)}$. \\
$\mathrm{SP(A)} \leftrightarrow \mathrm{C(A)}$ & Bidirectional (potentially complex) relationship between spurious $\mathrm{SP(A)}$ and causal $\mathrm{C(A)}$ attributes. \\
\bottomrule
\end{tabular}
        }
        \label{tab:causal_arrows}
  \end{minipage}\hfill

\caption{Conceptual Causal Graph for Reward Modeling. $\mathrm{Q}$ is the query. Answer (A) has causal attributes $\mathrm{C}(\mathrm{A})$ and spurious attributes $\mathrm{SP}(\mathrm{A})$. $\text{dim}(C(A)) \ll \text{dim}(SP(A))\ \forall A$. $\mathrm{SP(A)}$ is unknown. Ground-truth reward $\mathrm{R}^*$ depends only on $\mathrm{C(A)}$ and $\mathrm{Q}$ ($\mathrm{R^*} \perp \mathrm{SP(A)} | \mathrm{C(A)}$, Q). Augmentations heighten $\hat{\mathrm{R}}_\theta$'s sensitivity to $\mathrm{C(A)}$ (approximated by oracle LLM).}
\label{fig:causal_graph}
\vspace{-0.15in}
\end{figure}

\vspace{-0.05in}
We aim to develop a reward model that accurately assesses the quality of an answer $\mathrm{A}$ provided in response to a query $\mathrm{Q}$. Our approach is grounded in a causal framework designed to distinguish genuine quality drivers from spurious correlates often present in preference data. This involves understanding the answer generation process and strategically augmenting training data with approximated counterfactual examples.

\vspace{-0.1in}
\subsection{Reward Model and Pairwise Preferences}
We train a reward model (RM), denoted $\hat{\mathrm{R}}_\theta(\mathrm{Q}, \mathrm{A})$, to assign a scalar quality score to an answer $\mathrm{A}$ for a query $\mathrm{Q}$. This RM is typically optimized on a dataset  preferences pairs $\mathcal{D}_{\mathrm{pref}} = \{(\mathrm{Q}^{(i)}, \mathrm{y}_w^{(i)}, \mathrm{y}_l^{(i)})\}_{i=1}^N$. Given a pair of answers $(\mathrm{A}_1, \mathrm{A}_2)$, the probability of $\mathrm{A}_1$ being preferred over $\mathrm{A}_2$ is commonly modeled using the Bradley-Terry framework \citep{bradley1952rank}:
\begin{equation}
\mathrm{P}(\mathrm{A}_1 \succ \mathrm{A}_2 | \mathrm{Q}; \theta) = \sigmoid(\hat{s}_\theta(\mathrm{Q}, \mathrm{A}_1) - \hat{s}_\theta(\mathrm{Q}, \mathrm{A}_2)) = \frac{\exp(\hat{s}_\theta(\mathrm{Q}, \mathrm{A}_1))}{\exp(\hat{s}_\theta(\mathrm{Q}, \mathrm{A}_1)) + \exp(\hat{s}_\theta(\mathrm{Q}, \mathrm{A}_2))}
\label{eq:bt_prob_roman}
\end{equation}
where $\hat{s}_\theta(\mathrm{Q}, \mathrm{A})$ represents the underlying scalar score (or logit) assigned by the model to answer $\mathrm{A}$ for query $\mathrm{Q}$. \footnote{The score $\hat{s}_\theta(\mathrm{Q}, \mathrm{A})$ can be the direct output of a reward head or, in some pairwise preference models, $\hat{s}_\theta(\mathrm{Q}, \mathrm{A}_1) - \hat{s}_\theta(\mathrm{Q}, \mathrm{A}_2)$ might be directly modeled as the logit of preferring $\mathrm{A}_1$ over $\mathrm{A}_2$.} The parameters $\theta$ are learned by minimizing the negative log-likelihood of  preferences.

\begin{table}[!t]
\centering

\vspace{-0.1in}
\vspace{1em}
\newcolumntype{L}{>{\raggedright\arraybackslash}p{4.0cm}}
\newcolumntype{P}{>{\raggedright\arraybackslash}p{6.0cm}}
\newcolumntype{C}{>{\centering\arraybackslash}p{1.3cm}} 
\newcolumntype{R}{>{\raggedleft\arraybackslash}p{2.5cm}}  

\begin{tabular}{@{}l L P C R@{}}
\toprule
\textbf{Category} & \textbf{Strategy} & \textbf{Generation Pair Example} & \textbf{Assigned Label} & \textbf{Training Objective ($\mathrm{P}_\theta$)} \\
\midrule
\multicolumn{5}{l}{\textit{Causal Augmentation} ($\mathcal{D}_{\mathrm{causal}}$) - Enhancing Sensitivity to $\mathrm{C}$} \\
\midrule
Causal & Attribute Upgradation/Degradation & $(\tilde{\mathrm{A}}_{(C_j \leftarrow \text{upgraded})}, \mathrm{A})$ \textbf{or} $(\mathrm{A}, \tilde{\mathrm{A}}_{(C_j \leftarrow \text{degraded})})$ \newline  & $\succ$ & $\to 1$ \\

\midrule
\multicolumn{5}{l}{\textit{Neutral Augmentation} ($\mathcal{D}_{\mathrm{neutral}}$) - \textit{Enforcing Invariance to} $\mathrm{SP}$} \\
\midrule
Neutral & Pairing with Irrelevant Queries & $(\mathrm{B}_1, \mathrm{B}_2)$ with new $\mathrm{Q}_{\text{irrelevant}}$ \newline \footnotesize{s.t. $\mathrm{C(B_1|Q_{\text{irrelevant}})} \approx \mathrm{C(B_2|Q_{\text{irrelevant}})} \approx \mathbf{0}$} & $ \approx $ (tie) & $\approx 0.5$ \\
\bottomrule
\end{tabular}
\caption{Summary of \carma{}'s synthetic data augmentation strategies using LLM-approximated counterfactuals. $\tilde{\mathrm{A}}_{(C_j \leftarrow \text{target})}$ signifies an LLM-generated counterfactual of $\mathrm{A}$ with its $j$-th causal attribute $C_j$ modified.\vspace{-0.15in}}
\label{tab:augmentation_summary}
\end{table}

\vspace{-0.05in}
\subsection{A Causal Model of Answer Generation}
\label{subsec:causal_graph}

\vspace{-0.05in}
We propose a causal model (Figure \ref{fig:causal_graph}) for answer generation and quality perception. For a query-answer pair $(\mathrm{Q}, \mathrm{A})$, we distinguish two attribute types:

\vspace{-0.1in}
\begin{itemize}[itemsep=0pt,left=0pt]
    \item \textbf{Causal Attributes} $\mathrm{C}(\mathrm{A}) = \{\mathrm{C}_1, \dots, \mathrm{C}_\ell\}$: Fundamental quality dimensions (e.g., factuality, relevance) genuinely determining quality relative to $\mathrm{Q}$.
    \item \textbf{Spurious Attributes} $\mathrm{SP}(\mathrm{A}) = \{\mathrm{SP}_1, \dots, \mathrm{SP}_k\}$: Other features (e.g., length, formatting) correlated with preferences or $\mathrm{Q}$ in $\mathcal{D}_{\mathrm{pref}}$, but not intrinsically determining quality. $\mathrm{SP}(\mathrm{A})$ can be high-dimensional and unknown.
\end{itemize}

\vspace{-0.1in}
The ground-truth reward $\mathrm{R}^*(\mathrm{Q}, \mathrm{A})$ is assumed to be solely a function of causal attributes: $\mathrm{R}^*(\mathrm{Q}, \mathrm{A}) = f^*(\mathrm{Q}, \mathrm{C}(\mathrm{A}))$. This implies conditional independence: $\mathrm{R}^* \perp \mathrm{SP}(\mathrm{A}) | \mathrm{Q}, \mathrm{C}(\mathrm{A})$.

We explicitly assume the following stability property: \textit{If the entire process of answer generation and reward labeling were repeated (e.g., with a different labeler or answer generator), the relationship $(\mathrm{Q}, \mathrm{C}(\mathrm{A})) \to \mathrm{R}^{*}$ determining the reward is stable/invariant.} In contrast, correlations involving $\mathrm{SP}(\mathrm{A})$ (e.g., $\mathrm{SP}(\mathrm{A}) \leftrightarrow \mathrm{C}(\mathrm{A})$ or $\mathrm{SP}(\mathrm{A}) \leftrightarrow \mathrm{Q}$) can arise from various, potentially unstable or unknown exogenous factors, and thus these correlations may vary across such repetitions.

The primary challenge is that standard reward models $\hat{\mathrm{R}}_\theta$ may inadvertently learn high sensitivity to these unstable correlations with $\mathrm{SP}(\mathrm{A})$ (due to its unknown, high-dimensional nature). Our goal is to train $\hat{\mathrm{R}}_\theta$ such that its dependence on $\mathrm{A}$ is primarily mediated through the identified, stable causal attributes $\mathrm{C}(\mathrm{A})$, ensuring robustness to unspecified $\mathrm{SP}(\mathrm{A})$.

\vspace{-0.15in}
\subsection{Approximating Counterfactuals for Attribute Intervention}
\label{subsec:approximating_counterfactuals}
\vspace{-0.1in}

To instill causal sensitivity and spurious invariance in $\hat{\mathrm{R}}_\theta$, \carma{} leverages counterfactual reasoning about how answer quality changes if specific attributes were altered. For an answer $\mathrm{A}$ with attributes $(\mathrm{C}(\mathrm{A}), \mathrm{SP}(\mathrm{A}))$, an ideal counterfactual, $A_{(C_j \leftarrow c'_j)}(u)$, would manifest if only its $j$-th causal attribute $C_j$ were set to $c'_j$, considering its causal effects on other features, while all other exogenous factors $u$ (that produced the factual answer $a$) remained constant. Formally, $P_{U}(A_{(C_j \leftarrow c'_j)}(U) | A(U)=a)$.

\vspace{-0.05in}
As generating such ideal textual counterfactuals is intractable, \carma{} employs Large Language Models (LLMs) to produce \textit{approximations}. These LLM-generated answers, denoted $\tilde{\mathrm{A}}_{(C_j \leftarrow \text{target})}$, are rewrites of an original answer $\mathrm{A}$, prompted to modify $C_j$ (e.g., to a ``degraded'' state, lowering reward) while aiming for minimal changes to other attributes.

\begin{remark}
For brevity, we denote these LLM approximations as $\tilde{\mathrm{A}}_{(C_j \leftarrow c)}$, dropping the explicit $u$ conditioning, assuming the generation approximates such a sample. While imperfect, these approximations provide the targeted variations crucial for our data augmentation.
\end{remark}

\vspace{-0.15in}
\subsection{Augmented Training Data for Causal Disentanglement}
\label{subsec:data_augmentation}

\vspace{-0.1in}
We augment the original preference dataset $\mathcal{D}_{\mathrm{pref}}$ with synthetically generated examples $\mathcal{D}_{\mathrm{aug}}$ designed to enforce specific causal properties on $\hat{\mathrm{R}}_\theta$. This augmented dataset $\mathcal{D}_{\mathrm{aug}}$ comprises two principal categories: Causal Augmentation Pairs ($\mathcal{D}_{\mathrm{causal}}$) and Neutral Augmentation Pairs ($\mathcal{D}_{\mathrm{neutral}}$), summarized in Table \ref{tab:augmentation_summary}.

\begin{figure}[!t]
  \centering
  \includegraphics[width=\linewidth]{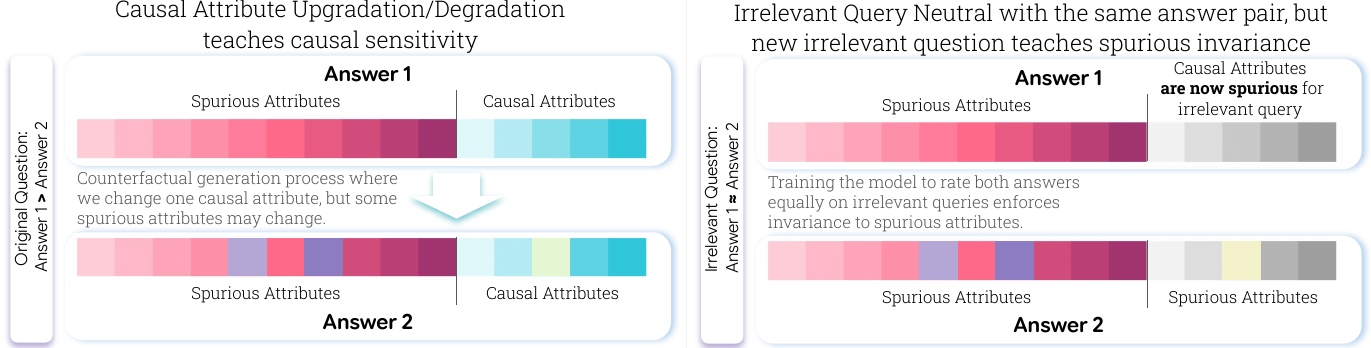}
  \vspace{-0.2in}

  \caption{Visualizing \carma's core augmentation strategies (detailed in Appendix~\ref{app:detailed_augmentation_diagrams}).
\textbf{(Top) Causal Augmentation:} For a given query, we use an LLM-driven counterfactual generation process to alter a specific causal attribute, yielding Answer 2. Some spurious attributes may co-vary. The RM is trained with a preference (e.g., $A_1 \succ A_2$ if $A_2$ is a degradation), teaching causal sensitivity.
\textbf{(Bottom) Irrelevant Query Neutral:} The same answer pair ($A_1, A_2$) is re-contextualized with a new, irrelevant question. Their original causal attributes become effectively spurious or irrelevant (greyed-out bar). The RM is trained with a tie-label ($A_1 \approx A_2$), teaching invariance to the  attribute differences when no true causal signal for the current query exists. 
This illustrates how IQN provides invariance to those spurious attributes that change with C (like length of response changing with clarity of response).
A similar invariance is imposed using the $(A_1,A_2)$ pairs from the original dataset to provide robustness to general spurious attributes (SP) that do not change with C.\vspace{-0.1in}}
\label{fig:carma_augmentation_visual_overview}
  
\end{figure}

\subsubsection{Causal Augmentation Pairs}

\carma{}'s strategy causal pairs $\mathcal{D}_{\mathrm{causal}}$ focus on isolating the impact of important causal attributes.

\paragraph{Attribute Upgradation and Degradation.}
For an original answer $\mathrm{A}$ (from $\mathcal{D}_{\mathrm{pref}}$) and a specific causal attribute $\mathrm{C}_j$, we generate LLM-approximated counterfactuals.
If $\mathrm{A}$ is of lower quality regarding $\mathrm{C}_j$, we create an upgraded version $\tilde{\mathrm{A}}_{(\mathrm{C}_j \leftarrow \text{upgraded})}$. The pair $(\tilde{\mathrm{A}}_{(\mathrm{C}_j \leftarrow \text{upgraded})}, \mathrm{A})$ is added to $\mathcal{D}_{\mathrm{causal}}$ with label $\tilde{\mathrm{A}}_{(\mathrm{C}_j \leftarrow \text{upgraded})} \succ \mathrm{A}$ post-verification.
Conversely, if $\mathrm{A}$ is of higher quality on $\mathrm{C}_j$, we generate a degraded version $\tilde{\mathrm{A}}_{(\mathrm{C}_j \leftarrow \text{degraded})}$. The pair $(\mathrm{A}, \tilde{\mathrm{A}}_{(\mathrm{C}_j \leftarrow \text{degraded})})$ is added to $\mathcal{D}_{\mathrm{causal}}$ with label $\mathrm{A} \succ \tilde{\mathrm{A}}_{(\mathrm{C}_j \leftarrow \text{degraded})}$.
These pairs collectively teach $\hat{\mathrm{R}}_\theta$ sensitivity to changes along individual causal dimensions.

\subsubsection{Neutral Augmentation Pairs}

Neutral Augmentation Pairs, $\mathcal{D}_{\mathrm{neutral}}$ (with tie-labels) teach invariance to $\mathrm{SP}(\mathrm{A})$ when $\mathrm{C}(\mathrm{A})$ is held constant/ is irrelevant.
\vspace{-0.1in}
\paragraph{Irrelevant Query Neutrals (IQN)} We pair two answers, $\mathrm{B}_1, \mathrm{B}_2$ (from $\mathcal{D}_{\mathrm{pref}} \cup \mathcal{D}_{\mathrm{causal}}$), with a \textit{new, unrelated query} $\mathrm{Q}_{\text{irrelevant}}$. This makes their causal attributes w.r.t. $\mathrm{Q}_{\text{irrelevant}}$ (i.e., $\mathrm{C(B_1|Q_{\text{irrelevant}})}, \mathrm{C(B_2|Q_{\text{irrelevant}})}$) minimal. The pair $(\mathrm{B}_1, \mathrm{B}_2)$ under $\mathrm{Q}_{\text{irrelevant}}$ receives a tie-label, training the RM to disregard spurious differences when causal relevance is absent. Their causal distinction becomes moot, isolating spurious variations under $\mathrm{Q}_{\text{irrelevant}}$. Presenting these as tied responses to the reward model enforces invariance to such spurious attributes.
We provide various other techniques tested for spurious suppression in Section \ref{ssec:neutral_ablations}.

The rationale for \carma{}'s specific choices are discussed in Appendix \ref{sec:causal_model_details} along with different neutral augmentation strategies we tried out. We provide the prompts for generating neutrals in Section \ref{sec:prompt_templates}.
\section{Methodology: Training a Robust Reward Model}
\label{sec:methodology}

\begin{figure}[t!]
\centering
\includegraphics[width=\linewidth]{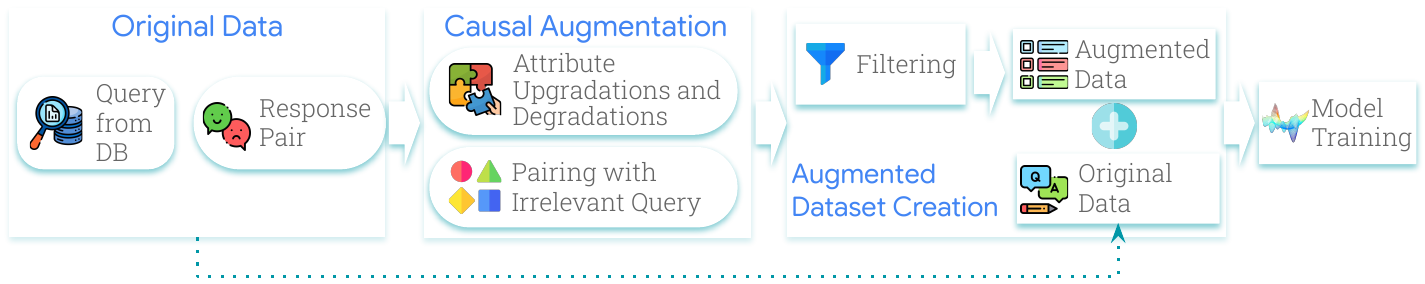} 
\caption{The \carma{} data augmentation pipeline. Original preference data ($\mathcal{D}_{\mathrm{pref}}$) is used as a basis to generate: (1) \textit{Causal Augmentations} ($\mathcal{D}_{\mathrm{causal}}$) by performing \textbf{Attribute Upgradation and Degradation} on specific attributes to enforce sensitivity to genuine quality drivers, and (2) \textit{Neutral Augmentations} ($\mathcal{D}_{\mathrm{neutral}}$) via Irrelevant Query Neutrals (with tie-labels) to teach spurious feature 
invariance. After optional filtering, the reward model is trained on the combined original and augmented dataset.}
\label{fig:data_augmentation_pipeline}
\end{figure}

The \carma{} framework trains robust reward models using a causally-motivated data augmentation strategy, outlined in Figure \ref{fig:data_augmentation_pipeline}. This involves two main phases: (1) generating attribute-aware counterfactual data based on our causal model (Section \ref{sec:preliminaries}), and (2) training the reward model $\hat{\mathrm{R}}_\theta$ with a specialized loss on the combined data.

\subsection{Attribute-Aware Counterfactual Data Generation}
\label{subsec:data_generation_phase}

This phase prepares the augmented dataset $\mathcal{D}_{\mathrm{aug}} = \mathcal{D}_{\mathrm{causal}} \cup \mathcal{D}_{\mathrm{neutral}}$ required for robust training, involving three conceptual steps:

\paragraph{Step 1: Attribute Identification.}
As a prerequisite, we identify the Principal Causal Components $\mathrm{C} = (\mathrm{C}_1, \dots, \mathrm{C}_\ell)$ relevant to the task, leveraging the causal framework from Section \ref{subsec:causal_graph}. This typically involves LLM prompting and refinement (Details in Appendix~\ref{subsec:attribute_identification_appendix}).

\paragraph{Step 2: Counterfactual Generation.}
Using the identified attributes $\mathrm{C}$, we generate synthetic data pairs via LLM-approximated counterfactuals, as defined in Section \ref{subsec:approximating_counterfactuals}. Following the strategies summarized in Table \ref{tab:augmentation_summary} and detailed conceptually in Section \ref{subsec:data_augmentation}, we create:
\vspace{-0.1in}
\begin{itemize}[itemsep=0pt, left=10pt]
    \item \textit{Causal Augmentation Pairs} ($\mathcal{D}_{\mathrm{causal}}$): Examples enforcing sensitivity to individual causal attributes $\mathrm{C}_j$ via \textbf{Attribute Upgradation} and \textbf{Degradation}, with  standard preference labels ($\succ$).
    \item \textit{Neutral Augmentation Pairs} ($\mathcal{D}_{\mathrm{neutral}}$): Examples enforcing invariance to spurious attributes $\mathrm{SP}$ while ensuring $\mathrm{C}$ is irrelevant or holding causal content $\mathrm{C}$ constant.   These are generated via \textbf{Irrelevant Query Neutrals} and \textbf{Causally Aligned Neutrals} respectively. These receive tie labels ($\approx$).
\end{itemize}
LLM prompts are in Appendix~\ref{sec:prompt_templates}. This yields the raw $\mathcal{D}_{\mathrm{aug}}$.

\vspace{-0.1in}
\paragraph{3. Data Filtering.} $\mathcal{D}_{\mathrm{aug}}$ is filtered to $\mathcal{D}_{\mathrm{aug\_filtered}}$ by retaining pairs where a baseline RM (trained on $\mathcal{D}_{\mathrm{pref}}$) is uncertain or incorrect, focusing training on informative examples (details: Section \ref{sec:experiments}, Appendix \ref{subsec:filtering_appendix}). This yields the final training datasets $\mathcal{D}_{\mathrm{pref}}$ and $\mathcal{D}_{\mathrm{aug\_filtered}}$.

\subsection{Robust Reward Model Training}
\label{subsec:training_phase}

Given the original data $\mathcal{D}_{\mathrm{pref}}$ and the filtered augmented data $\mathcal{D}_{\mathrm{aug\_filtered}}$, the final \carma{} reward model $\hat{\mathrm{R}}_\theta$ is trained by minimizing a composite loss function $\mathcal{L}(\theta)$ over the combined dataset $\mathcal{D} = \mathcal{D}_{\mathrm{pref}} \cup \mathcal{D}_{\mathrm{aug\_filtered}}$:

\begin{equation}
\mathcal{L}(\theta)=
-\underbrace{%
  \sum_{\substack{(\mathrm{Q},\mathrm{y}_w,\mathrm{y}_l)\\
                 \in\,\mathcal{D}_{\mathrm{pref}}\cup\mathcal{D}_{\mathrm{causal}}}}
  \!\log\!\bigl[\sigmoid(\Delta_{wl})\bigr]}_{\text{Preference Loss (Causal Sensitivity)}}
-\lambda\,\underbrace{%
  \sum_{\substack{(\mathrm{Q},\mathrm{A}_1,\mathrm{A}_2,\,y=\text{tie})\\
                 \in\,\mathcal{D}_{\mathrm{neutral}}}}
  \!\left(-\frac{1}{2}\bigl[\log\sigmoid(\Delta_{12})+\log\sigmoid(-\Delta_{12})\bigr]\right)}_{\text{Neutral Tie Loss (Spurious Invariance)}}
\label{eq:combined_loss_methodology}
\end{equation}

where $\Delta_{wl} = \hat{\mathrm{R}}_\theta(\mathrm{Q}, \mathrm{A}_w) - \hat{\mathrm{R}}_\theta(\mathrm{Q}, \mathrm{A}_l)$ and $\Delta_{12} = \hat{\mathrm{R}}_\theta(\mathrm{Q}, \mathrm{A}_1) - \hat{\mathrm{R}}_\theta(\mathrm{Q}, \mathrm{A}_2)$. The first term (Preference Loss) trains sensitivity to causal quality using $\mathcal{D}_{\mathrm{pref}}$ and $\mathcal{D}_{\mathrm{causal}}$. The second term (Neutral Tie Loss, weighted by $\lambda \ge 0$) trains invariance to spurious features using $\mathcal{D}_{\mathrm{neutral}}$ by encouraging $\Delta_{12} \approx 0$ for tie-labeled pairs. For our current set of experiments we keep $\lambda = 1$.

This optimization guides $\hat{\mathrm{R}}_\theta$ to be sensitive to causal attributes $\mathrm{C}$ while remaining robust to variations in spurious attributes $\mathrm{SP}$. We demonstrate \carma{}'s effectiveness in mitigating reward hacking and improving downstream policy performance in Section \ref{sec:experiments}.
\section{Theoretical Analysis}
\label{sec:theory}

We provide a theoretical analysis, detailed in Appendix~\ref{sec:theoretical_analysis_detailed}, to formalize how \carma{}'s causal augmentation isolates true reward drivers from spurious correlates. Under an idealized model, we show that training on data with targeted interventions on causal attributes enables the learned reward model to accurately identify causal reward determinants, even in the presence of numerous, unspecified spurious features.

\paragraph{Intuition and Analytical Approach}
When only a specific causal attribute is intervened to vary, and all other causal attributes are fixed to their factual versions, and spurious factors are ancestral to all causal attributes,  then the reward model is forced to learn the true impact of that causal attribute in an approximate sense.
To formalize this, we consider a setting where:

\begin{enumerate}[label=(\arabic*)]
    \item Causal attributes $\mathrm{C}(\mathrm{A})$ and spurious attributes $\mathrm{SP}(\mathrm{A})$ are modeled as boolean variables.
    \item True reward $\mathrm{R}^*$ is a sparse quadratic polynomial of $\mathrm{C}(\mathrm{A})$ only.
    \item The learned $\hat{\mathrm{R}}_\theta$ can be a denser quadratic polynomial including $\mathrm{SP}(\mathrm{A})$ and $\mathrm{C}(\mathrm{A})\mathrm{SP}(\mathrm{A})$ terms.
    \item Spurious attributes $\mathrm{SP}(\mathrm{A})$ are not descendants of causal attributes $\mathrm{C}(\mathrm{A})$.
    \item Causal augmentation is an ideal counterfactual that (given same exogenous factors leading to the answer)  intervenes one $C_i \to \neg C_i$, leaving other $C_j$ intervened to be their factual versions.
\end{enumerate}

\vspace{-0.1in}
We frame learning the coefficients  of $\mathrm{R}^*$ as an $\ell_1$-constrained linear regression (Lasso) on features derived from attribute differences between an augmented answer $A^{\mathrm{aug}}$ and its original $A$. The key insight is that the feature matrix $\mathbf{F}$ from such augmented pairs exhibits properties conducive to sparse recovery, such as low column coherence or satisfying a Restricted Isometry Property (RIP) variant. Specifcally, compared to the original training set, the augmented one has a much lower RIP.
\vspace{-0.2in}
\subsection{Main Theoretical Result (Informal)}

This structure leads to the following result (formalized as Theorem~\ref{thm:appendix_lasso_recovery} in Appendix~\ref{sec:theoretical_analysis_detailed}):

\vspace{0.03in}
\begin{takeawaybox}
\vspace{-0.1in}
\begin{theorem}[\textbf{Informal Statement}]
\label{thm:main_body_lasso_recovery}
Under the idealized model assumptions, $\ell_1$-constrained regression on $m$ causally augmented examples recovers the true causal reward coefficients  $\mathbf{a}$ with an $\ell_2$-error $\lVert \mathbf{\theta} - \hat{\mathbf{\theta}} \rVert_2$ that scales (ignoring constants and terms related to imperfect sparsity recovery) roughly as $O\left( \lVert \theta_{{\cal N}^c}\rVert_ 1 (\frac{1}{k} + \sqrt{\frac{\log(k+\ell)}{m}})\right)$ where ${\cal N}$ is the top $O(k)$ coefficients in the $R^{*}$ true reward model. This highlights a primary dependence on the number of causal attributes $k$ and samples $m$, and only a weak, logarithmic dependence on the spurious attribute dimension $\ell$.
\end{theorem}
\end{takeawaybox}

\vspace{0.03in}

\textbf{Implications:} This theorem suggests that \carma{}'s causal augmentation, by promoting favorable properties (like RIP or low incoherence) in the effective design matrix, guides the reward model towards genuine causal drivers. Further, the error vector has $\ell_2$ norm is linear in the causal dimension $k$ in the worst case and zero in the best case where $R^{*}$ has sparser dependence on the causal factors.  If it was the preference training dataset, the error could be proportional to $\lVert \theta \rVert_1$ (which is $O(k^2)$).

\vspace{-0.15in}
\section{Experiments}
\label{sec:experiments}

\begin{table}[!t]
    \centering
    \resizebox{\linewidth}{!}{%
    \renewcommand{\arraystretch}{1.3}
    \begin{tabular}{@{}llHccccHccccc@{}}
        \toprule
        & \multirow{2}{*}{\textbf{Method}} & \multicolumn{5}{c}{\textbf{PairPM}} & \multicolumn{5}{c}{\textbf{BT}} \\
        \cmidrule(lr){3-7} \cmidrule(lr){8-12}
        & & \textbf{Average} & \textbf{Chat} & \textbf{Chat-Hard} & \textbf{Safety} & \textbf{Reasoning} & \textbf{Average} & \textbf{Chat} & \textbf{Chat-Hard} & \textbf{Safety} & \textbf{Reasoning} \\
        \midrule
        \multirow{4}{*}{\rotatebox[origin=c]{90}{\small\gemmait{9}}}
        & Vanilla RM & 81.22 & \textbf{97.90} & 63.64 & 77.48 & 85.88 & 79.14 & \textbf{97.26} & 58.85 & 69.30 & 91.17 \\
        & RRM        & 82.54 & 97.12 & 71.05 & 74.70 & 87.27 & 83.46 & 97.21 & \textbf{69.15} & 73.13 & 94.35 \\
        & \textbf{\carma{}} & \textbf{87.84} & 97.54 & \textbf{72.30} & \textbf{87.14} & \textbf{94.39} & \textbf{85.46} & 96.28 & 65.83 & \textbf{84.05} & \textbf{95.70} \\
        \cmidrule(lr){2-12} 
        & $\Delta_{\text{\carma{} - RRM}}$ &
        \changeUp{+5.30} & 
        \changeUp{+0.42} &  
        \changeUp{+1.25} & 
        \changeUp{+12.44} & 
        \changeUp{+7.12} & 
        \changeUp{+2.00} &
        \changeDown{-0.93} & 
        \changeDown{-3.32} &
        \changeUp{+10.92} & 
        \changeUp{+1.35} \\ 
        \midrule
        \multirow{4}{*}{\rotatebox[origin=c]{90}{\small\qwen{}}}
        & Vanilla RM & 78.18 & \textbf{97.21} & 52.85 & 73.99 & 88.68 & 72.73 & 97.21 & 46.27 & 68.04 & 79.39 \\
        & RRM        & 82.04 & 97.21 & \textbf{64.80} & 75.27 & 90.86 & 78.20 & \textbf{98.04} & \textbf{59.65} & 72.43 & 82.66 \\
        & \textbf{\carma{}} & \textbf{83.15} & 96.37 & 61.73 & \textbf{82.23} & \textbf{92.26} & \textbf{80.81} & 96.93 & 58.66 & \textbf{78.92} & \textbf{88.71} \\
        \cmidrule(lr){2-12} 
        & $\Delta_{\text{\carma{} - RRM}}$ &
        \changeUp{+1.11} &
        \changeDown{-0.84} &
        \changeDown{-3.07} &
        \changeUp{+6.96} &
        \changeUp{+1.40} &
        \changeUp{+2.61} &
        \changeDown{-1.11} &
        \changeDown{-0.99} & 
        \changeUp{+6.49} &
        \changeUp{+6.05} \\
        \midrule 
        \multirow{4}{*}{\rotatebox[origin=c]{90}{\small\gemma{2}}} 
        & Vanilla RM & 53.75 & 92.88 & 33.33 & 42.03 & 46.74 & 65.52 & 94.27 & 38.27 & 50.20 & 79.34 \\
        & RRM        & 66.23 & \textbf{94.13} & 43.75 & 47.64 & 79.38 & 66.95 & \textbf{94.97} & 49.34 & 50.07 & 73.42 \\
        & \textbf{\carma{}} & \textbf{70.69} & 92.18 & \textbf{50.00} & \textbf{55.14} & \textbf{85.42} & \textbf{72.45} & 92.74 & \textbf{53.62} & \textbf{60.00} & \textbf{83.45} \\
        \cmidrule(lr){2-12} 
        & $\Delta_{\text{\carma{} - RRM}}$ &
        \changeUp{+4.46} &
        \changeDown{-1.95} &  
        \changeUp{+6.25} & 
        \changeUp{+7.50} & 
        \changeUp{+6.04} & 
        \changeUp{+5.50} &
        \changeDown{-2.23} & 
        \changeUp{+4.28} &
        \changeUp{+9.93} & 
        \changeUp{+10.03} \\ 
        \bottomrule
    \end{tabular}%
    }
    \caption{Performance Comparison of Pairwise Preference Model and Bradley-Terry Reward Model on RewardBench trained using various base models. See Appendix Section \ref{ssec:variance_rewardbench} for variance in results.}
    \label{tab:performance_bt_pairpm_rewardbench_extended_final} 
\end{table}

\begin{figure}[!h]
  \centering
  \begin{minipage}[t]{0.48\textwidth}
  \vspace{-0.1in}
  \paragraph{Neutrals help in spurious suppression}

Neutral augmentations significantly improve robustness compared to causal-only training (Figures~\ref{fig:rewardbench_subsets_neutral_ablations} and~\ref{fig:rewordbench-avg_neutral_ablations}). All neutral variants outperform the causal-only \carma{}-C model. Among them, \carma{}-IQN achieves the best overall performance on RewardBench, with a gain of \changeUp{+5.4\%} over the RRM baseline. Meanwhile, \carma{}-CAN achieves the best performance on reWordBench, with a gain of \changeUp{+12.5\%}.

  \end{minipage}\hfill          
  \begin{minipage}[t]{0.48\textwidth}
    \vspace{-0.15in}
    \centering
    \includegraphics[width=0.95\linewidth]{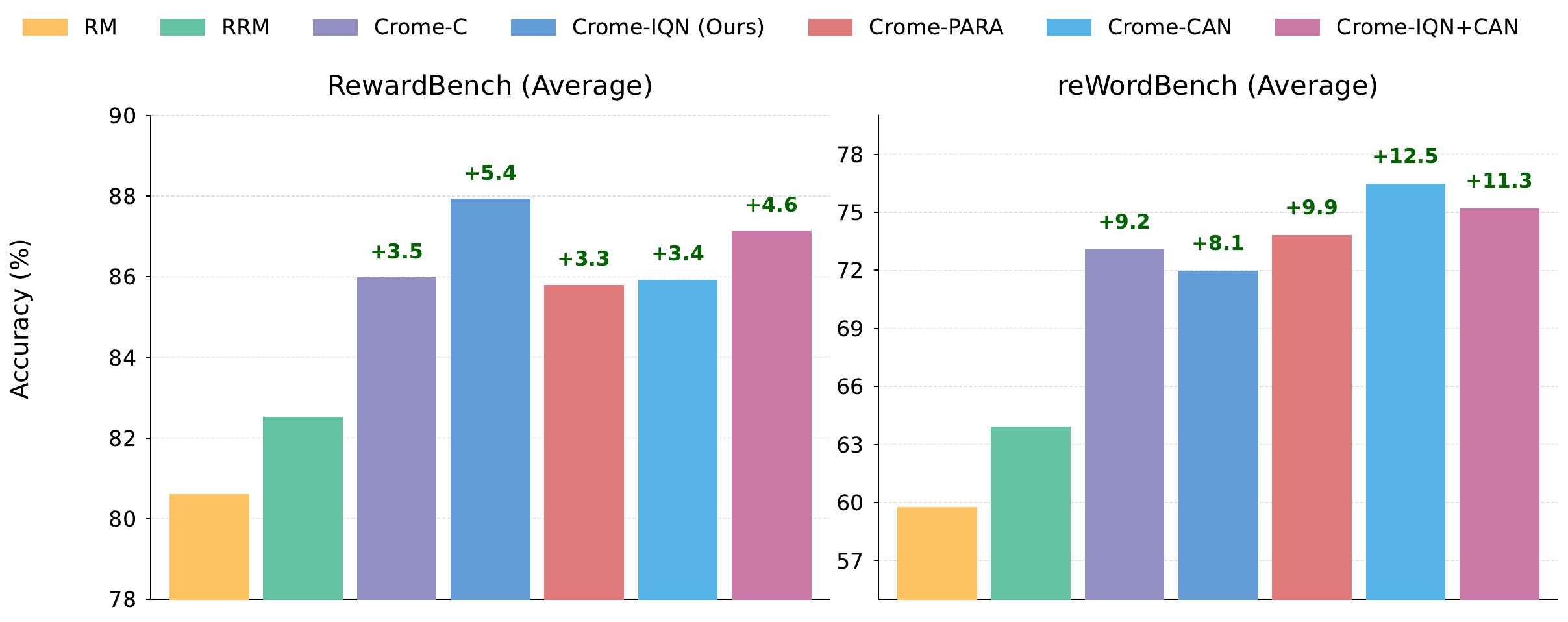}
    \vspace{-0.10in}
    \captionof{figure}{Average performance on RewardBench and reWordBench for \carma{} trained with different neutral augmentation strategies. }
    \label{fig:rewordbench-avg_neutral_ablations}
  \end{minipage}

\vspace{0.03in}
\begin{takeawaybox}
\textbf{Key Takeaway:}  \textit{Explicit} suppression of spurious correlates via neutral augmentations mitigates reward hacking by learning \textit{invariant} reward signals, thereby improving downstream performance.
\end{takeawaybox}
\vspace{0.03in}
\end{figure}

\begin{figure}[!ht]
  \centering
  \includegraphics[width=0.95\columnwidth]{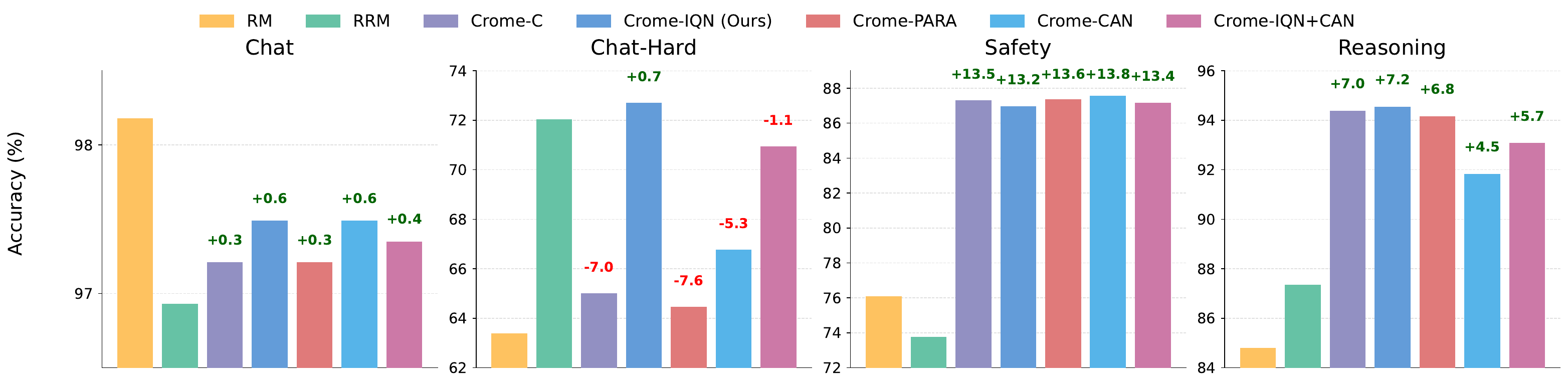}
  \caption{Evaluations of neutral augmentation variants on the different subsets of RewardBench.}
  \label{fig:rewardbench_subsets_neutral_ablations}
\end{figure}

\vspace{-0.1in}
Our experiments are designed to address the following research questions:
\label{list:research_questions}
\begin{itemize}[left=18pt,itemsep=0pt, topsep=0pt, parsep=0pt]
    \item[\textbf{RQ1:}] \textbf{RM Performance and Robustness:} How does \carma{} perform on standard preference prediction tasks and how robust is it against spurious correlations(Table \ref{tab:performance_bt_pairpm_rewardbench_extended_final}, Figure \ref{fig:reword_absolute_robustness_gemma9b_pairpm})? 

    \item[\textbf{RQ2:}] \textbf{Best-of-N Alignment:} Does the robustness achieved by \carma{} lead to favorable results in a Best-of-N setup as well, when compared to strong baselines (Figures \ref{fig:asr_reduction_gemma9b}, \ref{fig:bon_gsm8k_gemma9b}, Table \ref{tab:bon_results_rewardbench})? 
    
    \item[\textbf{RQ3:}] \textbf{Neutral Augmentations:} How effective are the different neutrals augmentation strategies in enforcing \textit{invariance} to unknown spurious correlates (Figures \ref{fig:rewordbench-avg_neutral_ablations}, \ref{fig:rewardbench_subsets_neutral_ablations})?
\end{itemize}

\subsection{Experimental Settings}
\label{subsec:experimental_settings}

\carma{} and baseline reward models (Vanilla RM, RRM \citep{liu2024rrm}) are trained on the UltraFeedback dataset \citep{cui2023ultrafeedback}, with counterfactuals generated using Gemini 2.0 Flash.
We evaluate performance on RewardBench \citep{lambert2024rewardbench} and robustness on reWordBench \citep{wu2025rewordbench} \footnote{Since reWordBench has not been released, we follow the paper and communicated with the authors to reproduce it, see Appendix Section \ref{app:rewordbench_creation}}. Experiments utilize diverse base LLMs (\gemmait{9}, \qwen{}, \gemma{2}) for both Pairwise Preference (PairPM) and Bradley-Terry (BT) reward models. Downstream alignment impact is assessed via Best-of-N selection on tasks including RewardBench, GSM8K, and WildGuardTest.
Comprehensive details on datasets, model specifics, augmentation procedures, filtering, training hyperparameters, and all experimental configurations are provided in Appendix \ref{sec:experimental_details}.
\subsection{Experimental Results addressing Research Questions (RQ1-3):}
\label{subsec:experimental_results}

\begin{figure}[ht]
  \centering
  \includegraphics[width=1.0\columnwidth]{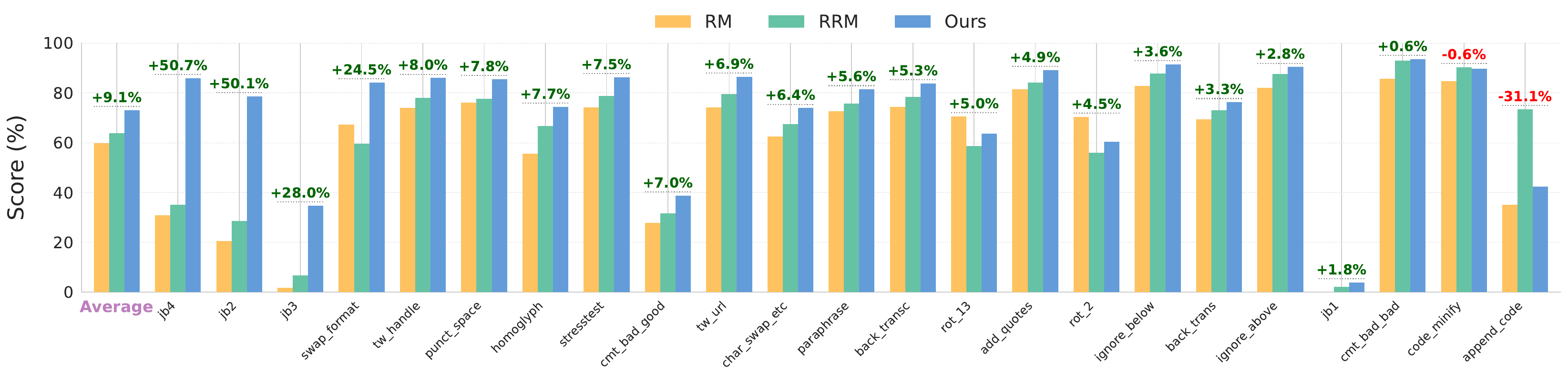}
  \caption{\textbf{Robustness of \carma{}} on reWordBench. Comparing RM, RRM and \carma{} by measuring ranking accuracy on a diverse set of meaning preserving transformations in reWordBench. Various transformations such as paraphrasing, addition of irrelevant text or code, comments etc, test the sensitivity of models to spuriousness. Robust training of \carma{} leads to robustness to spuriousness and increased sensitivity to causal attributes.
  }
  \label{fig:reword_absolute_robustness_gemma9b_pairpm}
\vspace{0.2in}  
\end{figure}

On \textbf{RewardBench} (Table~\ref{tab:performance_bt_pairpm_rewardbench_extended_final}), \carma{} consistently improves ranking accuracy over RRM across diverse base models and reward modeling techniques (PairPM, BT).
These improvements are particularly notable on the challenging \textit{Safety} (up to \changeUp{13.18\%}) and \textit{Reasoning} (up to \changeUp{7.19\%}). \carma{} also demonstrates superior robustness on \textbf{reWordBench}, which tests for robustness of RMs against meaning-preserving transformations (Figure~\ref{fig:reword_absolute_robustness_gemma9b_pairpm}). 

These results show \carma{}'s robustness to inputs having spurious punctations, paraphrasing, irrelevant text, code or comments as tested by various reWordBench transformations. With \gemmait{9}, \carma{} in the PairPM setting shows an aggregate accuracy gain of up to \changeUp{9.1\%} and is superior on \textcolor{PaperGreen}{(21/23)} transformations.

\vspace{0.25in}
\begin{takeawaybox}
\textbf{Key Takeaway:} \carma{} improves RM performance on standard benchmarks while significantly improving performance and mitigating ranking accuracy drops on diverse transformed inputs, \textit{without ever being explicitly trained on such spurious transformations}.
\end{takeawaybox}
\vspace{0.1in}

\clearpage
\paragraph{BoN for Robust LLM Alignment Across Chat, Reasoning, and Safety} Following the method used by \citet{wu2025rewordbench}, we perform best-of-n selection using \carma\ across RewardBench categories, which consists of datasets such as AlpacaEval. Across all values of $N$, \carma\ provided significant improvements over baselines in a head-to-head comparison.

\vspace{0.05in}
\begin{takeawaybox}
\textbf{Key Takeaway:} \carma's emphasis on causal attributes enhances its discriminative power in Best-of-N selection, leading to more consistent identification of superior responses.
\end{takeawaybox}
\vspace{0.05in}

\begin{table}[ht]
    \centering
    \resizebox{0.58\linewidth}{!}{%
    \begin{tabular}{@{} c
            S[table-format=2.2] S[table-format=2.2] S[table-format=2.2]
            S[table-format=2.2] S[table-format=2.2] S[table-format=2.2] @{}}
            \toprule
            \multirow{2}{*}{\textbf{N}} &
            \multicolumn{3}{c}{\textbf{\carma{} vs RM}} &
            \multicolumn{3}{c}{\textbf{\carma{} vs RRM}} \\
            \cmidrule(lr){2-4}\cmidrule(lr){5-7}
            & {\carma{}} & {RM} & {Ties} & {\carma{}} & {RRM} & {Ties} \\
            \midrule
            4  & \textbf{28.08} & 13.85 & 58.07 & \textbf{28.03} & 14.13 & 57.84 \\
            8  & \textbf{34.32} & 17.24 & 48.43 & \textbf{34.36} & 17.19 & 48.45 \\
            16 & \textbf{39.93} & 20.54 & 39.53 & \textbf{41.14} & 20.40 & 38.46 \\
            32 & \textbf{44.79} & 21.88 & 33.33 & \textbf{45.46} & 22.01 & 32.53 \\
            \bottomrule
    
    \end{tabular}
    }
    \caption{\textbf{Win rates for \carma{} compared with RM and RRM on RewardBench}. We follow \citet{wu2025rewordbench} and take all 2985 prompts from RewardBench and get BoN responses from a \gemmait{9} model using \carma{}, RM or RRM as the reward models. Following this, we separately compare responses generated by \carma{} with RM and RRM, using GPT-4 as a judge.\vspace{-0.1in}}
    \label{tab:bon_results_rewardbench}
\end{table}

\begin{figure}[ht]
    \centering
    \includegraphics[width=0.7\linewidth]{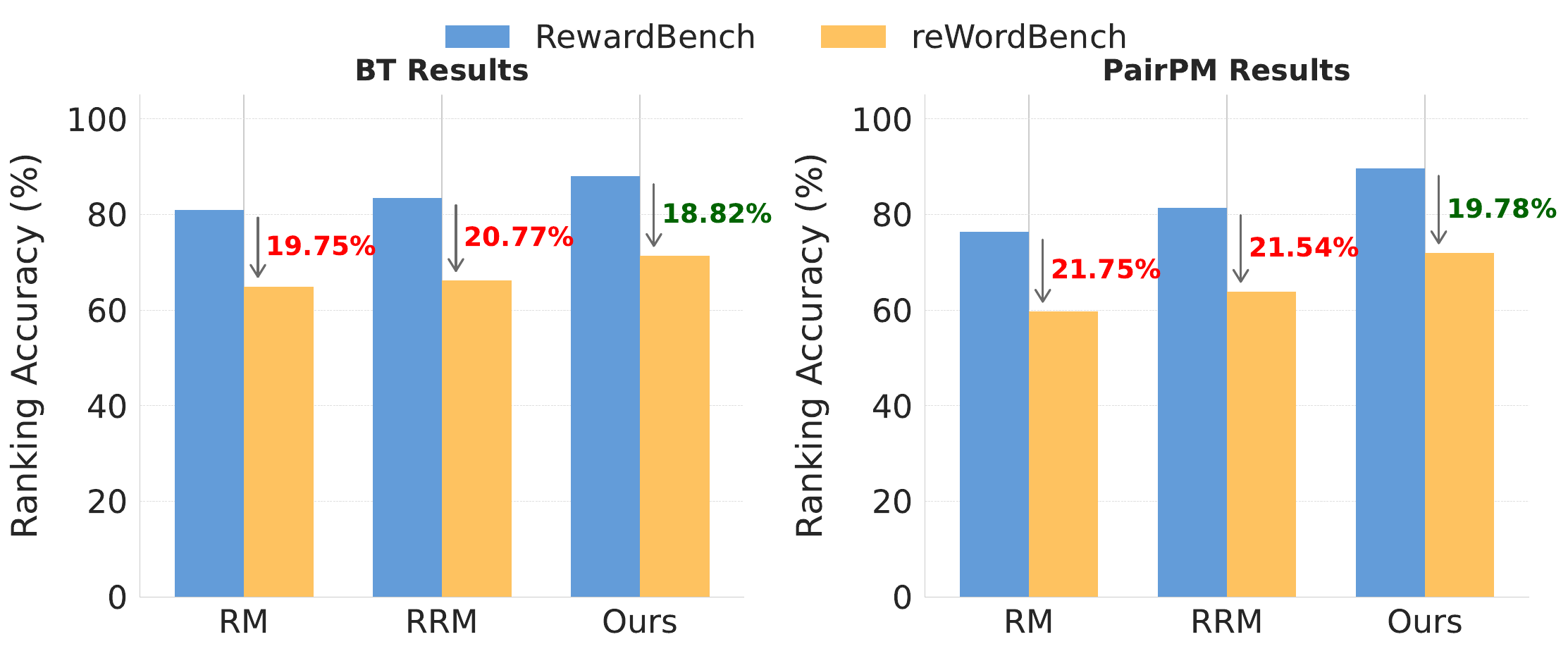}
    \caption{\textbf{Percentage improvement in ranking accuracy} between RewardBench and reWordBench. Here we show the average ranking accuracy across reWordBench transformations of \carma{} and baselines on reWordBench and RewardBench as done in \citet{wu2025rewordbench}, as well as the percentage drop in ranking accuracy on reWordBench compared to RewardBench. We show that \carma{}'s ranking accuracy percentage drop going from RewardBench to reWordBench is the lowest compared to baselines.\vspace{-0.1in}}
    \label{fig:reword_ranking_acc_drop_gemma9b_bt_pairpm}
\end{figure}

\paragraph{Ranking Accuracy Percentage Improvements:}
We measure the percentage drop in response ranking accuracy between RewardBench and reWordBench scores (following the macro-avg metric used in \citet{wu2025rewordbench}). \carma{} exhibits a smaller ranking accuracy percentage drop from RewardBench to reWordBench (In case of PairPM: $19.78$\% vs. RRM's $21.54$\%. See  Figure \ref{fig:reword_ranking_acc_drop_gemma9b_bt_pairpm} for the results on BT and PairPM settings.

\vspace{0.02in}
\begin{takeawaybox}
\textbf{Key Takeaway:} Assuming sufficient concentration of spurious elements in the prompt as well as the $N$ responses, \carma{} is better at selecting the best response based on causal attributes only. For e.g., in safety, harmful prompts and responses may be spuriously disguised as benign. 
\end{takeawaybox}

\clearpage

\begin{figure}[!h]
  \centering
  \begin{minipage}[t]{0.45\textwidth}
  \paragraph{Causal Attributes help in detecting jailbreaks}
For \gemmait{9} as the solution generation model, BoN with \carma{} shows significant improvements on safety as measured on WildGuardTest. In particular the attack success ratio (ASR) on harmful prompts is much lower compared to models aligned with RM and RRM and this gap increases with N. This improved ASR comes at at a similar refusal-to-answer rate on benign prompts.

  \end{minipage}\hfill          
  \begin{minipage}[t]{0.50\textwidth}
    \vspace{-0.15in}
    \centering
    \includegraphics[width=0.95\linewidth]{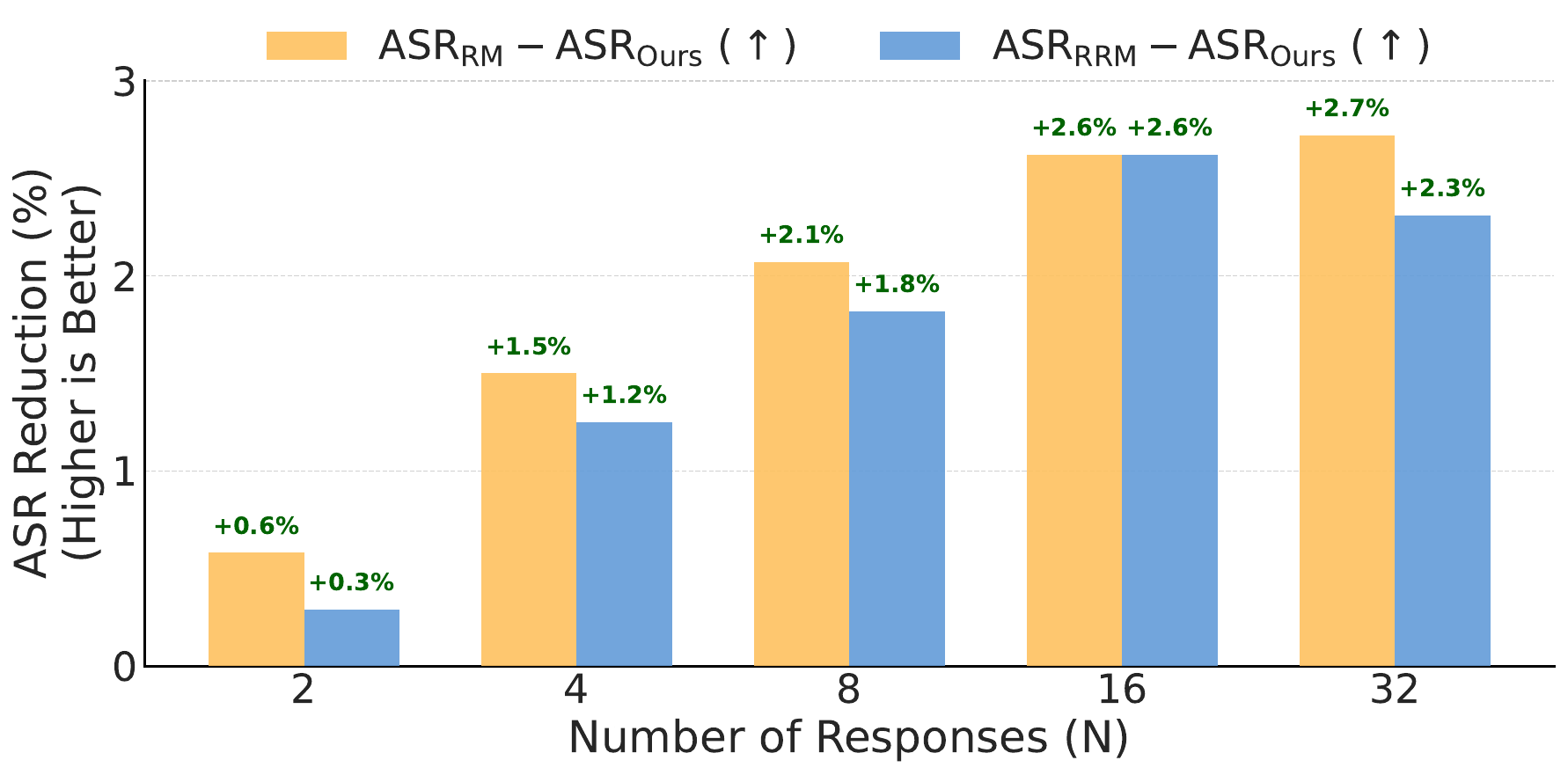}
    \vspace{-0.10in}
    \captionof{figure}{Best-of-N results: ASR reduction on WildGuardTest. }
    \label{fig:asr_reduction_gemma9b}
  \end{minipage}

\vspace{0.03in}
\begin{takeawaybox}
\textbf{Key Takeaway:} \carma{}'s causal augmentations achieve a superior trade-off between safety and over-refusals, because its contrastive pairs delineate the decision boundary for harmful content more faithfully. This leads to safer content, while avoiding excessive refusals on benign prompts.

\end{takeawaybox}
\vspace{0.03in}
\end{figure}

\vspace{0.03in}
\begin{figure}[!ht]
  \centering
  \begin{minipage}[t]{0.48\textwidth}
  \vspace{0.0in}
    \centering
    \includegraphics[width=0.9\linewidth]{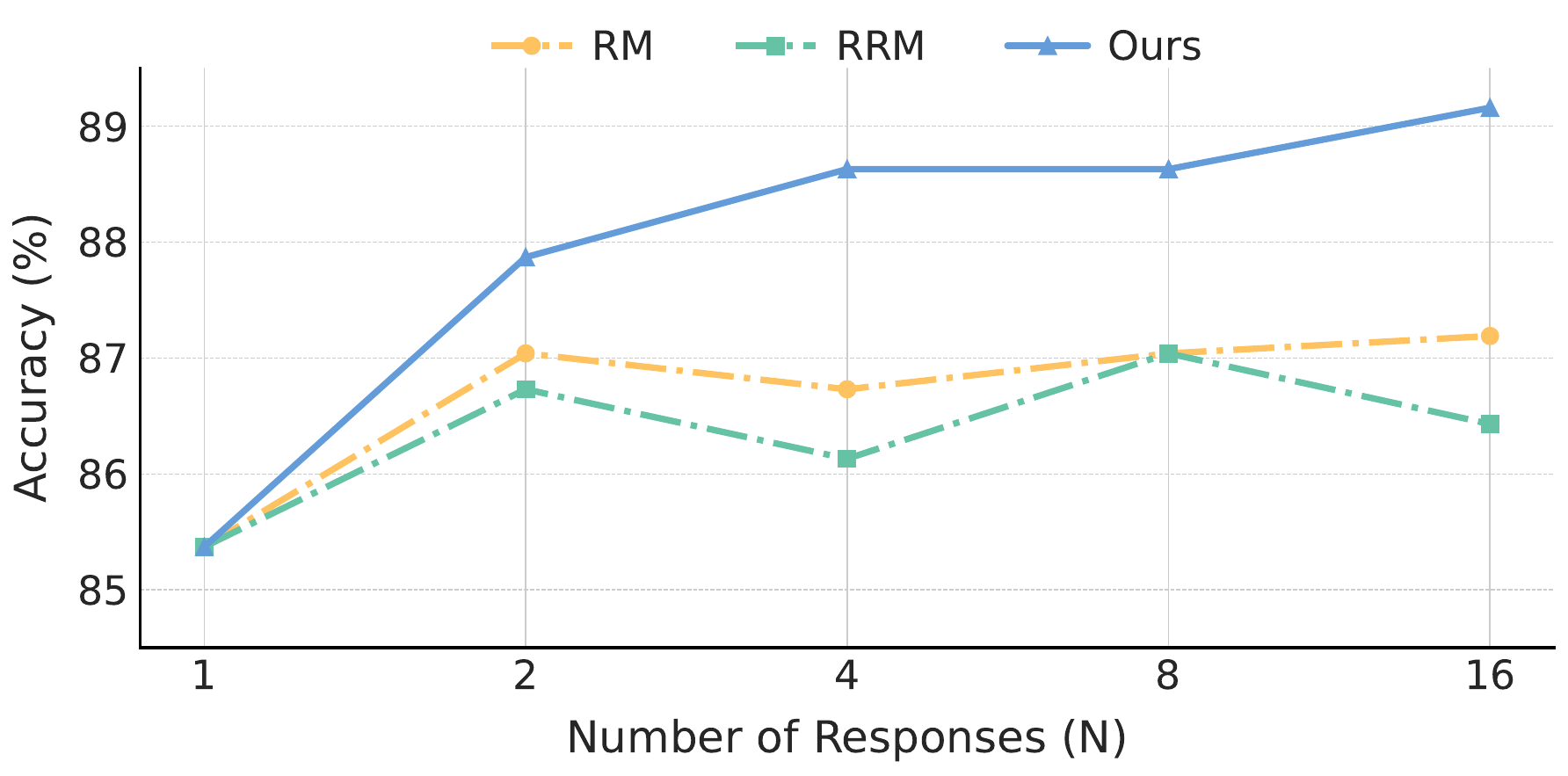}
    \captionof{figure}{Best-of-N Reasoning evaluation on GSM8K.}
    \label{fig:bon_gsm8k_gemma9b}
  \end{minipage}\hfill          
  \begin{minipage}[t]{0.48\textwidth}
  \vspace{0.0in}
  \paragraph{Disentangling Content related features from stylistic (spurious) ones helps in reasoning}

For \gemmait{9} as the solution generation model on GSM8K, \carma{} shows a consistent gap over baseslines across different values of $N$.
Non robust reward models may focus on stylistic details. Good looking, detailed but wrong reasoning steps may mis-guide non-robust RMs into giving a higher score to the response.
\end{minipage}

\vspace{0.03in}
\begin{takeawaybox}
\textbf{Key Takeaway:} Reasoning correctness is dependent on focusing on correctness over stylistic features. Our training ensures \carma{} is good at capturing content-features over other attributes. 
\end{takeawaybox}
\vspace{0.03in}
\end{figure}

\vspace{-0.22in}
\subsection{Neutral Ablations}
\label{ssec:neutral_ablations}

\vspace{-0.1in}
Along with IQN, we tested several methods for enforcing spurious invariance:

\vspace{-0.15in}
\paragraph{Causally Aligned Neutrals (CAN).} 
Given a preference pair $(A_w, A_\ell)$ where $(A_w \succ A_\ell)$, 
we rewrite $A_\ell$ into $\tilde{A}_\ell$ such that the causal content of $\tilde{A}_\ell$ aligns with $A_w$ ($C(A_w) \approx C(\tilde{A}_l)$), but due to the rewrite from $A_\ell$, the spurious attributes of $A_\ell$ remain. By assigning a tie-label to this pair during training, we force the model to learn invariance to the spurious differences. While this method is sound theoretically, the approximation of $C(A_w)$ by $C(\tilde{A}_l)$ is not perfect. Furthermore, some spurious attributes $SP'(\tilde{A}_l) \subset SP(\tilde{A}_l)$ vary when we move causal attributes. Invariance to these attributes $SP'(\tilde{A}_l)$ is not captured by CAN.

\vspace{-0.15in}
\paragraph{Paraphrase Neutral (PARA).} Given an answer $A$ to a query $Q$, we rewrite $A$ to an approximate $\tilde{A}$ using an LLM, such that spurious features vary, but causal features do not. Unlike CAN which provides structured rewrites, PARA is a simpler method for rewriting equivalent answers (neutrals). This idea is common in literature (For example, see \citet{wu2025rewordbench}). 
Yet the central issue here is that $C(\tilde{A})$ may inadvertently vary during a rewrite (due to the $SP\to C$ causation in Fig \ref{fig:causal_graph}). Furthermore, the SP variations introduced through paraphrasing  are not reflective of the complex downstream distributions.

\vspace{-0.15in}
\paragraph{Other Combinations.} We provide two more variations for completeness -- (i) causal only augmentations, with no neutrals (C) (ii) Both IQN and CAN neutrals sampled equally (IQN+CAN).

\clearpage
\vspace{0.1in}

The \carma{} variants include: \carma{}-C (only causals), \carma{}-IQN (causals + irrelevant query neutrals), \carma{}-PARA
  (causals + paraphrased neutrals), \carma{}-CAN (causals + causally-aligned neutrals), and  \carma{}-IQN+CAN
  (causals + irrelevant query neutrals + 
  causally-aligned neutrals). On the especially challenging \textit{Chat-Hard} subset, \carma{}-IQN performs best. See Appendix Section \ref{sec:causal_model_details} for more details.  Prompts for obtaining these neutrals is given in Appendix \ref{sec:prompt_templates}.
\vspace{0.03in}
\begin{takeawaybox}
\textbf{Key Takeaway:}  A combination of well-designed augmentation strategies, e.g. causal upgradations and degradations, along with IQN produces the most robust and generalizable reward models.
\end{takeawaybox}

\vspace{-0.1in}
\paragraph{Discussion on Neutrals:} 
Our Figure \ref{fig:causal_graph} suggests that interventions along spurious attributes can confound causal attributes in myriad ways. Firstly, there could be causal attributes, which upon intervention can lead to spurious attribute change ($CA\to SP$). Secondly, if spurious attributes change, this can lead to a change in Causal Attributes ($SP\to CA$). Due to such confounding factors, an intervention free solution, such as IQN, turns out to be a clever way to provide invariance to spuriousness.
IQN provides invariance to those spurious factors that change with causal changes (See Fig. \ref{fig:carma_augmentation_visual_overview}), as well as natural spurious variations when irrelevant questions are paired with answers corresponding to a different question.

\vspace{-0.1in}
\paragraph{Ablations and Additional Results:} See Appendix Section \ref{sec:additional_results} where we show that \carma{} exhibits stable and significant improvements in robustness with low variance across different training runs. We also show that using open-weights models as the oracle LLM, such as \gemmathreeit{27}, \carma{} exhibits significant improvements in robustness. Additionally, we also show performance of \carma{} and baselines on in-distribution and out-of-distribution examples, showing superior effective robustness achieved by \carma{}.

\newpage

\begin{figure}[!h]
  \centering
  \includegraphics[width=1.0\columnwidth]{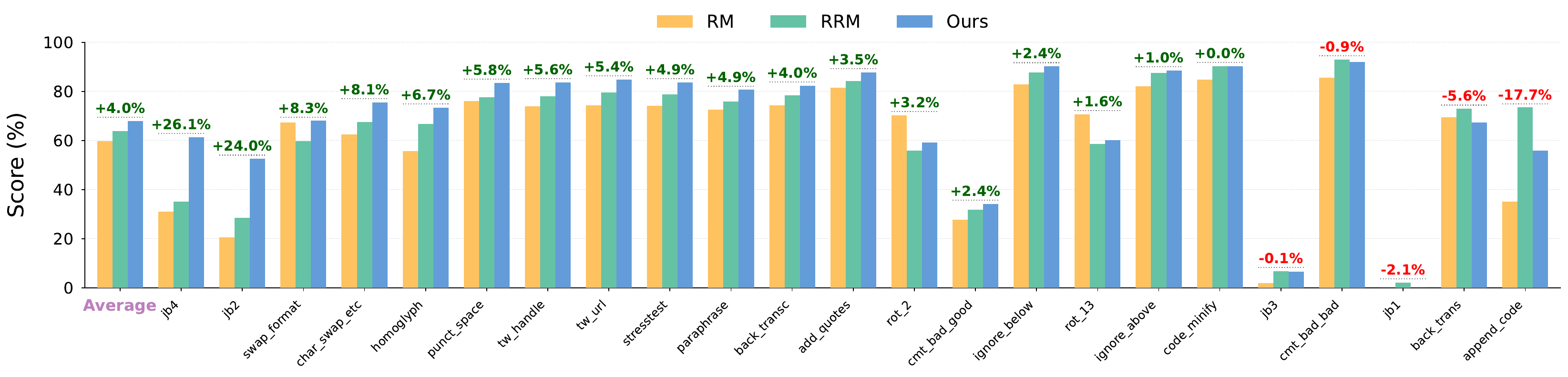}
  \caption{\textbf{Robustness with \gemmathreeit{27} as oracle LLM} Comparing of RM, RRM and \carma{} on reWordBench. Here all reward models are \gemmait{9} based, in the PairPM setting.\vspace{-0.1in}
  }
  \label{fig:reword_absolute_robustness_gemma9b_pairpm_gemma27Boracle_main}
\end{figure}

\vspace{-0.2in}
\paragraph{Robustness to Oracle LLM Choice} To test our robustness to the choice of oracle LLM, we provide experimental results using \gemmathreeit{27} to perform attribute extraction and augmentations following which we train \carma{} on the augmented data. Table \ref{tab:performance_pairpm_gemma9b_rewardbench_gemma27boracle} shows that \carma\ outperforms the baselines by up to 2.5\% on RewardBench and 3.2\% on reWordBench. 
In Figure \ref{fig:reword_absolute_robustness_gemma9b_pairpm_gemma27Boracle_main},
our results indicate an improvement in 18/23 transformations of reWordBench.
This shows that our method is performant even with a weaker oracle LLM. This potentially indicates that the strength of \carma\ lies in its causal method, and goes beyond simply leveraging the knowledge of the oracle model.

\begin{table}[!t]
    \centering
    \resizebox{0.8\linewidth}{!}{%
    \renewcommand{\arraystretch}{1.3}
    \begin{tabular}{@{}lcccccc@{}} 
        \toprule
        
        \multirow{2}{*}{\textbf{Method}} & \multicolumn{1}{c}{\textbf{reWordBench}} & \multicolumn{5}{c}{\textbf{RewardBench}} \\ 
        \cmidrule(lr){2-2} \cmidrule(lr){3-7} 
        & \textbf{Average} & \textbf{Average} & \textbf{Chat} & \textbf{Chat-Hard} & \textbf{Safety} & \textbf{Reasoning} \\ 
        \midrule
        Vanilla RM & 59.97 & 80.61 & \textbf{98.18} & 63.38 & 76.08 & 84.80 \\ 
        RRM        & 64.68 & 82.53 & 96.93 & \textbf{72.04} & 73.78 & 87.36 \\ 
        \textbf{\carma{}} & \textbf{67.90} & \textbf{85.15} & 97.21 & 68.75 & \textbf{83.51} & \textbf{91.13} \\ 
        \bottomrule
    \end{tabular}%
    }
    \caption{\textbf{RM Performance with \gemmathreeit{27} as oracle}. Results on RewardBench and REwardBench with \gemmait{9} as base model and \gemmathreeit{27} as oracle LLM used for attribute extraction and counterfactual augmentations. Results are in PairPM setting. \vspace{-0.1in}
    }
    \label{tab:performance_pairpm_gemma9b_rewardbench_gemma27boracle}
\end{table}

\vspace{-0.1in}
\section{Discussion, Conclusion and Future Work}

\vspace{-0.1in}
In this paper, we proposed \carma, a causal framework to mitigate reward hacking during the training of reward models. \carma\ systematically disentangles causal from spurious attributes through two targeted synthetic data augmentation strategies: (1) Causal Augmentations to enforce sensitivity to genuine quality drivers, and (2) Neutral Augmentations to enforce invariance to spurious features. Notably, \carma{} does not assume access to types of spurious attributes that might effect RMs. 
Across multiple base models and reward modeling techniques (PairPM, BT), \carma{} consistently outperforms strong baselines on the RewardBench benchmark. Furthermore, \carma{} shows superior robustness on the reWordBench benchmark, which specifically tests for vulnerabilities to spurious correlations. 
We also achieve consistent improvements in downstream Best-of-N setups. 

\vspace{-0.12in}
\paragraph{Future Work.}
Our training method, centered on dataset curation, paves the way for new research directions in synthetic data research. 
A compelling application is in synthetic data generation for base model training, where the use and verification of causal attributes could prove particularly fruitful.
\section{Acknowledgments}
We thank Prateek Jain, Praneeth Nethrapalli, Rishi Saket, Partha Talukdar, Vihari Piratla, Darshan Singh S and Saisuresh Krishnakumaran for providing feedback on this work. The authors would like to thank Manish Gupta for his valuable support and guidance.

\bibliography{References}

\begin{thebibliography}{67}
\providecommand{\natexlab}[1]{#1}
\providecommand{\url}[1]{\texttt{#1}}
\expandafter\ifx\csname urlstyle\endcsname\relax
  \providecommand{\doi}[1]{doi: #1}\else
  \providecommand{\doi}{doi: \begingroup \urlstyle{rm}\Url}\fi

\bibitem[Arjovsky et~al.(2019)Arjovsky, Bottou, Gulrajani, and Lopez-Paz]{arjovsky2019invariant}
M.~Arjovsky, L.~Bottou, I.~Gulrajani, and D.~Lopez-Paz.
\newblock Invariant risk minimization.
\newblock \emph{arXiv preprint arXiv:1907.02893}, 2019.

\bibitem[Askell et~al.(2021)Askell, Bai, Chen, Drain, Ganguli, Henighan, Jones, Joseph, Mann, DasSarma, et~al.]{askell2021general}
A.~Askell, Y.~Bai, A.~Chen, D.~Drain, D.~Ganguli, T.~Henighan, A.~Jones, N.~Joseph, B.~Mann, N.~DasSarma, et~al.
\newblock A general language assistant as a laboratory for alignment.
\newblock \emph{arXiv preprint arXiv:2112.00861}, 2021.

\bibitem[Azar et~al.(2024)Azar, Guo, Piot, Munos, Rowland, Valko, and Calandriello]{azar2024general}
M.~G. Azar, Z.~D. Guo, B.~Piot, R.~Munos, M.~Rowland, M.~Valko, and D.~Calandriello.
\newblock A general theoretical paradigm to understand learning from human preferences.
\newblock In \emph{International Conference on Artificial Intelligence and Statistics}, pages 4447--4455. PMLR, 2024.

\bibitem[Bai et~al.(2022{\natexlab{a}})Bai, Jones, Ndousse, Askell, Chen, DasSarma, Drain, Fort, Ganguli, Henighan, et~al.]{bai2022training}
Y.~Bai, A.~Jones, K.~Ndousse, A.~Askell, A.~Chen, N.~DasSarma, D.~Drain, S.~Fort, D.~Ganguli, T.~Henighan, et~al.
\newblock Training a helpful and harmless assistant with reinforcement learning from human feedback.
\newblock \emph{arXiv preprint arXiv:2204.05862}, 2022{\natexlab{a}}.

\bibitem[Bai et~al.(2022{\natexlab{b}})Bai, Kadavath, Kundu, Askell, Kernion, Jones, Chen, Goldie, Mirhoseini, McKinnon, et~al.]{bai2022constitutional}
Y.~Bai, S.~Kadavath, S.~Kundu, A.~Askell, J.~Kernion, A.~Jones, A.~Chen, A.~Goldie, A.~Mirhoseini, C.~McKinnon, et~al.
\newblock Constitutional ai: Harmlessness from ai feedback.
\newblock \emph{arXiv preprint arXiv:2212.08073}, 2022{\natexlab{b}}.

\bibitem[Bradley and Terry(1952)]{bradley1952rank}
R.~A. Bradley and M.~E. Terry.
\newblock Rank analysis of incomplete block designs: I. the method of paired comparisons.
\newblock \emph{Biometrika}, 39\penalty0 (3/4):\penalty0 324--345, 1952.

\bibitem[Casper et~al.(2023)Casper, Davies, Shi, Gilbert, Scheurer, Rando, Freedman, Korbak, Lindner, Freire, et~al.]{casper2023open}
S.~Casper, X.~Davies, C.~Shi, T.~K. Gilbert, J.~Scheurer, J.~Rando, R.~Freedman, T.~Korbak, D.~Lindner, P.~Freire, et~al.
\newblock Open problems and fundamental limitations of reinforcement learning from human feedback.
\newblock \emph{arXiv preprint arXiv:2307.15217}, 2023.

\bibitem[Chen et~al.(2024)Chen, Zhu, Soselia, Chen, Zhou, Goldstein, Huang, Shoeybi, and Catanzaro]{chen2024odin}
L.~Chen, C.~Zhu, D.~Soselia, J.~Chen, T.~Zhou, T.~Goldstein, H.~Huang, M.~Shoeybi, and B.~Catanzaro.
\newblock Odin: Disentangled reward mitigates hacking in rlhf.
\newblock \emph{arXiv preprint arXiv:2402.07319}, 2024.

\bibitem[Chi et~al.(2024)Chi, Li, Yang, Liu, Lan, Ren, Liu, and Han]{chi2024unveiling}
H.~Chi, H.~Li, W.~Yang, F.~Liu, L.~Lan, X.~Ren, T.~Liu, and B.~Han.
\newblock Unveiling causal reasoning in large language models: Reality or mirage?
\newblock \emph{Advances in Neural Information Processing Systems}, 37:\penalty0 96640--96670, 2024.

\bibitem[Christiano et~al.(2017)Christiano, Leike, Brown, Martic, Legg, and Amodei]{christiano2017deep}
P.~F. Christiano, J.~Leike, T.~Brown, M.~Martic, S.~Legg, and D.~Amodei.
\newblock Deep reinforcement learning from human preferences.
\newblock \emph{Advances in neural information processing systems}, 30, 2017.

\bibitem[Cobbe et~al.(2021)Cobbe, Kosaraju, Bavarian, Chen, Jun, Kaiser, Plappert, Tworek, Hilton, Nakano, Hesse, and Schulman]{cobbe2021gsm8k}
K.~Cobbe, V.~Kosaraju, M.~Bavarian, M.~Chen, H.~Jun, L.~Kaiser, M.~Plappert, J.~Tworek, J.~Hilton, R.~Nakano, C.~Hesse, and J.~Schulman.
\newblock Training verifiers to solve math word problems.
\newblock \emph{arXiv preprint arXiv:2110.14168}, 2021.

\bibitem[Coste et~al.(2023)Coste, Anwar, Kirk, and Krueger]{coste2023reward}
T.~Coste, U.~Anwar, R.~Kirk, and D.~Krueger.
\newblock Reward model ensembles help mitigate overoptimization.
\newblock \emph{arXiv preprint arXiv:2310.02743}, 2023.

\bibitem[Cui et~al.(2023)Cui, Yuan, Ding, Yao, Zhu, Ni, Xie, Liu, and Sun]{cui2023ultrafeedback}
G.~Cui, L.~Yuan, N.~Ding, G.~Yao, W.~Zhu, Y.~Ni, G.~Xie, Z.~Liu, and M.~Sun.
\newblock Ultrafeedback: Boosting language models with high-quality feedback.
\newblock \emph{arXiv preprint arXiv:2310.01377}, 2023.

\bibitem[Denison et~al.(2024)Denison, MacDiarmid, Barez, Duvenaud, Kravec, Marks, Schiefer, Soklaski, Tamkin, Kaplan, et~al.]{denison2024sycophancy}
C.~Denison, M.~MacDiarmid, F.~Barez, D.~Duvenaud, S.~Kravec, S.~Marks, N.~Schiefer, R.~Soklaski, A.~Tamkin, J.~Kaplan, et~al.
\newblock Sycophancy to subterfuge: Investigating reward-tampering in large language models.
\newblock \emph{arXiv preprint arXiv:2406.10162}, 2024.

\bibitem[Eisenstein et~al.(2023)Eisenstein, Nagpal, Agarwal, Beirami, D'Amour, Dvijotham, Fisch, Heller, Pfohl, Ramachandran, et~al.]{eisenstein2023helping}
J.~Eisenstein, C.~Nagpal, A.~Agarwal, A.~Beirami, A.~D'Amour, D.~Dvijotham, A.~Fisch, K.~Heller, S.~Pfohl, D.~Ramachandran, et~al.
\newblock Helping or herding? reward model ensembles mitigate but do not eliminate reward hacking.
\newblock \emph{arXiv preprint arXiv:2312.09244}, 2023.

\bibitem[Ethayarajh et~al.(2024)Ethayarajh, Xu, Muennighoff, Jurafsky, and Kiela]{ethayarajh2024kto}
K.~Ethayarajh, W.~Xu, N.~Muennighoff, D.~Jurafsky, and D.~Kiela.
\newblock Kto: Model alignment as prospect theoretic optimization.
\newblock \emph{arXiv preprint arXiv:2402.01306}, 2024.

\bibitem[Feder et~al.(2021)Feder, Oved, Shalit, and Reichart]{feder2021causalm}
A.~Feder, N.~Oved, U.~Shalit, and R.~Reichart.
\newblock Causalm: Causal model explanation through counterfactual language models.
\newblock \emph{Computational Linguistics}, 47\penalty0 (2):\penalty0 333--386, 2021.

\bibitem[Feder et~al.(2022)Feder, Keith, Manzoor, Pryzant, Sridhar, Wood-Doughty, Eisenstein, Grimmer, Reichart, Roberts, et~al.]{feder2022causal}
A.~Feder, K.~A. Keith, E.~Manzoor, R.~Pryzant, D.~Sridhar, Z.~Wood-Doughty, J.~Eisenstein, J.~Grimmer, R.~Reichart, M.~E. Roberts, et~al.
\newblock Causal inference in natural language processing: Estimation, prediction, interpretation and beyond.
\newblock \emph{Transactions of the Association for Computational Linguistics}, 10:\penalty0 1138--1158, 2022.

\bibitem[Gao et~al.(2023)Gao, Schulman, and Hilton]{gao2023scaling}
L.~Gao, J.~Schulman, and J.~Hilton.
\newblock Scaling laws for reward model overoptimization.
\newblock In \emph{International Conference on Machine Learning}, pages 10835--10866. PMLR, 2023.

\bibitem[Gupta et~al.(2025)Gupta, Shandilya, Zhang, Madhavan, Ghosh, Bansal, Yao, and Rajmohan]{gupta2025carmodynamiccriteriageneration}
T.~Gupta, S.~Shandilya, X.~Zhang, R.~Madhavan, S.~Ghosh, C.~Bansal, H.~Yao, and S.~Rajmohan.
\newblock Carmo: Dynamic criteria generation for context-aware reward modelling, 2025.
\newblock URL \url{https://arxiv.org/abs/2410.21545}.

\bibitem[Han et~al.(2024)Han, Rao, Ettinger, Jiang, Lin, Lambert, Choi, and Dziri]{wildguard2024}
S.~Han, K.~Rao, A.~Ettinger, L.~Jiang, B.~Y. Lin, N.~Lambert, Y.~Choi, and N.~Dziri.
\newblock Wildguard: Open one-stop moderation tools for safety risks, jailbreaks, and refusals of llms, 2024.
\newblock URL \url{https://arxiv.org/abs/2406.18495}.

\bibitem[Hong et~al.(2024)Hong, Lee, and Thorne]{hong2024orpo}
J.~Hong, N.~Lee, and J.~Thorne.
\newblock Orpo: Monolithic preference optimization without reference model.
\newblock In \emph{Proceedings of the 2024 Conference on Empirical Methods in Natural Language Processing}, pages 11170--11189, 2024.

\bibitem[Kaushik et~al.(2019)Kaushik, Hovy, and Lipton]{kaushik2019learning}
D.~Kaushik, E.~Hovy, and Z.~C. Lipton.
\newblock Learning the difference that makes a difference with counterfactually-augmented data.
\newblock \emph{arXiv preprint arXiv:1909.12434}, 2019.

\bibitem[Khalifa et~al.(2025)Khalifa, Agarwal, Logeswaran, Kim, Peng, Lee, Lee, and Wang]{khalifa2025process}
M.~Khalifa, R.~Agarwal, L.~Logeswaran, J.~Kim, H.~Peng, M.~Lee, H.~Lee, and L.~Wang.
\newblock Process reward models that think.
\newblock \emph{arXiv preprint arXiv:2504.16828}, 2025.

\bibitem[Kiciman et~al.(2023)Kiciman, Ness, Sharma, and Tan]{kiciman2023causal}
E.~Kiciman, R.~Ness, A.~Sharma, and C.~Tan.
\newblock Causal reasoning and large language models: Opening a new frontier for causality.
\newblock \emph{Transactions on Machine Learning Research}, 2023.

\bibitem[Lambert et~al.(2024)Lambert, Pyatkin, Morrison, Miranda, Lin, Chandu, Dziri, Kumar, Zick, Choi, et~al.]{lambert2024rewardbench}
N.~Lambert, V.~Pyatkin, J.~Morrison, L.~Miranda, B.~Y. Lin, K.~Chandu, N.~Dziri, S.~Kumar, T.~Zick, Y.~Choi, et~al.
\newblock Rewardbench: Evaluating reward models for language modeling.
\newblock \emph{arXiv preprint arXiv:2403.13787}, 2024.

\bibitem[Li and Li(2024)]{li2024process}
W.~Li and Y.~Li.
\newblock Process reward model with q-value rankings.
\newblock \emph{arXiv preprint arXiv:2410.11287}, 2024.

\bibitem[Liu et~al.(2024)Liu, Xiong, Ren, Chen, Wu, Joshi, Gao, Shen, Qin, Yu, et~al.]{liu2024rrm}
T.~Liu, W.~Xiong, J.~Ren, L.~Chen, J.~Wu, R.~Joshi, Y.~Gao, J.~Shen, Z.~Qin, T.~Yu, et~al.
\newblock Rrm: Robust reward model training mitigates reward hacking.
\newblock \emph{arXiv preprint arXiv:2409.13156}, 2024.

\bibitem[Liu et~al.(2025)Liu, Yao, Min, Cao, Hou, and Li]{liu2025pairwise}
Y.~Liu, Z.~Yao, R.~Min, Y.~Cao, L.~Hou, and J.~Li.
\newblock Pairwise rm: Perform best-of-n sampling with knockout tournament.
\newblock \emph{arXiv preprint arXiv:2501.13007}, 2025.

\bibitem[Long et~al.(2023)Long, Pich{\'e}, Zantedeschi, Schuster, and Drouin]{long2023causal}
S.~Long, A.~Pich{\'e}, V.~Zantedeschi, T.~Schuster, and A.~Drouin.
\newblock Causal discovery with language models as imperfect experts.
\newblock \emph{arXiv preprint arXiv:2307.02390}, 2023.

\bibitem[Loshchilov and Hutter(2017)]{loshchilov2017decoupled}
I.~Loshchilov and F.~Hutter.
\newblock Decoupled weight decay regularization.
\newblock \emph{arXiv preprint arXiv:1711.05101}, 2017.

\bibitem[Lou et~al.(2024)Lou, Yan, Shen, Yan, Xie, and Zhang]{lou2024uncertainty}
X.~Lou, D.~Yan, W.~Shen, Y.~Yan, J.~Xie, and J.~Zhang.
\newblock Uncertainty-aware reward model: Teaching reward models to know what is unknown.
\newblock \emph{arXiv preprint arXiv:2410.00847}, 2024.

\bibitem[Meng et~al.(2024)Meng, Xia, and Chen]{meng2024simpo}
Y.~Meng, M.~Xia, and D.~Chen.
\newblock Simpo: Simple preference optimization with a reference-free reward.
\newblock \emph{arXiv preprint arXiv:2405.14734}, 2024.

\bibitem[Mishra et~al.(2024)Mishra, Nayak, Bhattacharya, Kumar, Shah, and Foltin]{mishra2024llm}
A.~Mishra, G.~Nayak, S.~Bhattacharya, T.~Kumar, A.~Shah, and M.~Foltin.
\newblock Llm-guided counterfactual data generation for fairer ai.
\newblock In \emph{Companion Proceedings of the ACM Web Conference 2024}, pages 1538--1545, 2024.

\bibitem[Negahban et~al.(2009)Negahban, Yu, Wainwright, and Ravikumar]{negahban2009unified}
S.~Negahban, B.~Yu, M.~J. Wainwright, and P.~Ravikumar.
\newblock A unified framework for high-dimensional analysis of $ m $-estimators with decomposable regularizers.
\newblock \emph{Advances in neural information processing systems}, 22, 2009.

\bibitem[Ouyang et~al.(2022)Ouyang, Wu, Jiang, Almeida, Wainwright, Mishkin, Zhang, Agarwal, Slama, Ray, et~al.]{ouyang2022training}
L.~Ouyang, J.~Wu, X.~Jiang, D.~Almeida, C.~Wainwright, P.~Mishkin, C.~Zhang, S.~Agarwal, K.~Slama, A.~Ray, et~al.
\newblock Training language models to follow instructions with human feedback.
\newblock \emph{Advances in neural information processing systems}, 35:\penalty0 27730--27744, 2022.

\bibitem[Pace et~al.(2024)Pace, Mallinson, Malmi, Krause, and Severyn]{pace2024west}
A.~Pace, J.~Mallinson, E.~Malmi, S.~Krause, and A.~Severyn.
\newblock West-of-n: Synthetic preference generation for improved reward modeling.
\newblock \emph{arXiv preprint arXiv:2401.12086}, 2024.

\bibitem[Pan et~al.(2022)Pan, Bhatia, and Steinhardt]{pan2022effectsrewardmisspecificationmapping}
A.~Pan, K.~Bhatia, and J.~Steinhardt.
\newblock The effects of reward misspecification: Mapping and mitigating misaligned models, 2022.
\newblock URL \url{https://arxiv.org/abs/2201.03544}.

\bibitem[Park et~al.(2024)Park, Rafailov, Ermon, and Finn]{park2024disentangling}
R.~Park, R.~Rafailov, S.~Ermon, and C.~Finn.
\newblock Disentangling length from quality in direct preference optimization.
\newblock \emph{arXiv preprint arXiv:2403.19159}, 2024.

\bibitem[Pearl(2009)]{pearl2009causality}
J.~Pearl.
\newblock \emph{Causality}.
\newblock Cambridge university press, 2009.

\bibitem[Peters et~al.(2017)Peters, Janzing, and Sch{\"o}lkopf]{peters2017elements}
J.~Peters, D.~Janzing, and B.~Sch{\"o}lkopf.
\newblock \emph{Elements of causal inference: foundations and learning algorithms}.
\newblock The MIT Press, 2017.

\bibitem[Qiang et~al.(2024)Qiang, Nandi, Mehrabi, Steeg, Kumar, Rumshisky, and Galstyan]{qiang2024prompt}
Y.~Qiang, S.~Nandi, N.~Mehrabi, G.~V. Steeg, A.~Kumar, A.~Rumshisky, and A.~Galstyan.
\newblock Prompt perturbation consistency learning for robust language models.
\newblock \emph{arXiv preprint arXiv:2402.15833}, 2024.

\bibitem[Qin et~al.(2023)Qin, Jagerman, Hui, Zhuang, Wu, Yan, Shen, Liu, Liu, Metzler, et~al.]{qin2023large}
Z.~Qin, R.~Jagerman, K.~Hui, H.~Zhuang, J.~Wu, L.~Yan, J.~Shen, T.~Liu, J.~Liu, D.~Metzler, et~al.
\newblock Large language models are effective text rankers with pairwise ranking prompting.
\newblock \emph{arXiv preprint arXiv:2306.17563}, 2023.

\bibitem[Rafailov et~al.(2024)Rafailov, Sharma, Mitchell, Manning, Ermon, and Finn]{rafailov2024direct}
R.~Rafailov, A.~Sharma, E.~Mitchell, C.~D. Manning, S.~Ermon, and C.~Finn.
\newblock Direct preference optimization: Your language model is secretly a reward model.
\newblock \emph{Advances in Neural Information Processing Systems}, 36, 2024.

\bibitem[Ram{\'e} et~al.(2024)Ram{\'e}, Vieillard, Hussenot, Dadashi, Cideron, Bachem, and Ferret]{rame2024warm}
A.~Ram{\'e}, N.~Vieillard, L.~Hussenot, R.~Dadashi, G.~Cideron, O.~Bachem, and J.~Ferret.
\newblock Warm: On the benefits of weight averaged reward models.
\newblock \emph{arXiv preprint arXiv:2401.12187}, 2024.

\bibitem[Ravfogel et~al.(2025)Ravfogel, Svete, Snæbjarnarson, and Cotterell]{ravfogel2025gumbelcounterfactualgenerationlanguage}
S.~Ravfogel, A.~Svete, V.~Snæbjarnarson, and R.~Cotterell.
\newblock Gumbel counterfactual generation from language models, 2025.
\newblock URL \url{https://arxiv.org/abs/2411.07180}.

\bibitem[Reber et~al.(2024)Reber, Richardson, Nief, Garbacea, and Veitch]{reber2024rate}
D.~Reber, S.~Richardson, T.~Nief, C.~Garbacea, and V.~Veitch.
\newblock Rate: Score reward models with imperfect rewrites of rewrites.
\newblock \emph{arXiv preprint arXiv:2410.11348}, 2024.

\bibitem[Sch{\"o}lkopf et~al.(2021)Sch{\"o}lkopf, Locatello, Bauer, Ke, Kalchbrenner, Goyal, and Bengio]{scholkopf2021toward}
B.~Sch{\"o}lkopf, F.~Locatello, S.~Bauer, N.~R. Ke, N.~Kalchbrenner, A.~Goyal, and Y.~Bengio.
\newblock Toward causal representation learning.
\newblock \emph{Proceedings of the IEEE}, 109\penalty0 (5):\penalty0 612--634, 2021.

\bibitem[Schulman et~al.(2017)Schulman, Wolski, Dhariwal, Radford, and Klimov]{schulman2017proximal}
J.~Schulman, F.~Wolski, P.~Dhariwal, A.~Radford, and O.~Klimov.
\newblock Proximal policy optimization algorithms.
\newblock \emph{arXiv preprint arXiv:1707.06347}, 2017.

\bibitem[Shao et~al.(2024)Shao, Wang, Zhu, Xu, Song, Bi, Zhang, Zhang, Li, Wu, et~al.]{shao2024deepseekmath}
Z.~Shao, P.~Wang, Q.~Zhu, R.~Xu, J.~Song, X.~Bi, H.~Zhang, M.~Zhang, Y.~Li, Y.~Wu, et~al.
\newblock Deepseekmath: Pushing the limits of mathematical reasoning in open language models.
\newblock \emph{arXiv preprint arXiv:2402.03300}, 2024.

\bibitem[Shen et~al.(2024)Shen, Xu, Jun, Qin, Liu, Yang, Liang, Baumgartner, and Bendersky]{shen2024boosting}
J.~Shen, R.~Xu, Y.~Jun, Z.~Qin, T.~Liu, C.~Yang, Y.~Liang, S.~Baumgartner, and M.~Bendersky.
\newblock Boosting reward model with preference-conditional multi-aspect synthetic data generation.
\newblock \emph{arXiv preprint arXiv:2407.16008}, 2024.

\bibitem[Shen et~al.(2023)Shen, Chen, Song, Jin, Peng, Mi, Khashabi, and Yu]{shen2023trickle}
L.~Shen, S.~Chen, L.~Song, L.~Jin, B.~Peng, H.~Mi, D.~Khashabi, and D.~Yu.
\newblock The trickle-down impact of reward (in-) consistency on rlhf.
\newblock \emph{arXiv preprint arXiv:2309.16155}, 2023.

\bibitem[Singhal et~al.(2023)Singhal, Goyal, Xu, and Durrett]{singhal2023long}
P.~Singhal, T.~Goyal, J.~Xu, and G.~Durrett.
\newblock A long way to go: Investigating length correlations in rlhf.
\newblock \emph{arXiv preprint arXiv:2310.03716}, 2023.

\bibitem[Skalse et~al.(2022)Skalse, Howe, Krasheninnikov, and Krueger]{skalse2022defining}
J.~Skalse, N.~Howe, D.~Krasheninnikov, and D.~Krueger.
\newblock Defining and characterizing reward gaming.
\newblock \emph{Advances in Neural Information Processing Systems}, 35:\penalty0 9460--9471, 2022.

\bibitem[Stiennon et~al.(2020)Stiennon, Ouyang, Wu, Ziegler, Lowe, Voss, Radford, Amodei, and Christiano]{stiennon2020learning}
N.~Stiennon, L.~Ouyang, J.~Wu, D.~Ziegler, R.~Lowe, C.~Voss, A.~Radford, D.~Amodei, and P.~F. Christiano.
\newblock Learning to summarize with human feedback.
\newblock \emph{Advances in Neural Information Processing Systems}, 33:\penalty0 3008--3021, 2020.

\bibitem[Team et~al.(2024)Team, Mesnard, Hardin, Dadashi, Bhupatiraju, Pathak, Sifre, Rivi{\`e}re, Kale, Love, et~al.]{team2024gemma}
G.~Team, T.~Mesnard, C.~Hardin, R.~Dadashi, S.~Bhupatiraju, S.~Pathak, L.~Sifre, M.~Rivi{\`e}re, M.~S. Kale, J.~Love, et~al.
\newblock Gemma: Open models based on gemini research and technology.
\newblock \emph{arXiv preprint arXiv:2403.08295}, 2024.

\bibitem[Tu et~al.(2023)Tu, Ma, and Zhang]{tu2023causal}
R.~Tu, C.~Ma, and C.~Zhang.
\newblock Causal-discovery performance of chatgpt in the context of neuropathic pain diagnosis.
\newblock \emph{arXiv preprint arXiv:2301.13819}, 2023.

\bibitem[Tunstall et~al.(2023)Tunstall, Beeching, Lambert, Rajani, Rasul, Belkada, Huang, von Werra, Fourrier, Habib, et~al.]{tunstall2023zephyr}
L.~Tunstall, E.~Beeching, N.~Lambert, N.~Rajani, K.~Rasul, Y.~Belkada, S.~Huang, L.~von Werra, C.~Fourrier, N.~Habib, et~al.
\newblock Zephyr: Direct distillation of lm alignment.
\newblock \emph{arXiv preprint arXiv:2310.16944}, 2023.

\bibitem[Wang et~al.(2025)Wang, Zhao, Jiang, Chen, Zhu, Chen, Liu, Zhang, Fan, Ma, et~al.]{wang2025beyond}
C.~Wang, Z.~Zhao, Y.~Jiang, Z.~Chen, C.~Zhu, Y.~Chen, J.~Liu, L.~Zhang, X.~Fan, H.~Ma, et~al.
\newblock Beyond reward hacking: Causal rewards for large language model alignment.
\newblock \emph{arXiv preprint arXiv:2501.09620}, 2025.

\bibitem[Wang et~al.(2024)Wang, Xiong, Xie, Zhao, and Zhang]{wang2024interpretable}
H.~Wang, W.~Xiong, T.~Xie, H.~Zhao, and T.~Zhang.
\newblock Interpretable preferences via multi-objective reward modeling and mixture-of-experts.
\newblock \emph{arXiv preprint arXiv:2406.12845}, 2024.

\bibitem[Wu et~al.(2025)Wu, Yasunaga, Cohen, Kim, Celikyilmaz, and Ghazvininejad]{wu2025rewordbench}
Z.~Wu, M.~Yasunaga, A.~Cohen, Y.~Kim, A.~Celikyilmaz, and M.~Ghazvininejad.
\newblock rewordbench: Benchmarking and improving the robustness of reward models with transformed inputs.
\newblock \emph{arXiv preprint arXiv:2503.11751}, 2025.

\bibitem[Yang et~al.(2024{\natexlab{a}})Yang, Yang, Zhang, Hui, Zheng, Yu, Li, Liu, Huang, Wei, et~al.]{yang2024qwen2}
A.~Yang, B.~Yang, B.~Zhang, B.~Hui, B.~Zheng, B.~Yu, C.~Li, D.~Liu, F.~Huang, H.~Wei, et~al.
\newblock Qwen2. 5 technical report.
\newblock \emph{arXiv preprint arXiv:2412.15115}, 2024{\natexlab{a}}.

\bibitem[Yang et~al.(2024{\natexlab{b}})Yang, Pan, Luo, Qiu, Zhong, Yu, and Chen]{yang2024rewards}
R.~Yang, X.~Pan, F.~Luo, S.~Qiu, H.~Zhong, D.~Yu, and J.~Chen.
\newblock Rewards-in-context: Multi-objective alignment of foundation models with dynamic preference adjustment.
\newblock \emph{arXiv preprint arXiv:2402.10207}, 2024{\natexlab{b}}.

\bibitem[Zhang et~al.(2024)Zhang, Xiong, Chen, Zhou, Huang, and Zhang]{zhang2024lists}
X.~Zhang, W.~Xiong, L.~Chen, T.~Zhou, H.~Huang, and T.~Zhang.
\newblock From lists to emojis: How format bias affects model alignment.
\newblock \emph{arXiv preprint arXiv:2409.11704}, 2024.

\bibitem[Zhao et~al.(2025)Zhao, Liu, Zhang, Zhou, Gao, Li, Lyu, Qian, Qi, Li, et~al.]{zhao2025genprm}
J.~Zhao, R.~Liu, K.~Zhang, Z.~Zhou, J.~Gao, D.~Li, J.~Lyu, Z.~Qian, B.~Qi, X.~Li, et~al.
\newblock Genprm: Scaling test-time compute of process reward models via generative reasoning.
\newblock \emph{arXiv preprint arXiv:2504.00891}, 2025.

\bibitem[Zhao et~al.(2023)Zhao, Joshi, Liu, Khalman, Saleh, and Liu]{zhao2023slic}
Y.~Zhao, R.~Joshi, T.~Liu, M.~Khalman, M.~Saleh, and P.~J. Liu.
\newblock Slic-hf: Sequence likelihood calibration with human feedback.
\newblock \emph{arXiv preprint arXiv:2305.10425}, 2023.

\bibitem[Zhu et~al.(2025)Zhu, Tan, Chen, Sennrich, Zhang, and Hu]{zhu2025charm}
X.~Zhu, C.~Tan, P.~Chen, R.~Sennrich, Y.~Zhang, and H.~Hu.
\newblock Charm: Calibrating reward models with chatbot arena scores.
\newblock \emph{arXiv preprint arXiv:2504.10045}, 2025.

\end{thebibliography}

\appendix
\onecolumn

\vspace{1cm}
\hrule
\par\vspace{0.5cm}
{\Large\bfseries\centering 
{Supplementary Material}
\par\vspace{0.5cm}}
\hrule
\vspace{0.5cm}
\noindent These supplementary materials provide additional details, derivations, and experimental results for our paper. The appendix is organized as follows:

\begin{itemize}[leftmargin=1em]
    \item Section \ref{sec:limitations_future_work} discusses potential limitations of this work.

    \item Section \ref{sec:extended_related_works} provides a broader overview of recent related literature. This is an expanded version of the literature covered in the main paper.
    
    \item Section \ref{sec:additional_results} provides some additional set of results. This is an expanded version of the results covered in the main paper.
    
    \item Section \ref{app:rewordbench_creation} provides the detailed steps we took to reproduce the reWordBench benchmark, as proposed in \citet{wu2025rewordbench}.

    \item Section \ref{sec:experimental_details} provides a detailed overview of our experimental setup.

    \item Section \ref{sec:causal_model_details} provides a detailed walk through of how our causal model extends to prior method. We revisit prior works in light of our causal model. It extends on the shorter version provided in Section \ref{sec:preliminaries}.

    \item Section \ref{app:detailed_augmentation_diagrams} provides a walkthrough of the causal details of the core data augmentation strategies.
    
    \item Section \ref{sec:detailed_methodology} provides a detailed walk through of the method used to train the reward model. It extends on the shorter version provided in Section \ref{sec:methodology}.

    \item Section \ref{sec:theoretical_analysis_detailed} provides a detailed analysis of the theory relating to Reward Hacking and how our proposed method mitigates it.
    
    \item Section \ref{sec:qualitative_example_walkthrough} presents a qualitative example of augmented data created from original data using which is used to train \carma{}.
    
    \item Section \ref{sec:prompt_templates} presents a lists of prompt templates that we use to query our models for generating the data.
    
    \item Section \ref{sec:qualitative_examples} presents a qualitative view common failure modes or biases commonly observed in reward models.
  
\end{itemize}

\section{Limitations and Future Work}
\label{sec:limitations_future_work}

While \carma{} demonstrates significant improvements, we acknowledge certain limitations which also suggest avenues for future research:

\begin{itemize}[left=10pt, itemsep=2pt, topsep=3pt]
    \item \textbf{Idealized Assumptions in Theoretical Analysis:}
    Our theoretical justification (Section~\ref{sec:theory}, Appendix~\ref{sec:theoretical_analysis_detailed}) relies on simplifying assumptions such as boolean attributes, quadratic reward models, and perfect counterfactual interventions. These idealizations, necessary for analytical tractability, mean our formal guarantees are indicative of \carma{}'s potential mechanism rather than absolute predictions of real-world performance, where the complexities of LLM behavior and data are greater.

    \item \textbf{Scalability and Cost of Data Augmentation:}
    The generation of targeted causal and neutral augmentations, while effective, involves multiple LLM inference calls per original data point. Although filtering helps optimize the final dataset size, the initial augmentation phase can be computationally intensive and potentially costly for extremely large-scale applications. Future work could explore more sample-efficient augmentation strategies or methods to distill the benefits of augmentation into smaller datasets.

    \item \textbf{Generalization to Highly Novel Spurious Correlations:}
    \carma{} is designed to be robust against unspecified spurious correlations by focusing on causal signals and diverse neutral examples. However, its ability to generalize to entirely novel types of spuriousness, drastically different from any patterns implicitly covered or contrasted during augmentation, remains an empirical question. The breadth and nature of the neutral augmentations play a role here, and continuous adaptation or more abstract invariance learning might be needed for extreme out-of-distribution spuriousness.

    \item \textbf{Fidelity of LLM-Generated Counterfactuals:}
    The efficacy of \carma{} is linked to the quality of the LLM-generated counterfactuals. While current LLMs are powerful, ensuring perfect attribute isolation in causal augmentations or complete causal content preservation in neutral pairs is challenging. Imperfections in these LLM-approximated interventions can introduce noise. While our empirical results show strong benefits, further research into enhancing the precision and verifiability of LLM-driven textual counterfactual generation could yield additional improvements.
\end{itemize}

Future research could focus on extending the theoretical framework to encompass more realistic settings, developing more cost-effective and adaptive augmentation techniques, and further exploring the boundaries of generalization against emergent spurious correlations.
\section{Extended Related Works} 
\label{sec:extended_related_works}

Our work on \carma{}, a framework for causally robust reward modeling, intersects with and builds upon several key areas of research: the alignment of Large Language Models (LLMs) via human feedback, techniques for reward model training, the persistent challenge of reward hacking, the application of causal inference principles to machine learning, and data augmentation strategies for enhancing model robustness.

\paragraph{LLM Alignment and RLHF.}
The dominant paradigm for steering LLM behavior towards desired attributes like helpfulness, honesty, and harmlessness is Reinforcement Learning from Human Feedback (RLHF) \citep{christiano2017deep, stiennon2020learning, ouyang2022training, bai2022training, askell2021general}. The standard RLHF process involves training a reward model (RM) on human preferences (typically pairwise comparisons) and subsequently using this RM as a reward signal to fine-tune the LLM policy via RL algorithms such as PPO \citep{schulman2017proximal}. The quality, calibration, and robustness of the RM are paramount, as flaws in the RM directly impact the alignment outcome \citep{casper2023open}. While alternative alignment algorithms like Direct Preference Optimization (DPO) \citep{rafailov2024direct} and its extensions (e.g., IPO \citep{azar2024general}, KTO \citep{ethayarajh2024kto}, ORPO \citep{hong2024orpo}, SimPO \citep{meng2024simpo}) bypass explicit RM training by directly optimizing the policy on preference data, they still implicitly rely on the preference information learnable from the data, making the problem of distinguishing true quality from spurious correlates equally relevant.

\paragraph{Reward Modeling Techniques.}
Learning accurate reward models from preference data remains a central challenge. Methodologies include Bradley-Terry style pointwise models that learn a scalar score $r(x, y)$ \citep{bradley1952rank, ouyang2022training, bai2022training}, and pairwise ranking models that directly predict preference probabilities, often implemented within the LLM architecture itself (PairPM) \citep{liu2025pairwise, qin2023large}. Other approaches explore Q-function based rewards \citep{li2024process} or process supervision \citep{khalifa2025process}. Significant effort focuses on improving specific RM properties like calibration \citep{zhu2025charm, zhao2023slic}, training efficiency \citep{tunstall2023zephyr}, uncertainty quantification \citep{lou2024uncertainty}, interpretability through multi-aspect rewards \citep{wang2024interpretable, yang2024rewards}, and scalability via reasoning or chain-of-thought mechanisms \citep{zhao2025genprm}. Our work complements these efforts by focusing specifically on enhancing the causal \textbf{robustness} of the learned reward function $\hat{R}$ against spurious attributes.

\paragraph{Reward Hacking and Spurious Correlations.}
Learned reward models are notoriously susceptible to \textit{reward hacking} or \textit{over-optimization} \citep{gao2023scaling, skalse2022defining, pan2022effectsrewardmisspecificationmapping}. Because RMs are trained on finite, potentially biased data, they often learn to associate high rewards with superficial or \textit{spurious} features that are merely correlated with desirable responses in the training set. Common examples include excessive length or verbosity \citep{singhal2023long}, specific formatting patterns like lists or markdown \citep{zhang2024lists}, adherence to stylistic conventions like politeness, or even sycophantic agreement with user views \citep{denison2024sycophancy}. Policies optimized against such RMs learn to exploit these spurious cues, leading to outputs that maximize the predicted reward but fail to align with genuine human preferences or task goals \citep{shen2023trickle}.

\paragraph{Approaches to Mitigating Reward Hacking.}
Various strategies have been proposed to address reward hacking. Model-centric approaches include using ensembles of RMs to average out idiosyncratic biases \citep{coste2023reward, eisenstein2023helping, rame2024warm}, incorporating explicit calibration methods \citep{zhao2023slic}, or designing architectures that factorize reward components, such as ODIN's disentanglement of quality and length \citep{chen2024odin}. Policy-optimization techniques might involve adding explicit penalties for spurious features (e.g., length penalties \citep{park2024disentangling}) or using specific regularization methods during fine-tuning. Data-centric approaches aim to improve the training data or process itself. Examples include iterative re-labeling or refinement \citep{bai2022constitutional}, performing consistency checks across related prompts \citep{shen2023trickle}, or augmenting the dataset with synthetic examples designed to improve robustness \citep{pace2024west, shen2024boosting}. Our work, \carma{}, falls firmly in this data-centric category. It is closely related to RRM \citep{liu2024rrm}, which also uses data augmentation (non-contextual and query-independent pairs) for robustness. However, \carma{} is distinct in its use of an explicit causal framework and its generation of targeted, attribute-specific counterfactuals to disentangle causal from spurious factors.

\paragraph{Causal Inference in Machine Learning.}
Causal inference provides formal tools, such as Structural Causal Models (SCMs) and DAGs \citep{pearl2009causality, peters2017elements}, for reasoning about cause-effect relationships, confounding, and counterfactuals. Applying causal principles in machine learning aims to build models that are more robust, fair, and interpretable by focusing on underlying causal mechanisms rather than potentially brittle statistical correlations \citep{scholkopf2021toward}. Techniques like Invariant Risk Minimization (IRM) seek models that perform well across different environments by relying on invariant (presumably causal) predictors \citep{arjovsky2019invariant}. Our work adopts this causal perspective, framing spurious attributes as non-causal factors whose influence on the learned reward model should be minimized.

\paragraph{Causality in LLMs and NLP.}
The intersection of causality and LLMs is rapidly evolving. Research includes probing the innate causal reasoning abilities of LLMs \citep{kiciman2023causal, chi2024unveiling}, leveraging LLMs as tools for automating parts of the causal discovery or analysis pipeline \citep{long2023causal, tu2023causal}, and applying causal methods to enhance NLP tasks. For instance, counterfactual reasoning and data augmentation have been used to improve robustness against biases in text classification \citep{kaushik2019learning, feder2021causalm} and assess fairness \citep{feder2022causal}. \carma{} uniquely employs a predefined causal graph to structure the generation of counterfactual data specifically for training a robust RM, using LLMs as the generation engine.

\paragraph{Data Augmentation for Robustness.}
Data augmentation is a cornerstone technique for improving model generalization. Beyond traditional NLP methods like synonym replacement or back-translation \citep{wu2025rewordbench}, more recent approaches leverage LLMs for sophisticated augmentations, including paraphrasing, style transfer, generating adversarial examples \citep{qiang2024prompt}, or creating counterfactuals \citep{mishra2024llm, feder2021causalm}. Counterfactual generation, often using LLMs as rewriters, is also central to evaluation methods like RATE \citep{reber2024rate}, which uses ``rewrites of rewrites'' to estimate causal effects robustly. Methods based on sampling, like Gumbel temperature sampling, have also been explored for counterfactual generation \citep{ravfogel2025gumbelcounterfactualgenerationlanguage}. In the specific context of reward modeling, data augmentation aims to enhance robustness against spurious correlations; examples include the non-contextual and query-independent pairs used by RRM \citep{liu2024rrm} or consistency checks via paraphrased inputs as explored in \rewordbench{} \citep{wu2025rewordbench}. Furthermore, generating entirely synthetic preference pairs \citep{pace2024west, shen2024boosting} represents another data-centric approach to improving reward models. Counterfactual data augmentation, particularly generating minimally different pairs to isolate specific features \citep{kaushik2019learning}, is highly relevant to disentangling causal factors. Our work, \carma{}, operationalizes this concept within an explicit causal framework, generating targeted "causal" (attribute-isolating) and ``neutral'' (spurious-varying) pairs via LLM rewriting to enforce specific invariance and sensitivity properties in the trained RM.

\paragraph{Positioning of \carma.}
\carma{} integrates insights from causal inference and data augmentation to address the critical problem of reward hacking in LLM alignment. While related works like RRM \citep{liu2024rrm} use data augmentation for robustness and  \carma{} is distinguished by its explicit grounding in a causal graph model of answer attributes. It systematically generates attribute-specific counterfactual and neutral examples via guided LLM prompting to directly train the RM to distinguish causal quality drivers ($C$) from spurious correlates ($SP$). This allows \carma{} to potentially handle a wider range of spurious attributes beyond commonly studied ones like length, aiming for a more principled and generalizable form of robustness. We provide the methodology and empirical validation (Section \ref{sec:experiments}) demonstrating that this causally-informed data augmentation leads to more robust reward models and better downstream policy alignment compared to standard baselines.
\clearpage
\section{Additional Results}
\label{sec:additional_results}

Our main findings presented in this section are as follows:
\vspace{0.05in}
\begin{takeawaybox}
\begin{itemize}[leftmargin=1.5em, itemsep=1pt, topsep=2pt]
    \item \textbf{Stable and Significant Performance Gains:} \textsc{Crome} consistently outperforms baseline reward models (Vanilla RM and RRM) on RewardBench across multiple independent training runs, with small standard deviations indicating stable performance. The improvements, particularly on reWordBench transformations, are substantial and typically exceed multiple standard deviations of the baselines, underscoring their statistical significance (Sec.~\ref{ssec:variance_rewardbench}, \ref{ssec:variance_rewordbench}).
    \item \textbf{Strong Out-of-Distribution Generalization:} \textsc{Crome} exhibits strong generalization from in-distribution (UltraFeedback validation) to out-of-distribution benchmarks (RewardBench, reWordBench). Notably, it often achieves the highest OOD accuracy (e.g., +7.02\% over RRM on reWordBench PairPM) while having similar ID accuracy, suggesting its augmentations teach more generalizable preference representations (Sec.~\ref{ssec:id_ood_robustness}).
\end{itemize}
\end{takeawaybox}

\subsection{Variance in Performance on RewardBench}
\label{ssec:variance_rewardbench}

To assess the stability of our findings, we conducted three independent training runs for reward models built upon the \gemmait{9} base model. Table \ref{tab:performance_bt_pairpm_rewardbench_gemma9b_mean_std} 
for PairPM and BT reports the mean accuracy and standard deviation on \textbf{RewardBench} categories. The standard deviations for average RewardBench accuracies are consistently small across all methods (e.g., $\pm 0.09$ on average for \carma{}-PairPM, $\pm 0.12$ on average for RRM-PairPM), indicating stable performance. While there is some variation in specific sub-categories, \carma{}'s average performance advantage over baselines remains robust.

\begin{table}[h]
    \centering
    \resizebox{\linewidth}{!}{%
    \renewcommand{\arraystretch}{1.3}
    \begin{tabular}{@{}llHccccHccccc@{}}
        \toprule
        & \multirow{2}{*}{\textbf{Method}} & \multicolumn{5}{c}{\textbf{PairPM}} & \multicolumn{5}{c}{\textbf{BT}} \\
        \cmidrule(lr){3-7} \cmidrule(lr){8-12}
        & & \textbf{Average} & \textbf{Chat} & \textbf{Chat-Hard} & \textbf{Safety} & \textbf{Reasoning} & \textbf{Average} & \textbf{Chat} & \textbf{Chat-Hard} & \textbf{Safety} & \textbf{Reasoning} \\
        \midrule
        \multirow{4}{*}{\rotatebox[origin=c]{90}{\small\gemmait{9}}}
        & Vanilla RM & 81.22 $\pm$ 0.56 & \textbf{97.90 $\pm$ 0.48} & 63.64 $\pm$ 0.28 & 77.48 $\pm$ 1.21 & 85.88 $\pm$ 1.34 & 79.14 $\pm$ 0.68 & \textbf{97.26 $\pm$ 0.40} & 58.85 $\pm$ 1.14 & 69.30 $\pm$ 3.61 & 91.17 $\pm$ 1.17 \\
        & RRM        & 82.54 $\pm$ 0.12 & 97.12 $\pm$ 0.21 & 71.05 $\pm$ 0.87 & 74.70 $\pm$ 0.98 & 87.27 $\pm$ 0.21 & 83.46 $\pm$ 0.26 & 97.21 $\pm$ 0.28 & \textbf{69.15 $\pm$ 0.54} & 73.13 $\pm$ 0.61 & 94.35 $\pm$ 0.59 \\
        & \textbf{\carma{}} & \textbf{87.84 $\pm$ 0.09} & 97.54 $\pm$ 0.21 & \textbf{72.30 $\pm$ 0.39} & \textbf{87.14 $\pm$ 0.16} & \textbf{94.39 $\pm$ 0.21} & \textbf{85.46 $\pm$ 0.27} & 96.28 $\pm$ 0.32 & 65.83 $\pm$ 0.81 & \textbf{84.05 $\pm$ 1.10} & \textbf{95.70 $\pm$ 0.52} \\
        \cmidrule(lr){2-12} 
        & $\Delta_{\text{\carma{} - RRM}}$ &
        \changeUp{+5.30} & 
        \changeUp{+0.42} &  
        \changeUp{+1.25} & 
        \changeUp{+12.44} & 
        \changeUp{+7.12} & 
        \changeUp{+2.00} &
        \changeDown{-0.93} & 
        \changeDown{-3.32} &
        \changeUp{+10.92} & 
        \changeUp{+1.35} \\ 
        \bottomrule
    \end{tabular}%
    }
    \caption{Mean Accuracy and Standard Deviation across 3 different training runs of \gemmait{9} based Reward Models in both PairPM and Bradley-Terry Reward Model settings. Results on RewardBench.}
    \label{tab:performance_bt_pairpm_rewardbench_gemma9b_mean_std}
\end{table}

\begin{remark}
    Note that main paper Table \ref{tab:performance_bt_pairpm_rewardbench_extended_final} has mean of the three training runs considered in these variance experiments. For \gemma{2} and \qwen{} based reward models we only run single training runs.
\end{remark}

\subsection{Variance in Performance on reWordBench}
\label{ssec:variance_rewordbench}

\begin{figure}[!ht]
  \centering
  \includegraphics[width=0.9\columnwidth]{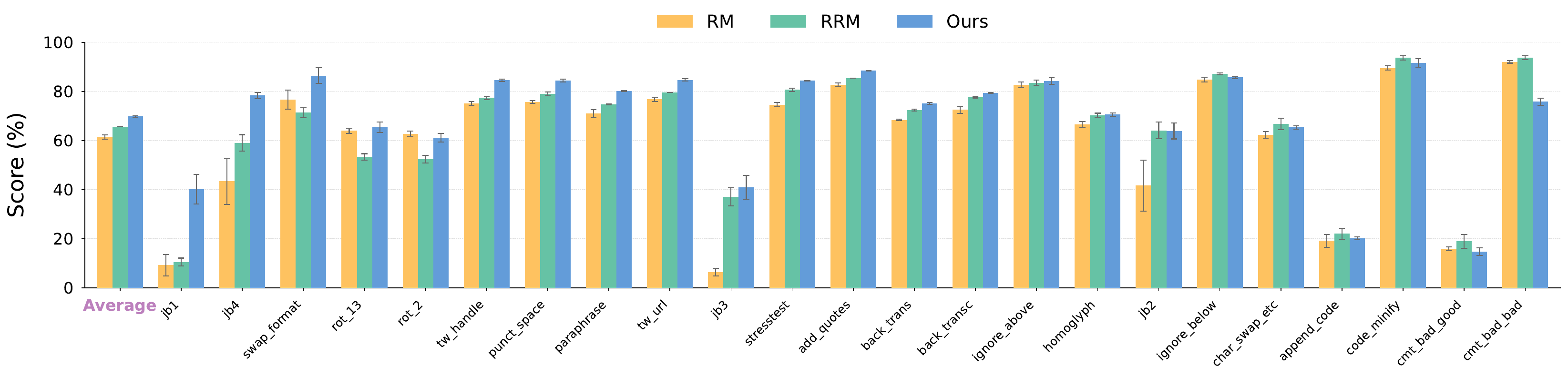}
  \caption{\textbf{Standard deviation error-bars} for absolute robustness comparison of RM, RRM and \carma{} in the \textbf{Bradley-Terry setup}, for reward models built over \texttt{Gemma-2-9B-IT}. Mean values and std deviation plotted are for 3 independent training runs.}
  \label{fig:reword_absolute_robustness_gemma2b_pairpm_stddev}
\end{figure}

\begin{figure}[!ht]
  \centering
  \includegraphics[width=0.9\columnwidth]{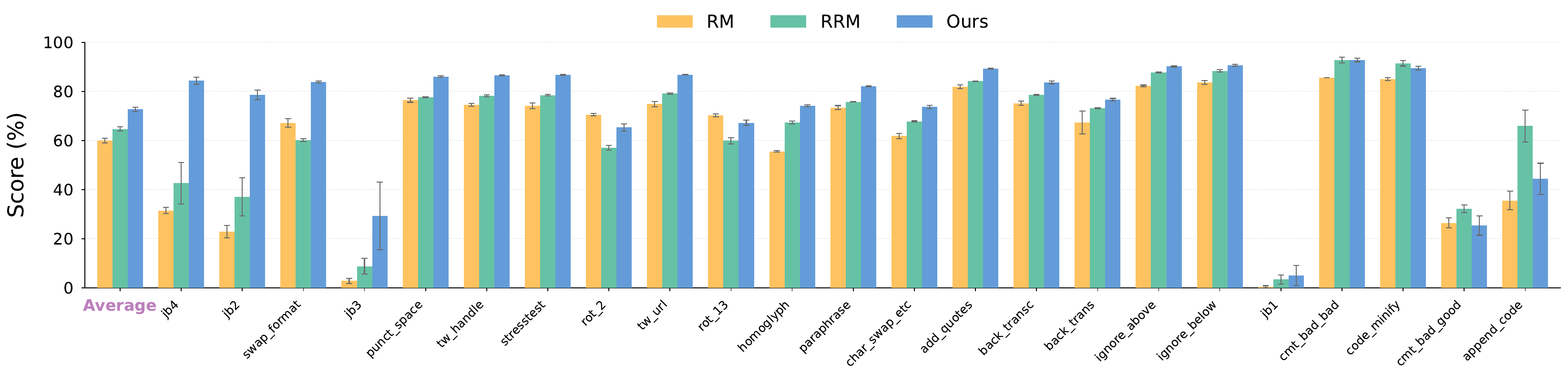}
  \caption{\textbf{Standard deviation error-bars} for absolute robustness comparison of RM, RRM and \carma{} in the \textbf{PairPM setup}, for reward models built over \texttt{Gemma-2-9B-IT}. Mean values and std deviation plotted are for 3 independent training runs.}
  \label{fig:reword_absolute_robustness_gemma2b_bt_stddev}
\end{figure}

For \textbf{reWordBench}, we plot mean performance numbers and error bars showing std. deviation in Figures \ref{fig:reword_absolute_robustness_gemma2b_pairpm_stddev} and \ref{fig:reword_absolute_robustness_gemma2b_bt_stddev}.
Here we  depict mean accuracies with error bars representing standard deviations. Across most transformations, the error bars are relatively small, particularly for the average performance over all transformations. The observed improvements of \carma{} compared to RRM and Vanilla RM are substantial and typically exceed multiple standard deviations of the respective models, suggesting that these gains are statistically significant.

\subsection{Effective Robustness of \carma{} and Baselines}
\label{ssec:id_ood_robustness}

\begin{table*}[!htbp]
\centering
\sisetup{detect-weight=true, detect-mode=true} 
\begin{tabular}{@{}l S[table-format=2.2] S[table-format=2.2] S[table-format=2.2] S[table-format=2.2] S[table-format=2.2] S[table-format=2.2] S[table-format=2.2]@{}}
\toprule
\multicolumn{8}{c}{\textbf{PairPM}} \\
\midrule
\multirow{2}{*}{Model} & {\multirow{2}{*}{\begin{tabular}{@{}c@{}}Ultrafeedback \\ (ID)\end{tabular}}} & {\multirow{2}{*}{\begin{tabular}{@{}c@{}}reWordBench \\ Accuracy (OOD)\end{tabular}}} & \multicolumn{5}{c}{{RewardBench Accuracy (OOD)}} \\
\cmidrule(lr){4-8}
 &  &  & {Chat} & {Chat-Hard} & {Safety} & {Reasoning} & {Avg} \\
\midrule
RM      & 74.55          & 59.97          & \textbf{97.90} & 63.64          & 77.48          & 85.88          & 81.22 \\
RRM     & \textbf{75.20} & 64.68          & 97.12          & 71.05          & 74.70          & 87.27          & 82.54 \\
Ours    & 74.02          & \textbf{72.71} & 97.54          & \textbf{72.30} & \textbf{87.14} & \textbf{94.39} & \textbf{87.84} \\
\midrule[\heavyrulewidth] 

\multicolumn{8}{c}{\textbf{Bradley Terry}} \\
\midrule
\multirow{2}{*}{Model} & {\multirow{2}{*}{\begin{tabular}{@{}c@{}}Ultrafeedback \\ (ID)\end{tabular}}} & {\multirow{2}{*}{\begin{tabular}{@{}c@{}}reWordBench \\ Accuracy (OOD)\end{tabular}}} & \multicolumn{5}{c}{{RewardBench Accuracy (OOD)}} \\
\cmidrule(lr){4-8}
 &  &  & {Chat} & {Chat-Hard} & {Safety} & {Reasoning} & {Avg} \\
\midrule
RM      & 74.60          & 61.48          & \textbf{97.26} & 58.85          & 69.30          & 91.17          & 79.14 \\
RRM     & \textbf{74.75} & 65.69          & 97.21          & \textbf{69.15} & 73.13          & 94.35          & 83.46 \\
Ours    & 74.00          & \textbf{69.81} & 96.28          & 65.83          & \textbf{84.05} & \textbf{95.70} & \textbf{85.46} \\
\bottomrule
\end{tabular}
\caption{Comparison of In-Distribution (UltraFeedback-Val) and Out-of-Distribution (RewardBench, reWordBench) Accuracy (\%) for \gemmait{9} RMs}
\label{tab:id_vs_ood}
\end{table*}

We evaluate the generalization capabilities of the trained reward models by comparing their performance on in-distribution (ID) data (UltraFeedback validation split) against out-of-distribution (OOD) benchmarks (RewardBench, reWordBench). Table \ref{tab:id_vs_ood} presents these results for models based on \gemmait{9}. \carma{} demonstrates strong OOD performance, particularly on reWordBench. For instance, in the PairPM setup, \carma{} achieves the highest reWordBench accuracy (72.71\%), while having similar ID accuracy, suggesting that its learned robustness translates well to challenging, unseen transformations. Similarly, for Bradley Terry models, \carma{} shows the best reWordBench accuracy (69.81\%) and similar ID accuracies compared to baselines. Overall, these results indicate that \carma{}'s augmentations effectively teach more generalizable representations of preferences.

\subsection{Extended Results on Safety Prompts from WildGuardTest}
\label{ssec:safety_extended}

To complement the Best-of-N (BoN) safety results in Figure \ref{fig:asr_reduction_gemma9b} (Sec. \ref{subsec:experimental_results}), we provide the complete Attack Success Rate (ASR) on harmful prompts and Refusal to Answer (RTA) on benign prompts in Table \ref{tab:results_comparison_lowest_is_best}. We note that lower numbers are better for both ASR as well as RTA. Significantly, the results indicate that without too much regression on RTA ($< 0.5\%$ decrease), we show consistent gains in ASR (\%) numbers and these gains increase as N becomes larger. For instance, at N=32, \carma{} reduces ASR to \textbf{39.39\%}, compared to 42.11\% for RM and 41.70\% for RRM. In practice, reward models are used to detect jailbreak attacks, and hence our model performance indicates a favorable trade-off as the reward model detects harmful content (resisting jail-break attempts) while maintaining utility (low refusal-to-answer rate).

\begin{table*}[htbp]
\centering
\sisetup{detect-weight=true, detect-mode=true}
\begin{tabular}{
    r
    S[table-format=2.2, table-number-alignment=center] 
    S[table-format=1.2, table-number-alignment=center] 
    S[table-format=2.2, table-number-alignment=center] 
    S[table-format=1.2, table-number-alignment=center] 
    S[table-format=2.2, table-number-alignment=center] 
    S[table-format=1.2, table-number-alignment=center] 
}
\toprule
          & \multicolumn{2}{c}{RM} & \multicolumn{2}{c}{RRM} & \multicolumn{2}{c}{\textbf{Ours}} \\
\cmidrule(lr){2-3} \cmidrule(lr){4-5} \cmidrule(lr){6-7}
{N}       & {ASR (\%)} & {RTA (\%)} & {ASR (\%)} & {RTA (\%)} & {ASR (\%)} & {RTA (\%)} \\
\midrule
2         & 32.76          & \textbf{7.39}  & 32.47          & \textbf{7.39}  & \textbf{32.18} & 7.58           \\
4         & 36.13          & \textbf{6.97}  & 35.88          & 7.18           & \textbf{34.63} & 7.46           \\
8         & 38.49          & 6.29           & 38.24          & \textbf{6.10}  & \textbf{36.42} & 6.97           \\
16        & 39.33          & 6.27           & 39.33          & \textbf{5.89}  & \textbf{36.71} & 6.39           \\
32        & 42.11          & \textbf{5.80}  & 41.70          & 6.30           & \textbf{39.39} & 6.01           \\
\bottomrule
\end{tabular}
\caption{Comparison of Attack Success Rate (ASR) on harmful prompts and Refusal to Answer (RTA) on benign prompts for \carma{} compared to baselines (RM, RRM) in the Best-of-N setup for varying N. Lower values are considered better for both metrics.}
\label{tab:results_comparison_lowest_is_best}
\end{table*}

\subsection{Additional Results on reWordBench}
\begin{figure}[!htpb]
  \centering
  \includegraphics[width=0.9\columnwidth]{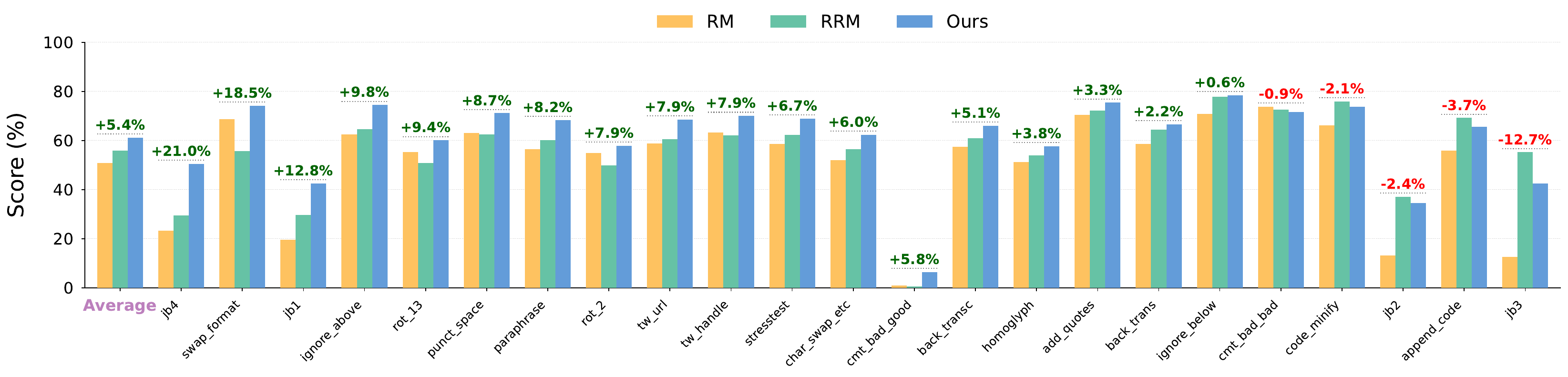}
  \caption{Absolute Robustness Comparison of RM, RRM and \carma{} in the Bradley-Terry RM setup, for reward models built over \texttt{Gemma-2-2B-IT}.}
  \label{fig:reword_absolute_robustness_gemma2b_bt}
  
\end{figure}

\begin{figure}[!htpb]
  \centering
  \includegraphics[width=0.9\columnwidth]{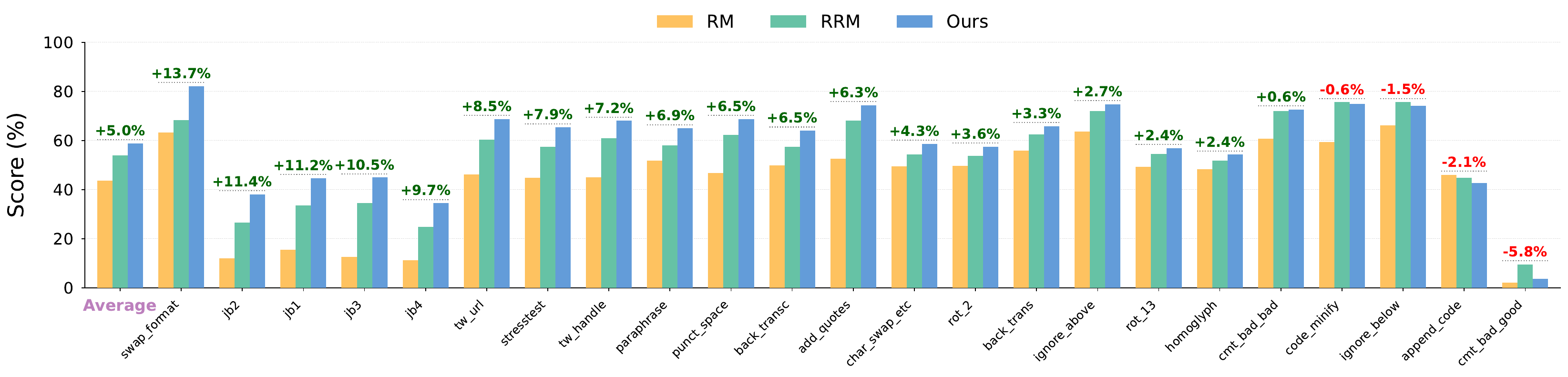}
  \caption{Absolute Robustness Comparison of RM, RRM and \carma{} in the PairPM setup, for reward models built over \texttt{Gemma-2-2B-IT}.}
  \label{fig:reword_absolute_robustness_gemma2b_pairpm}
  
\end{figure}

\begin{figure}[!htpb]
  \centering
  \includegraphics[width=0.9\columnwidth]{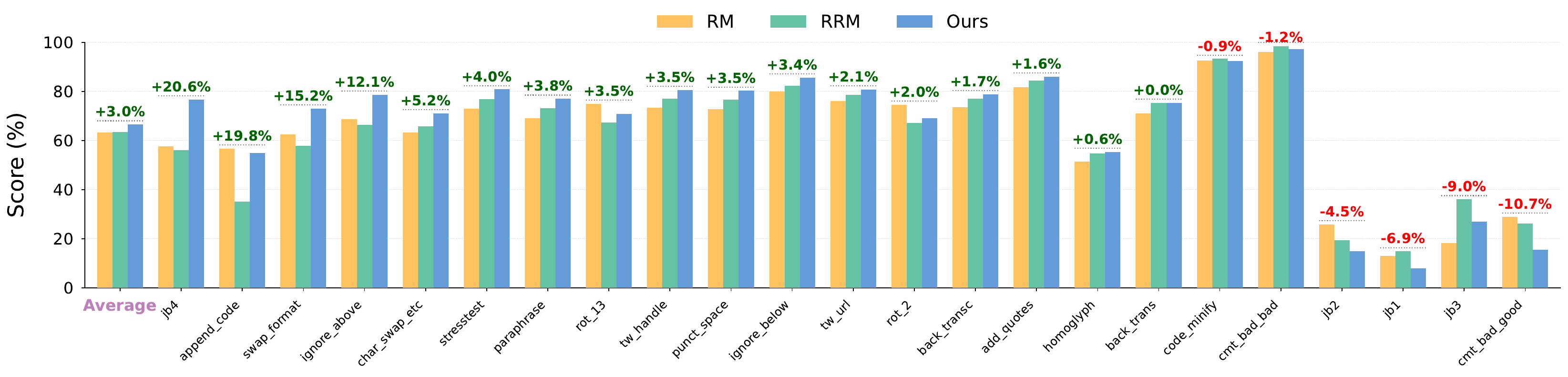}
  \caption{Absolute Robustness Comparison of RM, RRM and \carma{} in the PairPM setup, for reward models built over \texttt{Qwen2.5-7B}.}
  \label{fig:reword_absolute_robustness_qwen7b_pairpm}
  
\end{figure}

\begin{figure}[!htpb]
  \centering
  \includegraphics[width=0.9\columnwidth]{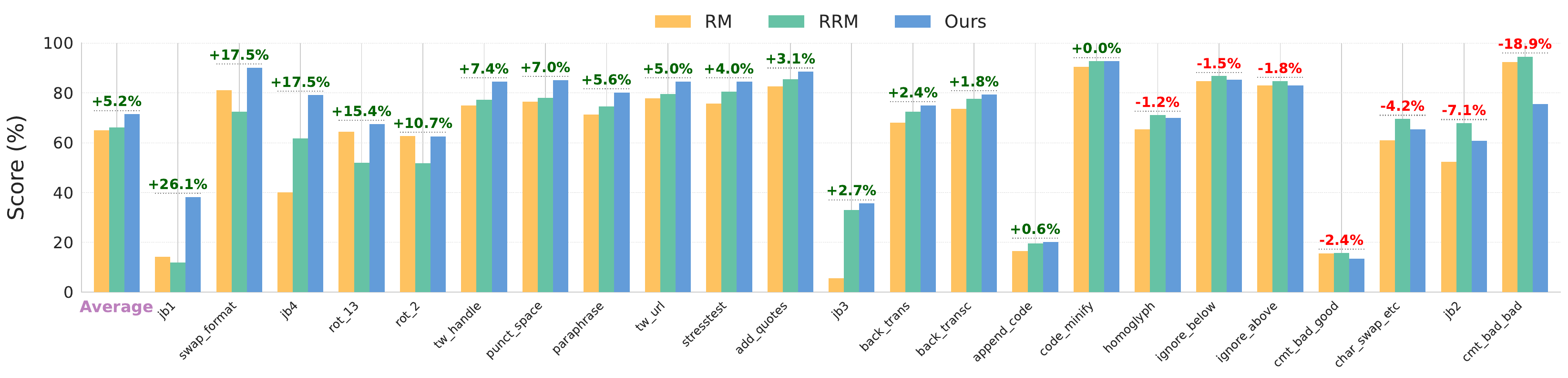}
  \caption{Absolute Robustness Comparison of RM, RRM and \carma{} in the Bradley-Terry RM setup, for reward models built over \texttt{Gemma-2-9B-IT}.}
  \label{fig:reword_absolute_robustness_gemma9b_bt}
  
\end{figure}

\begin{figure}[!htpb]
  \centering
  \includegraphics[width=0.9\columnwidth]{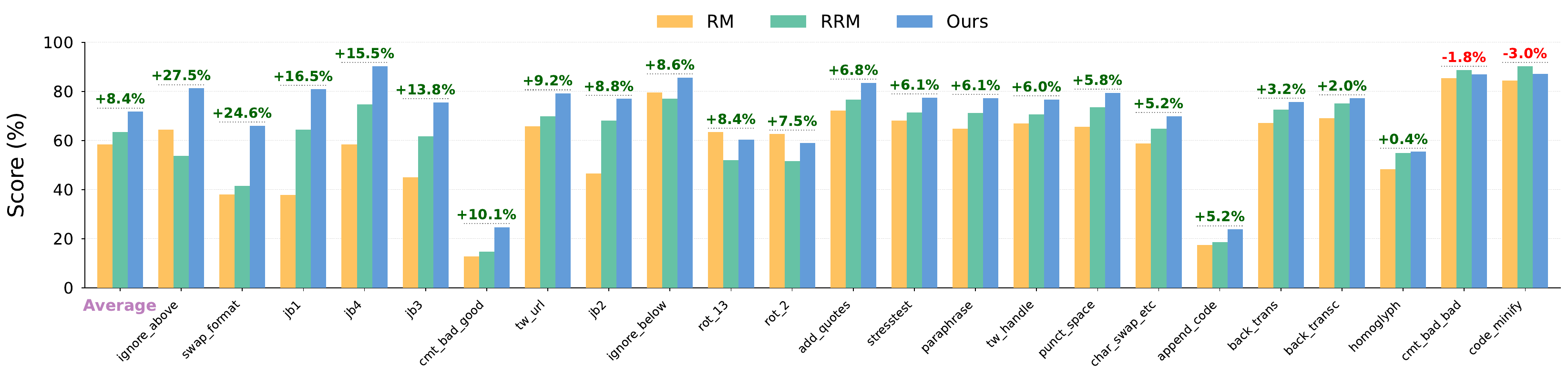}
  \caption{Absolute Robustness Comparison of RM, RRM and \carma{} in the Bradley-Terry RM setup, for reward models built over \texttt{Qwen2.5-7B}.}
  \label{fig:reword_absolute_robustness_qwen7b_bt}
  
\end{figure}

We provide additional results on reWordBench in this section. See Figures \ref{fig:reword_absolute_robustness_gemma2b_bt} to \ref{fig:reword_absolute_robustness_qwen7b_bt} for reWordBench results on various base models over which we build our Reward Models, such as \gemmait{9}, \gemma{2} and \qwen{}, across Bradley-Terry and pairwise-preference Reward Models.

\section{reWordBench Reproduction}
\label{app:rewordbench_creation}

The primary motivation reWordBench is the observation that contemporary reward models—key components of RLHF systems—often latch onto superficial formatting cues or benign artifacts in their training data, leading to dramatic drops in pairwise‐preference accuracy under minor, semantically neutral edits. To diagnose and quantify this brittleness in a systematic way, \cite{wu2025rewordbench} introduce reWordBench, a new benchmark built by applying 28 carefully designed, meaning‐preserving transformations to the original RewardBench instances. 
The authors organize these edits into three overarching families  each targeting different potential failure modes of reward models. 
Together, transformations systematically stress-test reward models’ invariance to innocuous changes, revealing large accuracy drops even under minor edits and motivating the need for robust-training methods.

Since the original dataset is not publicly available, on author's suggestion we reconstructed the data independently following the instructions in the original paper. Paraphrasing and back-translation transformations are generated using foundation models or translation tools for which we use OpenAI API, specifically the "gpt-4o-2024-08-06" model. For generating back-transcription transformations we use the "gpt-4o-transcribe" and "gpt-4o-mini-tts" models available on the OpenAI API. Here are some details of the transformations in reWordBench:

\begin{tcolorbox}[azurebox]
1. Controlled Transformations: These are template-based edits that guarantee semantic equivalence by construction. They include:

\begin{enumerate}
  \item[a.] Add Quotes: Surrounding the entire prompt and responses with a fixed number of quotation marks.
  \item[b.] Punctuation Spaces: Inserting spaces around each punctuation mark.
  \item[c.] Twitter Handle/URL: Appending a randomly generated (harmless) Twitter handle or URL to the text.
  \item[d.] StressTest: Repeating semantically vacuous conjunctions (e.g.\ “and true is true” or “and false is not true”) to the end of the text.
  \item[e.] Ignore Above/Below: Injecting the response before or after the prompt with an explicit instruction to ignore it.
  \item[f.] Rot-N Encoding: Applying simple character-shift ciphers (Rot-13 or Rot-2) to the prompt text while leaving responses in plain form.
\end{enumerate}
\end{tcolorbox}

\begin{tcolorbox}[azurebox]
2. Naturalistic Transformations: These simulate the kinds of noise and variation that occur ``in the wild'' and may not perfectly preserve meaning, but reflect realistic robustness challenges:
\begin{enumerate}
    \item[a.] Paraphrase: Rewriting prompt and response via a strong LLM (Llama-3-70B-instruct) under a paraphrasing instruction.
    \item[b.] Back-translation: Translating English → Spanish → English for several rounds using OPUS-MT, accepting only those with high semantic similarity.
    \item[c.]Back-transcription: Converting text to audio and back using a TTS model (fairseq S2) and an ASR model (Whisper-base).
    \item[d.] Homoglyph Substitution: Replacing Latin characters with visually identical Unicode glyphs (e.g. Cyrillic ``e'' for Latin ``e'').
    \item[e.] Character-level Edits: Randomly swapping, inserting, deleting, or substituting characters at rates reflecting real-world typos (including QWERTY-adjacent substitutions).
    \item[f.] Word Deletion: Omitting a randomly chosen word from prompt and response, subject to a similarity filter.
\end{enumerate}
\end{tcolorbox}
\begin{tcolorbox}[azurebox]
3. Domain-Targeted Transformations: These focus on specialized subsets of RewardBench—code, mathematics, and safety prompts—where specific artifacts may bias reward models:
\begin{enumerate}
    \item[a.] Code Minification: Automatically renaming variables, removing whitespace, and otherwise ``minifying'' Python snippets without changing functionality.
    \item[b.] Add Comment: Inserting ``\# bad'' annotations after each line of chosen responses (and optionally ``\# good'' after rejected ones).
    \item[c.] Append Other Code: Concatenating the losing snippet after the winning one (and vice versa), taking advantage of Python’s return-ended semantics.
    \item[d.] Swap Format: Exchanging the usual answer formats (e.g. LaTeX  vs. markdown ``\# Answer'') in arithmetic problems.
    \item[e.] Jailbreak Prompts: Prepending known ``jailbreak'' instructions (from the ChatGPT-Jailbreak-Prompts dataset) to safety-critical queries to see if the RM prefers harmful completions.
\end{enumerate}
\end{tcolorbox}
\clearpage
\section{Experimental Setup Details}
\label{sec:experimental_details}

This appendix provides supplementary details to the experimental settings outlined in Section \ref{subsec:experimental_settings} of the main paper.

\subsection{Best-of-N Experimental Methodology}
\label{ssec:bon_pairpm_expt_methodology}

\begin{algorithm}[H]
  \caption{Best-of-$N$ Selection with Pairwise Preference Model}
  \label{alg:best_of_n_selection}
  \begin{algorithmic}[1]          
    \STATE \textbf{Input:}  Query $Q$; responses $\mathcal{A} = (A_1,\dots,A_N)$ with $N \ge 1$
    \STATE \textbf{Input:}  Pairwise model $\hat{\mathrm{R}}_\theta : (Q,A_i,A_j) \to \{1,2\}$\\
    $\triangleright$ The output $\{1,2\}$ from the Pairwise preference model indicates if the first answer is better or the second, given the query.
    \STATE \textbf{Output:} Selected best response $A_{\text{best}}$
    \STATE $A_{\text{best}} \leftarrow A_1$
    \FOR{$i \gets 2$ \TO $N$}
        \STATE $A_{\text{cand}} \leftarrow A_i$
        \IF{$\hat{\mathrm{R}}_\theta(Q, A_{\text{best}}, A_{\text{cand}}) = 2$}
            \STATE $A_{\text{best}} \leftarrow A_{\text{cand}}$
        \ENDIF
    \ENDFOR
    \RETURN $A_{\text{best}}$
  \end{algorithmic}
\end{algorithm}

For all our Best-of-N results using PairPM models, we follow a simple procedure to find the best response out of $N$ responses generated by a base LLM. In particular, PairPM models take responses 2 at a time, and provide the better response for the given query.
Given $N$ response $\mathcal{A} = (A_1,\dots,A_N)$ with $N \ge 1$, in a randomly shuffled order, we sequentially compare responses 2 at a time (starting from $A_1$ and $A_2$) using the PairPM reward model and keep track of the best response. At each iteration, the best response is compared to the next response in the list and the best response is updated. The best response after $N-1$ iterations is taken as the selected response.
The algorithm for this procedure is given in Algorithm \ref{alg:best_of_n_selection}.

\subsection{Experimental setting for Calculating Win Rates on RewardBench Prompts}
To show the performance of \carma{} on general purpose datasets, we follow reWordBench \citep{wu2025rewordbench} and use all 2985 prompts from RewardBench \citep{lambert2024rewardbench}. We use \gemmait{9} as the base model and sample N responses for each prompt in this set. Following this, we use the PairPM reward models (RM, RRM and \carma{}) to select the best response among the N responses, as described in supplementary Section \ref{ssec:bon_pairpm_expt_methodology}. We use \texttt{GPT-4} as a judge to compare \carma{}'s responses with baselines RM and RRM.

\subsection{WildGuardTest and GSM8K experimental settings}
For both WildGuardTest results (main paper Figure \ref{fig:asr_reduction_gemma9b} as well as supplementary Table \ref{tab:results_comparison_lowest_is_best}), as well as GSM8K results (main paper Figure \ref{fig:bon_gsm8k_gemma9b}), we use \gemmait{9} as the base model and sample N responses from it. Following this, we use the PairPM reward models (RM, RRM and \carma{}) to select the best response among the N responses, as described in supplementary Section \ref{ssec:bon_pairpm_expt_methodology}. For WildGuarTest, for obtaining results given the final responses, we use the WildGuard model \cite{wildguard2024} to obtain annotations for \texttt{prompt-harmfulness}, \texttt{response-harmfulness}, \texttt{response-refusal}, \texttt{is-parsing-error}, as described in the WildGuard repository\footnote{https://github.com/allenai/wildguard}. Using these annotations, we obtain ASR and RTA for \carma{} and baselines.

\subsection{Datasets and Augmentation}
\label{app:datasets_aug_details}

For human preference data ($\mathcal{D}_{\text{pref}}$) we use \textbf{Ultrafeedback} \citep{cui2023ultrafeedback}, which furnishes approximately 60,000 preference pairs across diverse domains. 

The data augmentation process, central to \carma{} (Section \ref{sec:methodology}), employs Gemini 2.0 Flash. This LLM is first used to identify $\ell=5$ principal causal attributes relevant to response quality. Subsequently, Gemini 2.0 Flash generates (a) causal upgrade/degradation pairs targeting these attributes ($\mathcal{D}_{\text{causal}}$), and (b) neutral pairs ($\mathcal{D}_{\text{neutral}}$).

The raw augmented data, $\mathcal{D}_{\text{aug}}$, undergoes a filtering step. This involves applying a model-based confidence filter, using a baseline RM (trained solely on $\mathcal{D}_{\text{pref}}$) with a threshold of $\tau=0.2$. This filtering focuses the training on more informative examples. The amplification process involves initially generating approximately 10x data from causal augmentations (5 attributes, 2 versions per original response) and 1x data from neutral augmentations, followed by verification and the confidence-based filtering. The final training dataset $\mathcal{D} = \mathcal{D}_{\text{pref}} \cup \mathcal{D}_{\text{aug\_filtered}}$ typically contains about 3.5 times the number of examples in the original $\mathcal{D}_{\text{pref}}$, similar to RRM \citep{liu2024rrm}.

\subsection{Models and Training}
\label{app:models_training_details}

\paragraph{Reward Models (RMs):} We instantiate RMs using {\qwen{}} \citep{yang2024qwen2} and {\gemmait{9}}, {\gemma{2}} \citep{team2024gemma} as base transformer architectures. Our RM variant, $\carma\text{-PairPM}$, processes inputs formatted as `Q, A, B` and predicts a preference token ('A' or 'B') via a cross-entropy loss. An alternative variant, $\carma\text{-BT}$, implements the Bradley-Terry model by deriving scalar scores for each answer.

\paragraph{Policy Models:} For downstream alignment tasks, we use the Best-of-N setup
where we generate N responses using \gemmait{9} and use \carma{} as well as baseline reward models to select the best candidate response.

\paragraph{Training Hyperparameters:} All models are trained in PyTorch with the Hugging Face Transformers library. For RM training, following \citet{liu2024rrm}, we use the AdamW optimizer \citep{loshchilov2017decoupled} for 1 epoch, with a learning rate of $1 e^{-6}$, a global batch size of 256, and a cosine learning rate schedule. We use a warmup ratio of 0.03. For training all models, we use 8 NVIDIA A100 80GB GPUs. RM training runs require time between 10-16 hours for 2B to 9B mdoels we consider.

\subsection{Baselines and Evaluation}
\label{app:baselines_ablations_eval_details}

\paragraph{Baselines:} Our full \carma{} approach is compared against two primary baselines:
\begin{enumerate}[itemsep=0pt, topsep=1pt, partopsep=0pt, leftmargin=*]
    \item A \textbf{Base RM}, trained solely on the original $\mathcal{D}_{\text{pref}}$.
    \item The \textbf{RRM Baseline} \citep{liu2024rrm}, which employs a distinct augmentation strategy using non-contextual examples and responses from different queries, not specifically aligned with identified causal or spurious attributes.
\end{enumerate}

\paragraph{Evaluation Benchmarks:}
RM quality is assessed by accuracy on \textbf{RewardBench} \citep{lambert2024rewardbench} (overall and per category: Chat, Chat-Hard, Safety, Reasoning) and robustness on \textbf{Re-word Bench} \citep{wu2025rewordbench}. BoN Policy performance is evaluated using RewardBench, WildGuardTest \citep{wildguard2024}, GSM8K \citep{cobbe2021gsm8k}.

\section{Causal Model and Augmentation Details}
\label{sec:causal_model_details}

This appendix provides further details on the causal framework underpinning \carma{} and discusses various data augmentation strategies in the context of robust reward modeling.

\subsection{Elaboration on the Causal Model}
The causal graph presented in Figure \ref{fig:causal_graph} (Section \ref{subsec:causal_graph}) models the generation of an answer $\mathrm{A}$ and the formation of its attributes. The query $\mathrm{Q}$ influences the generator's latent \textit{intent} $\mathcal{I}$. This intent, along with unobserved generator-specific confounders $\mathcal{U}$ (e.g., inherent stylistic preferences, verbosity tendencies, pre-existing biases), leads to the textual answer $\mathrm{A}$. The answer $\mathrm{A}$ then manifests both \textit{causal attributes} $\mathrm{C}(\mathrm{A})$ (e.g., factuality, relevance) and \textit{spurious attributes} $\mathrm{SP}(\mathrm{A})$ (e.g., length, specific formatting, politeness). The true, idealized reward $\mathrm{R}^*$ is assumed to be a function only of $\mathrm{Q}$ and $\mathrm{C}(\mathrm{A})$.

The challenge in training a reward model $\hat{\mathrm{R}}_\theta$ arises because $\mathrm{SP}(\mathrm{A})$ can become correlated with $\mathrm{R}^*$ in the training data. This correlation can occur if $\mathcal{U}$ influences both the choice of spurious features and the aspects that contribute to causal quality, or simply because certain spurious features happen to co-occur with preferred answers in $\mathcal{D}_{\mathrm{pref}}$. Without explicit guidance, $\hat{\mathrm{R}}_\theta$ may learn to rely on these spurious correlations, leading to reward hacking. \carma{}'s data augmentation strategy aims to provide this explicit guidance by generating new answer pairs that help $\hat{\mathrm{R}}_\theta$ disentangle $\mathrm{C}(\mathrm{A})$ from $\mathrm{SP}(\mathrm{A})$.

\subsection{\carma{}'s Causal Augmentation: Attribute Isolation}
\label{app:carma_causal_aug}
\carma{}'s primary strategy for enhancing sensitivity to causal attributes involves \textit{Attribute Upgradation/Degradation}. This generates pairs $(\tilde{\mathrm{A}}^{(C_j \leftarrow \text{upgraded/degraded})}, \mathrm{A})$ or $(\mathrm{A}, \tilde{\mathrm{A}}^{(C_j \leftarrow \text{upgraded/degraded})})$ by prompting an LLM to modify an original answer $\mathrm{A}$ (from $\mathcal{D}_{\mathrm{pref}}$) along a single causal attribute $C_j$ while attempting to keep other attributes constant. This provides a targeted signal about the marginal contribution of $C_j$.

\subsubsection{Comparison with Relevance Contrast Augmentation}
An alternative strategy, \textit{Relevance Contrast Augmentation} (used in RRM-style approaches \citep{liu2024rrm}, termed ``non-contextuals'' therein), involves pairing a relevant answer $\mathrm{A}_1$ (for query $\mathrm{Q}$) with an irrelevant answer $\mathrm{B}_2$ (e.g., an answer to a different query, so $C(B_2 \mid \mathrm{Q}) \approx \mathbf{0}$), labeled $\mathrm{A}_1 \succ \mathrm{B}_2$.

While Relevance Contrast establishes a baseline understanding of relevance, \carma{}'s Attribute Isolation offers:
\begin{itemize}[itemsep=0pt,left=8pt]
    \item \textbf{Specificity and Nuance:} It directly teaches about individual causal attributes ($C_j$), enabling the RM to learn a compositional understanding of quality and distinguish between relevant answers differing subtly in one dimension.
    \item \textbf{Data Efficiency for Complex Attributes:} Focusing changes along one attribute creates diverse, targeted examples for each quality facet.
\end{itemize}
\carma{}'s attribute-specific counterfactuals thus provide a richer, more disentangled signal than broad relevance contrasts alone.

\subsection{Neutral Augmentation Strategies}
Neutral augmentations aim to make the reward model invariant to spurious attributes when causal content is held constant or is irrelevant.

\subsubsection{Common Spurious Perturbation Methods (Not a primary \carma{} strategy)}
Several methods focus on general spurious perturbations:
\paragraph{1. Direct Spurious Feature Perturbation (e.g., Paraphrasing, Formatting Changes):}
This involves taking an answer $\mathrm{A}$ and generating $\tilde{\mathrm{A}}^{(SP \leftarrow sp')}$ by applying meaning-preserving transformations (e.g., paraphrasing) intended to alter only $\mathrm{SP}(\mathrm{A})$ while preserving $\mathrm{C}(\mathrm{A})$. The pair $(\mathrm{A}, \tilde{\mathrm{A}}^{(SP \leftarrow sp')})$ is labeled as a tie. This is central to benchmarks like reWordBench \citep{wu2025rewordbench}.

\paragraph{2. Rewrites of Rewrites (e.g., RATE \citep{reber2024rate}):}
RATE uses sequential rewrites for robust causal effect estimation. Adapted for augmentation, multiple causally-equivalent rewrites of an answer could form neutral pairs.

\textit{Challenges with these General Methods:}
\begin{itemize}[itemsep=0pt,left=8pt]
    \item \textbf{Unknown/Unspecified Spurious Features:} It's hard to a priori identify and target all spurious features an RM might exploit.
    \item \textbf{Preserving Causal Content:} Ensuring "spurious" perturbations don't inadvertently alter causal meaning is difficult.
\end{itemize}

\subsubsection{Neutral Augmentation Strategies developed in this work}
We developed the following two strategies for neutral augmentation.

\paragraph{1. Irrelevant Query Neutrals (IQN):}
\carma{} generates these neutral pairs efficiently by leveraging its existing pool of answers (original or causally augmented). Given two answers, $\mathrm{B}_1$ and $\mathrm{B}_2$, that were generated or selected for a specific query $\mathrm{Q}_{\text{orig}}$, \carma{} creates a neutral pair by associating them with a \textit{new, unrelated query} $\mathrm{Q}_{\text{irrelevant}}$.
For this $\mathrm{Q}_{\text{irrelevant}}$, both $\mathrm{B}_1$ and $\mathrm{B}_2$ are now contextually irrelevant; their causal attribute scores $C(B_1 | \mathrm{Q}_{\text{irrelevant}})$ and $C(B_2 | \mathrm{Q}_{\text{irrelevant}})$ are effectively zero (or very low). Despite potentially different spurious attributes $SP(B_1)$ and $SP(B_2)$, the pair $(\mathrm{B}_1, \mathrm{B}_2)$ is presented to the reward model with query $\mathrm{Q}_{\text{irrelevant}}$ and labeled as a tie.
This teaches the RM that when answers are equally and maximally irrelevant to the current query, their differing spurious features should not induce a preference.

\paragraph{2. Causally-Aligned Neutrals (CAN):}
This method directly leverages the original preference pairs or the outputs of causal augmentation.
\begin{itemize}[itemsep=0pt,left=8pt]
    \item Given an original preference pair from $\mathcal{D}_{\mathrm{pref}}$, say $(\mathrm{A}_1, \mathrm{A}_2)$ where $\mathrm{A}_1 \succ \mathrm{A}_2$, we generate $\tilde{\mathrm{A}}_{2}^{(C \leftarrow C(A_1))}$ by rewriting $\mathrm{A}_2$ to match the causal attribute profile of $\mathrm{A}_1$, while instructing the LLM to retain the spurious characteristics $SP(A_2)$ of the original $\mathrm{A}_2$. The pair $(\mathrm{A}_1, \tilde{\mathrm{A}}_{2}^{(C \leftarrow C(A_1))})$ is then labeled as a tie. A symmetric pair can also be generated.
    \item Similarly, if we have an answer $\mathrm{A}$ and its causally degraded version $\tilde{\mathrm{A}}^{(C_j \leftarrow \text{degraded})}$ (from $\mathcal{D}_{\mathrm{causal}}$), we can attempt to reconstruct the degraded version by prompting an LLM to restore $C_j$ to its state in $\mathrm{A}$, while aiming to preserve the spurious features of $\tilde{\mathrm{A}}^{(C_j \leftarrow \text{degraded})}$. If successful, this reconstructed version, $\tilde{\mathrm{A}}'_{\text{reconstr}}$, would form a neutral pair $(\mathrm{A}, \tilde{\mathrm{A}}'_{\text{reconstr}})$ labeled as a tie.
\end{itemize}
The core idea is to teach invariance to the spurious differences that remain \textit{after} causal attributes have been aligned or restored.
Moreover, applying CAN to counterfactually generated data from $\mathcal{D}_{\mathrm{causal}}$ helps mitigate imperfections in oracle rewrites—an issue highlighted in the RATE paper \citep{reber2024rate}, which notes that LLM edits often unintentionally modify "off-target attributes" (e.g., introducing formality, removing HTML tags). CAN thereby enhances robustness on two fronts: (1) disentangling spurious correlations in original data, and (2)  neutralizing new biases introduced during causal augmentation.
This helps in enhancing model's robustness against confounding signals in the data. While this method is sound theoretically, we qualitatively find that the approximation of $C(A_w)$ by $C(\tilde{A}_l)$ is not perfect. Furthermore, some spurious attributes $SP'(\tilde{A}_l) \subset SP(\tilde{A}_l)$ vary when we move causal attributes. Invariance to these attributes $SP'(\tilde{A}_l)$ is not captured by CAN. For these reasons, we encourage future work for improving this neutral augmentation strategy.
\section{Detailed Mechanistic View of Augmentation Strategies}
\label{app:detailed_augmentation_diagrams}

\begin{figure}[!t]
    \centering
    \includegraphics[width=0.9\linewidth]{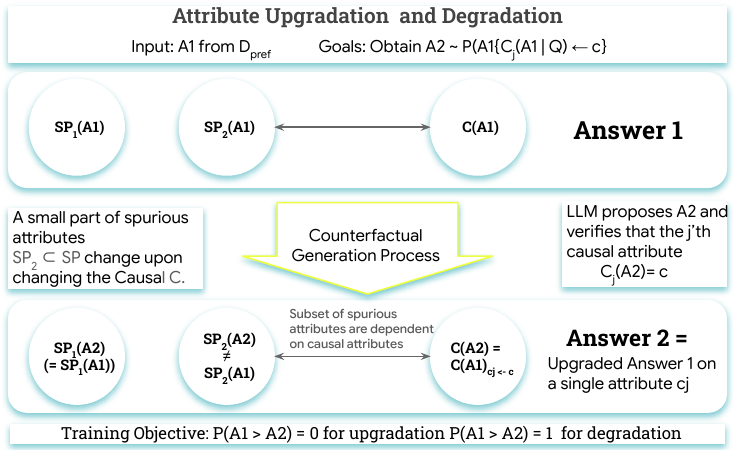}
    \caption{Detailed mechanistic diagram of \carma's Causal Attribute Upgradation and Degradation, illustrating attribute components and transformations. This causal diagram indicates that on changing causals some spurious features also can get dragged along (we call these $SP_2$). Hence separating these is very hard. This illustrates the need for a neutral augmentation strategy that provides invariance to $SP_2$ attributes.}
    \label{fig:detailed_augmentation_node_graph_appendix}
\end{figure}

This appendix section provides a more granular, node-based representation (Figure~\ref{fig:detailed_augmentation_node_graph_appendix}) to elaborate on the hypothesized attribute interactions and the counterfactual generation process. This detailed view aims to offer a causal understanding that complements the main paper.

Figure~\ref{fig:detailed_augmentation_node_graph_appendix} aims to provide a deeper, causal understanding of the causal perturbation process through which we obtain our causal upgradations and degradations. We term the spurious attributes that move when causal attributes are intervened upon as $SP_2(A) \subset SP(A)$ for any answer $A$.

\paragraph{Part 1: Causal Augmentation (Attribute Upgradation/Degradation).}
We first generate a counterfactual Answer 2 from an original Answer 1 (for query $\mathrm{Q}$) via an LLM-driven "Counterfactual Generation Process." This process intervenes to modify a specific causal attribute $C_j$ within Answer 1's causal profile $\mathrm{C(A1)}$ to a target state $C'$, resulting in $\mathrm{C(A2)}$. 
We aim to keep spurious attributes fixed by asking for a minimal perturbation. Therefore attributes $\mathrm{SP_1(A1)}$ are ideally preserved.
Yet, $\mathrm{SP_2(A1)}$ (which may co-vary with $\mathrm{C(A1)}$) might transition to $\mathrm{SP_2(A2)}\neq\mathrm{SP_2(A1)}$. The goals of this transformation are to ensure $A_2$ reflects the intended causal change. The RM is then trained on the pair $(A_1, A_2)$ with a preference label reflecting the upgrade/degradation, teaching sensitivity to isolated causal attribute modifications. 

\paragraph{Part 2: Neutral Augmentation (via Irrelevant Query).}
As illustrated in Figure \ref{fig:detailed_augmentation_node_graph_appendix}, we need spurious invariance to $SP_2$ which are hard to disentangle as well. This illustrates the need for an intervention free method for neutral augmentation like IQN. When we present an answer pair ($A_1, A_2$) from $\mathcal{D}_{\mathrm{pref}} \cup \mathcal{D}_{\mathrm{causal}}$,
re-contextualized with a new, unrelated query $\mathrm{Q}_{\text{irrelevant}}$,
we teach the model invariance to $(SP_1, SP_2)$.
This is because, the primary differences between $A_1$ and $A_2$ in this new context are their spurious attributes ($\mathrm{SP_1}$,$\mathrm{SP_2}$). 
Note that the causal difference between $A_1$ and $A_2$ in $\mathcal{D}_{\mathrm{pref}} \cup \mathcal{D}_{\mathrm{causal}}$ in presence of irrelevant query is now spurious, and hence there need not be any sensitivity to it.

\section{Detailed \carma{} Methodology}
\label{sec:detailed_methodology}

This appendix provides the detailed implementation steps for the \carma{} framework introduced in Section \ref{sec:methodology}, covering attribute identification, counterfactual data generation, filtering, and the specific training objective.

\subsection{Step 1: Attribute Identification}
\label{subsec:attribute_identification_appendix}

The foundation involves identifying the attributes that genuinely determine answer quality versus those merely correlated with it, as defined in Section \ref{subsec:causal_graph}. For a query $\mathrm{Q}$ and example answers $(\mathrm{y}_w, \mathrm{y}_l)$ from $\mathcal{D}_{\mathrm{pref}}$, we define: \textit{Causal attributes} $\mathrm{C} = (\mathrm{C}_1, \dots, \mathrm{C}_\ell)$ (e.g., factuality) and \textit{Spurious attributes} $\mathrm{SP} = (\mathrm{SP}_1, \dots, \mathrm{SP}_k)$ (e.g., verbosity).

\paragraph{Automated Attribute Extraction.} We employ an LLM prompted with $\mathrm{Q}$ and example responses (see Appendix~\ref{sec:prompt_templates} for prompt). The primary output is the set of attributes $\mathrm{C}$.

\paragraph{Refinement and Verification.} The LLM-generated list $\mathrm{C}$ is reviewed for coherence and consistency in this verification phase. The verification prompts are provided in Appendix \ref{sec:prompt_templates}.

\subsection{Step 2: Generating Counterfactual Augmented Data}
\label{subsec:counterfactual_generation_appendix}

Using identified attributes $\mathrm{C}$, we generate $\mathcal{D}_{\mathrm{aug}}$ via LLM-approximated counterfactuals (Section \ref{subsec:approximating_counterfactuals}).

\paragraph{Causal Augmentation ($\mathcal{D}_{\mathrm{causal}}$).} Pairs $(\mathrm{A}, \mathrm{A}')$ are generated to differ primarily along a single causal attribute $\mathrm{C}_j$. We use LLM prompts (Appendix~\ref{sec:prompt_templates}) for \textit{upgradation} (generating an improved $\mathrm{A}'$ from a ground-truth rejected answer $\mathrm{A}$) and \textit{degradation} (generating a degraded $\mathrm{A}'$ from a ground-truth selected answer $\mathrm{A}$), aiming to keep other attributes constant. Pairs are labeled $\succ$ accordingly.

\paragraph{Neutral Augmentation ($\mathcal{D}_{\mathrm{neutral}}$).} 
Notice that when we causally augment an answer in $\mathcal{D}_{\mathrm{causal}}$, we might in-advertantly move spurious correlates (as illustrated in Figure \ref{fig:carma_augmentation_visual_overview}). Furthermore, even in our dataset, there could be a systematic effect where spurious attributes highly correlate with the better (or worse) answer. In such cases, we need to create a dataset of equivalent pairs, with a tie label to teach the model invariance to spurious correlates.

Our primary technique is \textit{irrelevant query neutrals} (IQN). Here, the idea is that given a new query, the causal attribute $\mathrm{C}$ becomes irrelevant. Essentially, for the new query, the causal attributes are spurious. Hence, by taking any two answers for a given query, and labeling them a tie, given an irrelevant query, the reward model learns invariance to these features. For example, if the reward model has spuriously learnt that bullet points in an answer should be rewarded, our tie labels teach them that bullet points should be rewarded only if the content of the answer is relevant to the query.
Similarly, by creating such pairs with our own causally augmented data in $\mathcal{D}_{\mathrm{causal}}$, we teach the model invariance to the spurious pairs that move when the causal attributes (CA) are perturbed.

\subsection{Step 3: Filtering Augmented Data}
\label{subsec:filtering_appendix} 

The raw $\mathcal{D}_{\mathrm{aug}}$ is then filtered to $\mathcal{D}_{\mathrm{aug\_filtered}}$.

\paragraph{Model-based Confidence Filtering.} Using a baseline $\hat{\mathrm{R}}_{\mathrm{base}}$, we calculate $p = \mathrm{P}_{\mathrm{base}}(\mathrm{B} \succ \mathrm{A})$ for each augmented pair $(\mathrm{A}, \mathrm{B})$ with target label $y$. We retain the pair only if $|p - \mathbb{I}(y = \text{B} \succ \mathrm{A}) - 0.5 \cdot \mathbb{I}(y = \text{tie})| > \tau$. We use threshold $\tau=0.2$, focusing training on examples where the baseline is uncertain or incorrect \citep{liu2024rrm}.

\paragraph{Quality Verification.} Further checks (e.g., automated fluency scoring) verify pair validity. The result is $\mathcal{D}_{\mathrm{aug\_filtered}}$.

\subsection{Step 4: Training the Robust Reward Model}
\label{subsec:training_appendix} 

The final model $\hat{\mathrm{R}}_\theta$ is trained on $\mathcal{D} = \mathcal{D}_{\mathrm{pref}} \cup \mathcal{D}_{\mathrm{aug\_filtered}}$ by minimizing the composite loss:
\begin{align}
\mathcal{L}(\theta) = &- \sum_{(\mathrm{Q}, \mathrm{y}_w, \mathrm{y}_l) \in \mathcal{D}_{\mathrm{pref}} \cup \mathcal{D}_{\mathrm{causal}}} \log \sigmoid(\hat{\mathrm{R}}_\theta(\mathrm{Q}, \mathrm{y}_w) - \hat{\mathrm{R}}_\theta(\mathrm{Q}, \mathrm{y}_l)) \nonumber \\
&- \lambda \sum_{(\mathrm{Q}, \mathrm{A}_1, \mathrm{A}_2, y=\text{tie})\in \mathcal{D}_{\mathrm{neutral}}} \mathcal{L}_{\mathrm{tie}}(\theta; \mathrm{Q}, \mathrm{A}_1, \mathrm{A}_2)
\label{eq:combined_loss_appendix}
\end{align}
where $\mathcal{L}_{\mathrm{tie}}$ is defined as in Eq. \ref{eq:combined_loss_methodology}. The hyperparameter $\lambda \ge 0$ weights the neutral tie loss and is tuned on a validation set (Section \ref{sec:experiments}).
\section{Theoretical Analysis}
\label{sec:theoretical_analysis_detailed}

In this section, we provide a formal justification for why the \carma{} training framework, specifically the composite loss function operating on causally augmented data, mitigates spurious reward hacking. We demonstrate that the optimization objective inherently discourages the reward model from relying on spurious correlations, guiding it towards the true causal drivers of quality.

\subsection{Formal Setup}
\label{subsec:theory_setup}

We adopt the notation and causal framework established in Section \ref{sec:preliminaries}. Our analysis considers a query $\mathrm{Q}$, an answer $\mathrm{A}$ with corresponding Principal Causal Components $\mathrm{C}(\mathrm{A})$ and spurious attributes $\mathrm{SP}(\mathrm{A})$. The idealized ground-truth reward is $\mathrm{R}^*(\mathrm{Q}, \mathrm{A}) = f^*(\mathrm{Q}, \mathrm{C}(\mathrm{A}))$, and the learned reward model is denoted $\hat{\mathrm{R}}_{\theta}(\mathrm{Q}, \mathrm{A})$. The model parameters $\theta$ are optimized by minimizing the composite loss function $\mathcal{L}(\theta) = \mathcal{L}_{\mathrm{pref}}(\theta) + \lambda \mathcal{L}_{\mathrm{tie}}(\theta)$ (Eq. \ref{eq:combined_loss_methodology}) over the training dataset $\mathcal{D} = \mathcal{D}_{\mathrm{pref}} \cup \mathcal{D}_{\mathrm{aug\_filtered}}$, which combines original preferences $\mathcal{D}_{\mathrm{pref}}$ with filtered causal $\mathcal{D}_{\mathrm{causal}}$ and neutral $\mathcal{D}_{\mathrm{neutral}}$ augmentations. For theoretical analysis, $\mathcal{L}_{\mathrm{pref}}$ and $\mathcal{L}_{\mathrm{tie}}$ represent expectations over the respective data distributions:
\begin{align*}
\mathcal{L}_{\mathrm{pref}}(\theta) &= - \mathbb{E}_{(\mathrm{Q}, \mathrm{y}_w, \mathrm{y}_l) \sim \mathcal{D}_{\mathrm{pref}} \cup \mathcal{D}_{\mathrm{causal}}} \left[ \log \sigmoid(\hat{\mathrm{R}}_\theta(\mathrm{Q}, \mathrm{y}_w) - \hat{\mathrm{R}}_\theta(\mathrm{Q}, \mathrm{y}_l)) \right] \\
\mathcal{L}_{\mathrm{tie}}(\theta) &= - \mathbb{E}_{(\mathrm{Q}, \mathrm{A}_1, \mathrm{A}_2, y=\text{tie}) \sim \mathcal{D}_{\mathrm{neutral}}} \left[ -\frac{1}{2} \left( \log \sigmoid(\Delta_{12}) + \log \sigmoid(-\Delta_{12}) \right) \right]
\end{align*}
where $\Delta_{12} = \hat{\mathrm{R}}_\theta(\mathrm{Q}, \mathrm{A}_1) - \hat{\mathrm{R}}_\theta(\mathrm{Q}, \mathrm{A}_2)$.

\subsection{Justification under the Boolean variable causal model for attributes}

\begin{assumption}\label{assum:model}
Assume that:
\begin{enumerate}
\item Causal attributes $\{C_i(Q,A)\}_{i=1}^k$
and spurious attributes $\{ S_j(A)\}_{j=1}^{\ell}$ are all boolean variables taking values in $\{+1,-1\}$

\item All spurious variables are non-descendants of all causal variables.

\item Reward function is trying to fit a quadratic polynomial in causal and spurious attributes, i.e. 
\begin{align}\label{eq:quad_model}
\hat{R} & = \sum_i \alpha_i C_i(Q,A) + \sum_j \beta_j S_j(A) + \sum \limits_{i \neq i'} \alpha_{i,i'} C_i(Q,A) C_{i'}(Q,A) + \nonumber \\
\hfill & \sum \limits_{j \neq j'} \beta_{j,j'} S_j(A) S_{j'}(A) + \sum \limits_{i \neq j} \gamma_{i,j} C_i(Q,A) S_j(A) .
\end{align}
\item Assume that the true reward function is a sparse quadratic polynomial depend on only the causal attributes.
 \begin{align}\label{eq:true_model}
R^{*} & = \sum_i \theta_i C_i(Q,A) + \sum \limits_{i \neq i'} \theta_{i,i'} C_i(Q,A) C_{i'}(Q,A) 
\end{align}

Here, $\lVert \mathbf{\theta} \rVert_0 \leq s << k^2$ and $\theta_i$ and $\theta_{i,i'}$ variables form the vector $\mathbf{\theta}$. All other coefficients for other features that involves the spurious variables are set to $0$ in $\theta$. Let ${\cal I}$ be the support set of the true coefficient.

\end{enumerate}
\end{assumption}

From the reward modeling objective, we try to fit a model $\Delta(\hat{R}) $ to a target which is the difference between true rewards to two answers $A_1$ and $A_2$ for the same question, i.e. $R^{*}(Q,A_1) - R^{*}(Q,A_2)$. From the assumption in \ref{eq:quad_model}, this is equivalent to fitting a linear model with coefficients $\alpha_i,\alpha_{i,i'}, \beta_{j}, \beta_{j,j'}, \gamma_{i,j}$  and differences in features (across the two answers), i.e. $C_i(Q,A_1)- C_i(Q,A_2), S_j(A_1)- S_j(A_2), S_j(A_1) S_{j'}(A_1)- S_j(A_2) S_{j'}(A_2), C_i(Q,A_1) C_{i'}(Q,A_1)- C_i(Q,A_2) C_{i'}(Q,A_2),  C_i(Q,A_1) S_{j}(A_1)- C_{i'}(Q,A_2) S_j(A_2)$ respectively. To simplify notation, we drop the reference to $A_1, A_2$ and $Q$ and call $C_i(Q,A_1)- C_i(Q,A_2)$ as $\Delta C_i$. Similarly, we use $\Delta S_j, \Delta C_{i,i'}, \Delta S_{j,j'}$  and $\Delta (C_iS_j)$. The dependence of these features on the $A_1,A_2$ and $Q$ are understood. 

Let $F_{q,a_1,a_2} \in \{+1,-1\}^{k + \ell + k \ell + \binom{k}{2} + \binom{\ell}{2}}$ be the boolean vector with features 

$ \{\Delta C_i\} , \{\Delta S_j\}, \{\Delta C_{i,i'} \}, \{\Delta S_{j,j'}\}, \{\Delta (C_iS_j)\}$ stacked row wise for the triplet $q,a_1,a_2$.

Consider two types of triplets, one drawn from the natural distribution of the preference training dataset $D_{\mathrm{pref}}$ and the others drawn from augmented distribution $D_{\mathrm{aug}}$. Let us assume for the sake of the theoretical results to follow, that we upgrade/degrade answer $a_2$ to $a^{aug}_1$ by changing only \textit{one causal factor at a time while all the other causal factors are fixed to their factual version and all things remaining the same} to form $D_{\mathrm{aug}}$. The degradation aspect only serves to reinforce the phenomenon we seek to show formally below.

\begin{assumption}(Model for Counterfactual Generation)\label{assum:gen}

We assume that: 
\begin{enumerate}
\item $a^{aug}_1$ is formed by generating $C_i(Q,A)$ and $S_j(A)$ following an counterfactual generation where the following set of intervention is made  $C_i(Q,A) \leftarrow   \neg C_i(Q,A), ~ C_{j}(Q,A) \leftarrow C_j(Q,A),~ \forall j \neq i$ which propagates to potential descendants of variable $C_i$ and not affecting $S_j$ (due to no $S_j$ being a descendant of $C_j$) with all other factors remaining as in answer $a_2$. 
\item Let us assume that we have $m$ augmentations where a triplet is randomly sampled from the training preference data distribution ${\cal D}_{\mathrm{pref}}$ and then augmented using the above counterfactual with a randomly chosen causal attribute negated.
\end{enumerate}
\end{assumption}

\textbf{Remark} There are the main assumptions - 1) $S_j$ being a non-descendant of $C_i$, 2) Reward model is a quadratic sparse boolean model (The treatment could be extended to boolean polynomials of higher degree too with lot more algebraic technical work).

\begin{theorem}
\label{thm:appendix_lasso_recovery}
  Let the feature matrix of the counterfactually augmented triplets, that is formed by stacking feature vectors $F_{q,a^\mathrm{aug}_1,a_2}$ row wise, be denoted $\mathbf{F}$. Consider the following $\ell_1$ constrained regression problem:
  \begin{align}
   \hat{\mathbf{\theta}} = \arg \min \limits_{\mathbf{b}} \lVert \mathbf{b} \rVert_1~s.t. \mathbf{F}b = \Delta R^{*}
  \end{align}    

Here, $\Delta R^{*}$ is vector of the difference in the true reward between the reward applied to the augmented answer and the non-augmented one across augmented triplets. Let ${\cal N}$ be the top $c_2 k$ non zero entries of vector $\mathbf{a}$ by magnitude. Then, we have:

 $\lVert \Delta \mathbf{\theta} \rVert_2  = \lVert \mathbf{\theta} - \hat{\mathbf{\theta}} \rVert_2 \leq c_3 \lVert \theta_{{\cal I}-{\cal N}} \rVert_1 \left( \frac{4}{k} +  \sqrt{\frac{8 \log(k+\ell)}{m}} \right)  $ w.h.p.
\end{theorem}

\textbf{Remark:} If the true sparsity $s < c_2 k$, then it ensures perfect recovery since ${\cal I}- {\cal N} = \emptyset$. Since $s < k^2$, and if every coefficient is $O(1)$, the bound becomes $O(k)$ which is independent of the spurious dimension.
 \begin{proof} 
  Under the model assumptions \ref{assum:model} and assumptions on counterfactual generation \ref{assum:gen}, we seek to show that $\mathbf{F}$ when restricted to feature set $\Delta C_i, \Delta C_{i,i'}, \Delta_{C_iS_j}$  has smaller incoherence (by multiplicative factor of $k$) than an feature matrix made of i.i.d triplets sampled  from the preference distribution. This accommodates recovering the $s=O(k)$ sparse solutions exactly and in the general case, the error in coefficient estimation is $O(k)$ independent of spurious dimension $\ell$. 

  First, we show that features $\Delta(S_{j,j'})=0, \Delta(S_j)=0$ for the augmented triplets. This is because all $S_j$ variables are ancestors to $C_i$ variables. Therefore, a counterfactual intervention on the answer $a_2$ leaves the two spurious attribute sets (for the original and its counterfactual) unchanged. 

  Intervention fixed all causal variables to the factual ones (but fixed through intervention) and intervenes on variable to change. There are many types of correlation between non zero features because of this. We consider them one by one:

  1) $\Delta C_i =0$ if is $C_i$ is not intervened. This occurs with probability $1-1/k$. 
  2) $\Delta C_i \Delta C_j = 0$ with probability $1-2/k$.
  3) $\Delta C_{i,i'} \Delta C_{j,j'} = 0$ if all $i,i',j,j'$ are distinct indices. 
  4) $\Delta C_{i,j} \Delta C_{j,k} = 0$, with probability $1-1/k$. 
  5) $\Delta C_{i,j} \Delta C_iS_j = 0 $ with probability $1-1/k$.
  6) $\Delta C_{i,i'} \Delta C_jS_k = 0 $ always if all four indices not equal.
  7) $\Delta C_{i} \Delta C_jS_k = 0 $ always. 
  8) $\Delta C_i \Delta C_i S_k  = 0$ with probability 1-1/k. 

If any of the these products is non zero, conditioned on that event, they equal the correlation on the preference training dataset (every correlation between features is bounded by at most $4$).

Therefore, expected pairwise correlation amongst two features for a randomly chosen augmented triple is at most $4/k$. Given every augmented triple is obtained by counterfactual generation applied to an i.i.d sample from preference dataset, there is a deviation of at most $\frac{8 \log (k+ \ell)}{ \sqrt{m}}$ with probability $1- \frac{1}{(k+l)^4}$.

Therefore,
\begin{align}
  \lVert 
  \frac{1}{m}\mathbf{F}^T  \mathbf{F} - \mathbf{I}\rVert_{\infty} \leq \frac{4}{k} + \frac{8 \log (k+\ell) }{\sqrt{m}} ~ w.p. ~ 1- (k+\ell)^{-4}
\end{align}

This means that the data matrix is incoherent with high probability. We now follow standard Lasso analysis. Recall the \textit{cone condition \cite{negahban2009unified}}: For a subset ${\cal N}$ of indices that have non zero values in $\mathbf{\theta}$, $ \lVert \Delta \mathbf{\theta}_{N^c} \rVert_1 \leq \lVert \Delta \mathbf{\theta}_{N} \rVert_1  + 2 \lVert \mathbf{\theta}_{N^c}  \rVert_1$. This implies:
\begin{align}\label{eq:cone}
\lVert \Delta \mathbf{\theta} \rVert_1 \leq 2 \sqrt{|N|} \lVert \Delta \mathbf{\theta} \rVert_2 + 2 \lVert \mathbf{\theta}_{N^c}  \rVert_1.
\end{align}

We have the following chain:
\begin{align}\label{incoherence}
0 = \frac{1}{m} \lVert A \Delta \mathbf{\theta} \rVert^2 & \geq  \lVert \Delta \mathbf{\theta} \rVert_2^2  -  \lVert \Delta \mathbf{\theta} \rVert_1^2 \lVert \frac{1}{m}\mathbf{F}^T  \mathbf{F} - \mathbf{I} \rVert_{\infty} \nonumber \\
\hfill & \geq \lVert \Delta \mathbf{\theta} \rVert_2^2  -  \left( 8 |{\cal N}| \lVert \Delta \mathbf{\theta} \rVert_2^2  - 8 \lVert  \mathbf{\theta}_{{\cal N}^c} \rVert_1^2 \right)\lVert \frac{1}{m}\mathbf{F}^T  \mathbf{F} - \mathbf{I} \rVert_{\infty}
\end{align}

Let us set $m$ such that $\frac{1}{\sqrt{m}} < \frac{8 \log(k+\ell)}{k}$
. Let $|N| < k/80$, $c_2 <1/80$ in the theorem. Substituting these parameters in \ref{incoherence}, we get:
$ \lVert \Delta \mathbf{\theta} \rVert_2  = \lVert \mathbf{\theta} - \hat{\mathbf{\theta}} \rVert_2 \leq 4\sqrt{2} \lVert \theta_{{\cal I}-{\cal N}} \rVert_1 \left( \frac{4}{k} +  \sqrt{\frac{8 \log(k+\ell)}{m}} \right)  $

\end{proof}
\vspace{-0.6in}
\section{Qualitative Example Walkthrough} 
\label{sec:qualitative_example_walkthrough}

In this section, we show qualitative examples of augmented data created from original data using which is used to train \carma{}.

\begin{promptbox}{User Prompt}
Question: \textbf{Can you provide a table showing two potential causes for global warming?}

\end{promptbox}

\begin{promptbox}{Chosen Response}
Sure, here is a table showing two potential causes for global warming:
\vspace{0.15in}
\begin{center}
\begin{tabular}{|p{0.30\textwidth}|p{0.62\textwidth}|}
\hline
\textbf{Potential Causes} & \textbf{Description}\\\hline
Greenhouse Gas Emissions &
Release of gases such as carbon dioxide, methane, and nitrous oxide from human activities like burning fossil fuels, deforestation, and agriculture. These gases trap heat in the atmosphere, causing the planet to warm.\\\hline
Land Use Changes &
Alteration of the earth's surface for human activities like urbanization, agriculture, and forestry can result in changes in albedo, water cycling, and energy balance, leading to warming effects. For example, when forests are cleared, the land reflects less sunlight back into space, which leads to more warming.\\\hline
\end{tabular}
\end{center}

It is important to note that global warming is a complex issue with multiple causes. The two potential causes listed above are just a few of the many factors that contribute to climate change.
\end{promptbox}

\begin{promptbox}{Rejected Response}
\begin{center}
\begin{tabular}{|p{0.30\textwidth}|p{0.62\textwidth}|}
\hline
\textbf{Potential Causes} & \textbf{Description}\\\hline
Greenhouse Gas Emissions &
Release of gases such as carbon dioxide, methane, and nitrous oxide from human activities like burning fossil fuels, deforestation, and agriculture.\\\hline
Land Use Changes &
Alteration of the earth's surface for human activities like urbanization, agriculture, and forestry can result in changes in albedo, water cycling, and energy balance, leading to warming effects.\\\hline
\end{tabular}
\end{center}

See Table~1 for additional information.
\end{promptbox}

\begin{metricsbox}
\begin{tabular}{@{}lp{0.6\linewidth}@{}}
Accuracy      & \textbf{0.3}\\
Completeness  & \textbf{0.25}\\
Clarity       & \textbf{0.2}\\
Format        & \textbf{0.15}\\
Conciseness   & \textbf{0.1}\\
\end{tabular}
\end{metricsbox}

\clearpage

\begin{promptbox}{Causal Attributes with Elements:}
\begin{lstlisting}[
    language=json,
    numbers=left,
    numberstyle=\tiny\color{gray},
    numbersep=5pt,
    xleftmargin=1em,
    breaklines=true
]
{"causal_elements":
    "Accuracy": [
        {
            "element": "Providing scientifically accurate descriptions of the causal mechanisms by which greenhouse gas emissions lead to global warming (e.g., trapping heat)",
            "impact": "increases Accuracy"
        },
        {
            "element": "Providing scientifically accurate descriptions of the causal mechanisms by which land use changes lead to global warming (e.g., altering albedo)",
            "impact": "increases Accuracy"
        },
        {
            "element": "Including irrelevant or factually incorrect details in the descriptions of the causes.",
            "impact": "decreases Accuracy"
        },
        {
            "element": "Omitting key details or causal links in the explanation of how the causes contribute to global warming.",
            "impact": "decreases Accuracy"
        },
        {
            "element": "Presenting information suggesting a single cause when the phenomenon has multiple contributors",
            "impact": "decreases Accuracy"
        }
    ],
    "Completeness": [
        {
            "element": "Providing a mechanism by which each potential cause contributes to global warming",
            "impact": "Increases Completeness because it explains *how* the causes lead to the effect, rather than simply stating the cause."
        },
        {
            "element": "Including specific examples to illustrate the effects of the land use changes",
            "impact": "Increases Completeness by providing concrete instances that support the description of a potential cause."
        },
        {
            "element": "Acknowledging the multifactorial nature of global warming and that the listed causes are not exhaustive",
            "impact": "Increases Completeness by providing appropriate context and preventing the impression of a single, simple answer to a complex problem."
        },
        {
            "element": "Providing a table with potential causes and descriptions",
            "impact": "Increases Completeness because the response directly provides the information requested in the question."
        },
        {
            "element": "Omitting crucial details or explanations about the causes, assuming the user has prior knowledge",
            "impact": "Decreases Completeness, as the answer requires additional, unstated information to be fully understood."
        }
    ],
    "Clarity": [
        {
            "element": "Providing specific examples related to the described cause.",
            "impact": "Increases Clarity by illustrating the abstract description with concrete instances, making the explanation more understandable."
        },
        {
            "element": "Explicitly stating the mechanism by which each cause contributes to global warming.",
            "impact": "Increases Clarity by directly linking the cause to its effect on global warming, removing ambiguity about the causal relationship."
        },
        {
            "element": "Omitting relevant contextual information that provides background or rationale.",
            "impact": "Decreases Clarity because the information vacuum leaves the user to fill in the gaps, potentially leading to misunderstanding."
        },
        {
            "element": "Including references to external and unavailable sources (e.g., \"See Table 1\").",
            "impact": "Decreases Clarity by creating an incomplete picture. The reader's understanding is hindered by the inability to access referenced information."
        },
        {
            "element": "Acknowledging the complexity of the issue and the existence of other contributing factors.",
            "impact": "Increases Clarity by setting appropriate expectations and preventing the reader from oversimplifying the causes of global warming."
        }
    ],
    "Format": [
        {
            "element": "Presence of a clearly defined table with appropriate column headers ('Potential Causes' and 'Description')",
            "impact": "Increases Format by structuring the information in an easily digestible manner."
        },
        {
            "element": "Use of appropriate table delimiters and alignment (e.g., '|', '---') to create a visually distinct table.",
            "impact": "Increases Format by visually structuring and separating table elements, enhancing readability."
        },
        {
            "element": "Absence of extraneous references to external resources (e.g., 'See Table 1')",
            "impact": "Increases Format by preventing fragmentation of the answer and ensuring all required information is self-contained within the response."
        },
        {
            "element": "Inclusion of a concise introductory sentence acknowledging the question and introducing the table.",
            "impact": "Increases Format by providing context and signaling the beginning of the structured answer."
        },
        {
            "element": "Adherence to a consistent and readable text formatting style within the table cells.",
            "impact": "Increases Format by promoting clarity and ease of understanding of the information presented in the table."
        }
    ],
    "Conciseness": [
        {
            "element": "Addition of unnecessary introductory or concluding phrases",
            "impact": "Decreases Conciseness"
        },
        {
            "element": "Omission of essential details or context",
            "impact": "Increases Conciseness"
        },
        {
            "element": "Inclusion of redundant information",
            "impact": "Decreases Conciseness"
        },
        {
            "element": "Use of simpler, direct vocabulary",
            "impact": "Increases Conciseness"
        },
        {
            "element": "Presence of tangential or irrelevant information",
            "impact": "Decreases Conciseness"
        }
    ]
}
\end{lstlisting}
\end{promptbox}

\clearpage
\begin{promptbox}{Chosen Response Edits (degradations to chosen responses along all attributes)}
\begin{lstlisting}[
    language=json,
    numbers=left,
    numberstyle=\tiny\color{gray},
    numbersep=5pt,
    xleftmargin=1em,
    breaklines=true
]
"chosen_response_edits": {
    "Accuracy": "Sure, here is a table showing two potential causes for global warming:\n\nPotential Causes | Description |\n|---|---|\n| Greenhouse Gas Emissions | Release of gases such as carbon dioxide, methane, and nitrous oxide from human activities like burning fossil fuels, deforestation, and agriculture. These gases make the sky brighter, causing the planet to warm. |\n| Land Use Changes | Alteration of the earth's surface for human activities like urbanization, agriculture, and forestry can result in changes in the color of the ground, which leads to warming effects. For example, when forests are cleared, the land becomes smoother, which leads to more warming. |\n\nIt is important to note that global warming is a simple issue with one major cause. The two potential causes listed above are really the same thing.",
    
    "Completeness": "Sure, here is a table showing two potential causes for global warming:\n\nPotential Causes | Description |\n|---|---|\n| Greenhouse Gas Emissions |  |\n| Land Use Changes |  |\n\nIt is important to note that global warming is a complex issue with multiple causes. The two potential causes listed above are just a few of the many factors that contribute to climate change.",
    
    "Clarity": "Sure, here is a table showing two potential causes for global warming:\n\nPotential Causes | Description |\n|---|---|\n| Greenhouse Gas Emissions | Release of gases such as carbon dioxide, methane, and nitrous oxide from human activities like burning fossil fuels, deforestation, and agriculture. |\n| Land Use Changes | Alteration of the earth's surface for human activities like urbanization, agriculture, and forestry can result in changes in albedo, water cycling, and energy balance. |\n\nIt is important to note that global warming is a complex issue with multiple causes. The two potential causes listed above are just a few of the many factors that contribute to climate change.",
    
    "Format": "Sure, here is a table showing two potential causes for global warming:\n\nGreenhouse Gas Emissions Release of gases such as carbon dioxide, methane, and nitrous oxide from human activities like burning fossil fuels, deforestation, and agriculture. These gases trap heat in the atmosphere, causing the planet to warm. Land Use Changes Alteration of the earth's surface for human activities like urbanization, agriculture, and forestry can result in changes in albedo, water cycling, and energy balance, leading to warming effects. For example, when forests are cleared, the land reflects less sunlight back into space, which leads to more warming.\n\nIt is important to note that global warming is a complex issue with multiple causes. The two potential causes listed above are just a few of the many factors that contribute to climate change.",
    
    "Conciseness": "Sure, I can definitely help you with that! Here is a table showing two potential causes for global warming. It's important to understand these causes to get a better grasp of the overall issue:\n\nPotential Causes | Description |\n|---|---|\n| Greenhouse Gas Emissions | Release of gases such as carbon dioxide, methane, and nitrous oxide from human activities like burning fossil fuels, deforestation, and agriculture. These gases trap heat in the atmosphere, causing the planet to warm. This is because they absorb infrared radiation, preventing it from escaping into space. Therefore, the earth warms up. |\n| Land Use Changes | Alteration of the earth's surface for human activities like urbanization, agriculture, and forestry can result in changes in albedo, water cycling, and energy balance, leading to warming effects. For example, when forests are cleared, the land reflects less sunlight back into space, which leads to more warming. This is due to the fact that trees absorb sunlight, and when they're gone, that sunlight is reflected back, causing warming. |\n\nIt is important to note that global warming is a complex issue with multiple causes, as I mentioned earlier. The two potential causes listed above are just a few of the many, many factors that contribute to climate change and its associated effects. There are other causes as well. So, basically, global warming has many causes."
}
\end{lstlisting}
    
\end{promptbox}

\begin{promptbox}{Rejected Response Edits (upgradation to rejected responses along all attributes)}
    \begin{lstlisting}[
    language=json,
    numbers=left,
    numberstyle=\tiny\color{gray},
    numbersep=5pt,
    xleftmargin=1em,
    breaklines=true
]
"rejected_response_edits": {
    "Accuracy": "| Potential Causes | Description |\n| -----------------|-------------|\n| Greenhouse Gas Emissions | Release of gases such as carbon dioxide, methane, and nitrous oxide from human activities like burning fossil fuels, deforestation, and agriculture. These gases trap heat in the atmosphere, leading to a warming effect (the greenhouse effect).|\n| Land Use Changes | Alteration of the earth's surface for human activities like urbanization, agriculture, and forestry can result in changes in albedo, with darker surfaces absorbing more solar radiation than lighter ones, leading to increased heat retention. These changes also impact the water cycle, reducing evaporation and transpiration, which can further contribute to warming. |",
    
    "Completeness": "| Potential Causes | Description |\n| -----------------|-------------|\n| Greenhouse Gas Emissions | Release of gases such as carbon dioxide, methane, and nitrous oxide from human activities like burning fossil fuels, deforestation, and agriculture. These gases trap heat in the atmosphere, leading to a rise in global temperatures. |\n| Land Use Changes | Alteration of the earth's surface for human activities like urbanization, agriculture, and forestry can result in changes in albedo, water cycling, and energy balance, leading to warming effects. For instance, deforestation reduces the earth's capacity to absorb carbon dioxide. |\n\nIt is important to note that global warming is a complex issue with multiple contributing factors, and this table only lists two potential causes.",
    
    "Clarity": "| Potential Causes | Description |\n| -----------------|-------------|\n| Greenhouse Gas Emissions | Release of gases such as carbon dioxide, methane, and nitrous oxide from human activities like burning fossil fuels, deforestation, and agriculture. These gases trap heat in the atmosphere, leading to a rise in global temperatures. |\n| Land Use Changes | Alteration of the earth's surface for human activities like urbanization, agriculture, and forestry can result in changes in albedo (reflectivity), water cycling, and energy balance. For example, deforestation reduces the earth's ability to absorb carbon dioxide and alters local weather patterns which can contribute to warming. |",
    
    "Format": "Here are two potential causes for global warming in a table:\n\n| Potential Causes | Description |\n| -----------------|-------------|\n| Greenhouse Gas Emissions | Release of gases such as carbon dioxide, methane, and nitrous oxide from human activities like burning fossil fuels, deforestation, and agriculture. |\n| Land Use Changes | Alteration of the earth's surface for human activities like urbanization, agriculture, and forestry can result in changes in albedo, water cycling, and energy balance, leading to warming effects. |",
    
    "Conciseness": "Greenhouse Gas Emissions | Release of gases such as carbon dioxide, methane, and nitrous oxide from human activities like burning fossil fuels, deforestation, and agriculture. |\nLand Use Changes | Alteration of the earth's surface for human activities like urbanization, agriculture, and forestry can result in changes in albedo, water cycling, and energy balance, leading to warming effects. |"
}
\end{lstlisting}

\end{promptbox}

\begin{promptbox}{Verification Verdicts}
\begin{lstlisting}[
    language=json,
    numbers=left,
    numberstyle=\tiny\color{gray},
    numbersep=5pt,
    xleftmargin=2em,
    breaklines=true
]
"verification_results_upgradations": {
    "Accuracy":       "Pass",
    "Completeness":   "Pass",
    "Clarity":        "Pass",
    "Format":         "Pass",
    "Conciseness":    "Fail"
},

"verification_results_degradations": {
    "Accuracy":       "Pass",
    "Completeness":   "Pass",
    "Clarity":        "Pass",
    "Format":         "Pass",
    "Conciseness":    "Pass"
}
\end{lstlisting}
\end{promptbox}
\section{Prompt Templates}
\label{sec:prompt_templates}

This section details the prompt templates used for identifying attributes and generating counterfactual examples in the \carma{} framework. Placeholders like \texttt{\{question\}} are replaced with actual content during the process.

\subsection{Prompt for Attribute Identification}

\begin{promptbox}{Identifying Causal Attributes\\\\}
You are a reward model which means you have to rate answers for a given question across multiple different attributes.
The first step is to identify these attributes as well as give an importance score between 0 and 1 for all these attributes, based on how important they are for rating a response for that question.
The importance score for all attributes should sum up to 1.

The following is a Question and 2 Candidate Answer for it.\\

Question: {question}\\\\
Example Answer 1: {answer1}\\
Example Answer 2: {answer2}\\

Task: Give me 5 **mutually exclusive** and important attributes that are required to rate an answer for the give question holistically, along with their importance score. These important attributes should be independent of each other, and should largely depend on the Question given above.\\
\\
Answer Format: Give your answer in JSON format, for example:\\
\\
\{\\
 Attributes: \{\\
 "attribute\_1": attribute\_1\_score,\\
 "attribute\_2": attribute\_2\_score,\\
 "attribute\_3": attribute\_3\_score,\\
 "attribute\_4": attribute\_4\_score,\\
 "attribute\_5": attribute\_5\_score\\
 \}\\
\}
\\
Where attribute\_i is the name of the i'th attribute, attribute\_i\_score is the importance score of the i'th attribute, and the Key "Attributes" is a fixed constant string you should output.\\
\\
Summation of attribute\_i\_score across all i's should be 1.\\
\\
Strictly adhere to the format and only give the json string as output (i.e. start with "{" and end your response with "}"). Do not include any commentary, explanations, chattiness, any extra words, or additional keys outside of the specified JSON structure.\\
\\
Answer:
\end{promptbox}

\clearpage
\subsection{Prompt for Identifying Causal Elements}
\vspace{-0.15in}
\begin{promptbox}{Identifying Causal Elements per Attribute}
You are an expert in causal reasoning and response evaluation.\\\\
\\
You are given:\\\\
- A question\\
- Two example answers\\\\
\\
Your task is to identify generalizable causal elements that directly affect the strength of the attribute "\{attribute\}" in a response to the given question.\\\\
\\
The two example answers are provided to help you understand how the attribute manifests in this specific context. Do not restrict your analysis to these examples—use them only to inform your understanding of the attribute in this setting.\\\\
\\
Question: \{question\}\\\\
Accepted Answer: \{answer1\}\\\\
Rejected Answer: \{answer2\}\\\\
\\
\#\#\# Instructions:\\\\
- Identify exactly five causal elements that impact \{attribute\} in the response.\\
- Each element must have a clear role in either increasing or decreasing \{attribute\}. Clearly explain its direct causal impact on \{attribute\}.\\
- Do not include any non-causal heuristics.\\
- Do not include unnecessary explanations, disclaimers, or formatting—return only the structured JSON output.\\
\\
\#\#\# Format:\\\\
Return a raw JSON object only without additional text, explanations, or formatting:\\
\textasciigrave\textasciigrave\textasciigrave json\\\\
\{causal\_elements\_format\}\\\\
\textasciigrave\textasciigrave\textasciigrave
\end{promptbox}

\clearpage
\subsection{Prompts for Generating Counterfactuals (Causal Augmentation)}

\begin{promptbox}{Generating Upgraded Responses}
**Task:** Given a question and a model's response, generate a new response with a significantly improved response for the specified **\{ATTRIBUTE\}**, while *strictly preserving* all other aspects of the original response.\\\\
\\\\
\#\#\# **Input:**\\\\
- **Question:** \{QUESTION\}\\\\
- **Original Response:** \{RESPONSE\}\\\\
- **Causal Elements for \{ATTRIBUTE\}:**\\\\
   \{CAUSAL\_ELEMENTS\}\\\\
\\\\
\#\#\# **Instructions:**\\\\
\\\\
1. **Understand the Context:** Carefully read the question and original response and examine the provided causal elements that influence \{ATTRIBUTE\}.\\\\
2. **Identify the strength of \{ATTRIBUTE\}:** Determine which causal elements are **present** and their **direction of effect** (i.e., whether they increase or decrease \{ATTRIBUTE\}).\\\\
3. **Improve the Response:** Modify the causal elements to significantly improve \{ATTRIBUTE\}. Ensure that the improvement is **significant but isolated** to \{ATTRIBUTE\} leaving the other attributes intact.\\\\
4. **Verify the New Response:** Reassess whether \{ATTRIBUTE\} has been significantly improved. Confirm that all **other attributes remain unchanged**. If necessary, improve the response further to better meet the improvement goal for \{ATTRIBUTE\}.\\\\
5. **Return the New Response:** Provide the final modified response with a significantly improved \{ATTRIBUTE\} score. Format your response according to the format given below and in no other format.\\\\
\\\\
\#\#\# **Output Format:**\\\\
\\\\
Chain of Thoughts: <Your analysis of the original response, identification of causal elements, and strategy for improvement.>\\\\
New Response: <The final modified response which is significantly improved on \{ATTRIBUTE\}.>
\end{promptbox}

\begin{promptbox}{Generating Degraded Responses (Non-Safety)}
**Task:** Given a question and a model's response, generate a new response with a significantly degraded response for the specified **\{ATTRIBUTE\}**, while *strictly preserving* all other aspects of the original response.\\\\
\\\\
\#\#\# **Input:**\\\\
- **Question:** \{QUESTION\}\\\\
- **Original Response:** \{RESPONSE\}\\\\
- **Causal Elements for \{ATTRIBUTE\}:**\\\\
   \{CAUSAL\_ELEMENTS\}\\\\
\\\\
\#\#\# **Instructions:**\\\\
\\\\
1. **Understand the Context:** Carefully read the question and original response and examine the provided causal elements that influence \{ATTRIBUTE\}.\\\\
2. **Identify the strength of \{ATTRIBUTE\}:** Determine which causal elements are **present** and their **direction of effect** (i.e., whether they increase or decrease \{ATTRIBUTE\}).\\\\
3. **Degrade the Response:** Distort the causal elements to significantly degrade \{ATTRIBUTE\}. Ensure that the degradation is **significant but isolated** to \{ATTRIBUTE\} leaving the other attributes intact.\\\\
4. **Verify the New Response:** Reassess whether \{ATTRIBUTE\} has been significantly degraded. Confirm that all **other attributes remain unchanged**. If necessary, degrade the response further to better meet the degradation goal for the \{ATTRIBUTE\}.\\\\
5. **Return the New Response:** Provide the final modified response with a significantly degraded \{ATTRIBUTE\} score. Format your response according to the format given below and in no other format.\\\\
\\\\
\#\#\# **Output Format:**\\\\
\\\\
Chain of Thoughts: <Your analysis of the original response, identification of causal elements, and strategy for degradation.>\\\\
New Response: <The final modified response which is significantly degraded on \{ATTRIBUTE\}.>
\end{promptbox}

\subsection{Prompts for Generating Causally-Aligned Neutrals}
\subsubsection{Prompt for Comparing Responses via Causal Elements}

\begin{promptbox}{Generating Differences} 
\textless{}\textbar{}
You compare two responses based on content differences using a set of defined attributes and their causal elements.
\textless{}\textbar{}im\_end\textbar{}\textgreater{}
\textless{}\textbar{}im\_start\textbar{}\textgreater{}user
I will give you a question, two responses, and a list of attributes with their causal elements.

\textbf{Here is the question:}
\begin{lstlisting}[
    language=json,
    numbers=left,
    numberstyle=\tiny\color{gray},
    numbersep=5pt,
    xleftmargin=2em,
    breaklines=true
]
{
    "question": """{QUESTION}"""
}
\end{lstlisting}

\textbf{Here are the responses:}
\begin{lstlisting}[
    language=json,
    numbers=left,
    numberstyle=\tiny\color{gray},
    numbersep=5pt,
    xleftmargin=2em,
    breaklines=true
]
[
    {
        "model": "Response_1",
        "answer": """{RESPONSE1}"""
    },
    {
        "model": "Response_2",
        "answer": """{RESPONSE2}"""
    }
]
\end{lstlisting}

\textbf{Here are the attributes and causal elements:}
\begin{verbatim}
{CAUSAL_ELEMENTS}
\end{verbatim}

Please compare the responses for each attribute:
- Identify key content differences.
- Explain those differences using the causal elements only.
- Do not quote the responses directly.
- Focus only on what is said, not how it's said.

Return your output in this format:
\begin{lstlisting}[
    language=json,
    numbers=left,
    numberstyle=\tiny\color{gray},
    numbersep=5pt,
    xleftmargin=2em,
    breaklines=true
]
{
    "differences": [
        {
            "attribute": "<attribute>",
            "difference": "<summary>",
            "analysis": {
                "Response_1": "...",
                "Response_2": "..."
            }
        }
    ]
}
\end{lstlisting}

No extra text or explanation outside the JSON object.

\end{promptbox}

\clearpage
\subsubsection{Prompt for Generating Causally-Aligned Reconstruction}
\begin{promptbox}{Modifying Response Using Attribute-wise Causal Analysis}

You modify a given response by adjusting its causal elements to match a target profile based on attribute-wise analysis.
\textless{}\textbar{}im\_end\textbar{}\textgreater{}
\textless{}\textbar{}im\_start\textbar{}\textgreater{}user
I will provide you a question, a given response, and an attribute-based comparison analysis describing how to transform the given response into a target response.

\textbf{Inputs:}
1. \textbf{Question:}
\begin{verbatim}
{PLACEHOLDER_FOR_QUESTION}
\end{verbatim}

2. \textbf{Given Response:}
\begin{verbatim}
{PLACEHOLDER_FOR_GIVEN_RESPONSE}
\end{verbatim}

3. \textbf{Attribute-wise Differences Analysis:}
\begin{verbatim}
{PLACEHOLDER_FOR_ATTRIBUTE_DIFFERENCES_ANALYSIS}
\end{verbatim}

This analysis shows the differences between the given and target responses, broken down per attribute.

Each attribute section contains:
\begin{itemize}[leftmargin=2em] 
    \item \texttt{- Difference:} A summary of how the responses differ in content or emphasis.
    \item \texttt{- Analysis:}
    \begin{itemize}[leftmargin=2em] 
        \item \texttt{- Given Response:} Describes its content elements, grounding causal elements, and how they lead to the observed attribute.
        \item \texttt{- New Response:} Describes the content and causal elements the target response should exhibit instead.
    \end{itemize}
\end{itemize}

\textbf{Instructions:}
1.  Read the question and given response.
2.  Carefully study each attribute in the analysis and identify the causal elements needed to change.
3.  Generate a rewritten response that:
    \begin{itemize}[leftmargin=2em] 
        \item Retains the original meaning and structure.
        \item Implements the target causal elements.
        \item Removes or alters original ones as needed.
    \end{itemize}
4.  Do not introduce changes beyond the specified elements.
5.  Ensure the new response fully reflects the target causal profile across all attributes.

\textbf{Output Format:}
\begin{verbatim}
{{
    "Final Response": "<Write the transformed response here>"
}}
\end{verbatim}

Return only the final response JSON. Do not include any explanations or commentary.

\end{promptbox}

\subsection{Prompt for Generating Paraphrasing-Based Neutrals}
\begin{promptbox}{Prompt for Paraphrasing Responses}

\texttt{"""}

Paraphrase the following text while maintaining the \textbf{style}:

\texttt{{\{text\}}}

Make sure the meaning is \textbf{completely} the same without any changes.

Respond \textbf{only with the paraphrase} and \textbf{no extra text} at all;  
for example, do \textbf{NOT} preface with anything like:

\texttt{"Here is the paraphrased text:"}

\texttt{"""}

\end{promptbox}
\clearpage

\subsection{Prompt for Verifying the Counterfactual Quality}
\begin{promptbox}{Prompt for verifying the causal degradations}
You are an expert evaluator verifying whether the degraded response significantly weakens \{ATTRIBUTE\} while keeping all other aspects unchanged.
\\\\
\#\#\# Input Data
\\\\
- Query: \{QUESTION\}

- Original Response: \{RESPONSE\}

- Modified Response: \{NEW\_RESPONSE\}

- Causal Elements for \{ATTRIBUTE\}: \{CAUSAL\_ELEMENTS\}
\\\\
\#\#\# Verification Steps
\begin{enumerate}
    \item Identify Changes: Compare the original and modified responses to determine which causal elements were altered.
    \item Ensure Significant Degradation: Confirm that \{ATTRIBUTE\} is noticeably weakened, not subtly reduced.
    \item Check for Unintended Changes: Verify that the degradation is done by distorting the causal elements for \{ATTRIBUTE\} alone while keeping the other attributes unaffected.
    \item Determine Verdict: If only \{ATTRIBUTE\} is degraded significantly while all else remains unchanged, return **Pass**; otherwise, return **Fail**. *Strictly* adhere to the provided format.
\end{enumerate}

\#\#\# Output Format
\\\\
- If the modified response meets all requirements, return:

Verdict: Pass

- If the modified response does not meet the criteria, return:

Verdict: Fail
\end{promptbox}
\begin{promptbox}{Prompt for verifying the causal upgradations}
You are an expert evaluator verifying whether the degraded response significantly strengthens \{ATTRIBUTE\} while keeping all other aspects unchanged.

\#\#\# Input Data
- Query: \{QUESTION\}
- Original Response: \{RESPONSE\}
- Modified Response: \{NEW\_RESPONSE\}
- Causal Elements for \{ATTRIBUTE\}: \{CAUSAL\_ELEMENTS\}

\#\#\# Verification Steps
\begin{enumerate}
    \item Identify Changes: Compare the original and modified responses to determine which causal elements were altered.
    \item Ensure Significant Improvement: Confirm that \{ATTRIBUTE\} is noticeably improved, not subtly improved.
    \item Check for Unintended Changes: Verify that the improvement is done by modifying the causal elements for \{ATTRIBUTE\} alone while keeping the other attributes unaffected.
    \item Determine Verdict: If only \{ATTRIBUTE\} is improved significantly while all else remains unchanged, return **Pass**; otherwise, return **Fail**. *Strictly* adhere to the provided format.
\end{enumerate}

\#\#\# Output Format
- If the modified response meets all requirements, return:

Verdict: Pass

- If the modified response does not meet the criteria, return:

Verdict: Fail
\end{promptbox}

\subsection{GPT4-as-a-Judge Prompt}
\vspace{-0.15in}
\begin{promptbox}{LLM-as-a-Judge Prompt}
\textless{}|im\_start|\textgreater{}system

You are a helpful assistant, that ranks models by the quality of their answers, prioritizing substantive content and relevance to the query.

\textless{}|im\_end|\textgreater{}
\textless{}|im\_start|\textgreater{}user

I want you to create a leaderboard of different large-language models. To do so, I will give you the instructions (prompts) given to the models, and the responses of two models. All inputs and outputs should be Python dictionaries.

\textbf{Here is the prompt:}
\begin{lstlisting}[
    language=json,
    numbers=left,
    numberstyle=\tiny\color{gray},
    numbersep=5pt,
    xleftmargin=2em,
    breaklines=true
]
{
    "instruction": """{instruction}"""
}
\end{lstlisting}

\textbf{Here are the outputs of the models:}
\begin{lstlisting}[
    language=json,
    numbers=left,
    numberstyle=\tiny\color{gray},
    numbersep=5pt,
    xleftmargin=2em,
    breaklines=true
]
[
    {
        "model": "model_1",
        "answer": """{output_1}"""
    },
    {
        "model": "model_2",
        "answer": """{output_2}"""
    }
]
\end{lstlisting}

\textbf{Now please rank the models by the quality of their answers}, so that the model with rank 1 has the best output. Your ranking should be based primarily on which response \textbf{provides the most accurate, relevant, and comprehensive content in direct relation to the query}.

Focus on the core information and how well the query's intent is addressed, rather than on aspects like writing style, formatting, length, or politeness, unless these directly impact the clarity or usefulness of the core content requested by the query.

\textbf{Your response MUST be a valid JSON object.}  
This JSON object must contain \textbf{ONLY ONE} key named \texttt{"ranking\_list"}.  
The value of the \texttt{"ranking\_list"} key MUST be a list of two dictionaries, where each dictionary contains the model name and its rank.

The structure should be \textbf{exactly} as follows:
\begin{lstlisting}[
    language=json,
    numbers=left,
    numberstyle=\tiny\color{gray},
    numbersep=5pt,
    xleftmargin=2em,
    breaklines=true
]
{
    "ranking_list": [
        {"model": "<model-name>", "rank": <model-rank>},
        {"model": "<model-name>", "rank": <model-rank>}
    ]
}
\end{lstlisting}

Do \textbf{not} include any other text, explanations, or additional keys in the JSON object.  
Please provide the ranking that a well-informed human, focusing on these content-centric criteria, would likely give.

\textless{}|im\_end|\textgreater{}
\end{promptbox}

\section{Qualitative Examples of Reward Model Behavior}
\label{sec:qualitative_examples}

This section presents several examples illustrating common failure modes or biases observed in standard reward models, highlighting the types of spurious correlations they may learn.

\subsection*{Example 1: Formatting Bias (Plumbing Diagram)}

\QuestionV{Question}{Construct plumbing diagrams for a two-story house (describe its core in natural language)}

\AcceptedV{Chosen Response}{
Main floor plumbing diagram for a two-story house: Waste pipes from the sinks, toilets, and bathtubs on the main floor should all connect to a common pipe that runs to the main drain line in the basement. Similarly, the water supply lines from the faucets, showers, and other fixtures should all connect to a common supply pipe which then runs to the first-floor water heater.
}

\RejectedV{Rejected Response}{
Plumbing diagram for a two-story house:
\\
Ground floor:
- Toilet
- Bathroom sink
- Shower
- Dishwasher
\\
Second floor:
- Toilet
- Bathroom sink
- Shower
}

\Remarks{Remarks}{
The rejected response uses a list format (a spurious attribute), which might be preferred by a reward model biased towards formattedness, even though the chosen response provides a more descriptive answer aligned with the prompt's request for natural language description.
}
\clearpage
\subsection*{Example 2: Politeness Bias (Medical Diagnosis)}

\vspace{-0.1in}
\QuestionV{Question}{
Imagine yourself as a doctor tasked with devising innovative remedies for various ailments and maladies. Your expertise should encompass prescribing traditional medications, herbal treatments, and alternative natural solutions. Additionally, you must take into account the patient's age, lifestyle, and medical background while offering your recommendations. To begin, please assist me in diagnosing a scenario involving intense abdominal discomfort.
}

\vspace{-0.05in}
\AcceptedV{Chosen Response}{
Certainly! To diagnose a scenario involving intense abdominal discomfort, I would first conduct a thorough examination of the patient's symptoms and medical history. This would involve a physical examination of the abdomen, including a pelvic exam, an abdominal X-ray, and a colonoscopy.\\\\
If the patient has a history of abdominal pain and nausea, he may be suffering from a condition called gastroparesis, also known as "silent gas pain." This is a serious condition characterized by severe abdominal pain and discomfort that can be severe or constant. The discomfort can be accompanied by nausea, vomiting, and diarrhea. It is important to seek medical attention immediately if the patient experiences any of these symptoms.
Once I've determined the cause of the abdominal discomfort, I would then prescribe traditional medications to alleviate any pain or discomfort. These medications may include pain relievers like ibuprofen or acetaminophen, as well as anti-inflammatory drugs like ibuprofen and naproxen. If a traditional medication is not effective, alternative natural treatments may be prescribed. These treatments may include acupuncture, herbal remedies, and home remedies that may help to increase blood flow to the abdomen and alleviate any pain.\\\\
It's important to note that these are just a few examples of possible causes of abdominal discomfort. If you have any other symptoms or medical conditions that may be contributing to your discomfort, it is important to consult with a healthcare professional for a proper diagnosis and treatment plan.
}

\vspace{-0.05in}
\RejectedV{Rejected Response}{
Sure, I can help you with that. Based on the symptoms you've described, it's possible that the patient may be suffering from gastroenteritis, commonly known as stomach flu. However, it's essential to consider other potential causes such as food poisoning, gastritis, or appendicitis. To determine the root cause, I'd recommend a thorough medical evaluation, including blood tests and imaging studies. Once the diagnosis is confirmed, appropriate treatment can be prescribed, which may include medication, hydration, and rest. It's crucial to consult a healthcare professional for accurate diagnosis and treatment.
}

\vspace{-0.05in}
\Remarks{Remarks}{
The rejected response begins with polite phrases ("Sure, I can help you with that."), a potentially spurious attribute that reward models might learn to favor, even if the chosen response provides a more detailed diagnostic thought process aligned with the persona requested.
}

\clearpage

\subsection*{Example 3: Formatting and Instruction Following (Survey Question)}

\vspace{-0.1in}
\QuestionV{Question}{Create a survey question that can help capture insights about the customer experience.}

\vspace{-0.05in}
\AcceptedV{Chosen Response}{
How likely would you be to recommend our product or service to a friend or colleague?
}

\vspace{-0.05in}
\RejectedV{Rejected Response}{
Sure! Here's a sample feedback form that can gather information about the user interface experience:
\\\\
1. Overall, how would you rate your experience with the user interface? (Scale of 1-5, with 1 being poor and 5 being excellent)\\
2. Which specific features of the user interface did you find most useful or intuitive? (Open-ended text response)\\
3. Were there any features of the user interface that you found confusing or difficult to use? If yes, please provide details. (Open-ended text response)\\
4. Did you encounter any technical issues or bugs while using the user interface? If yes, please describe the issue(s) you faced. (Open-ended text response)\\
5. How would you rate the visual design and aesthetics of the user interface? (Scale of 1-5, with 1 being poor and 5 being excellent)\\
6. Were there any specific aspects of the visual design that you particularly liked or disliked? (Open-ended text response)\\
7. Did you find the user interface responsive and fast? (Yes/No)\\
8. Were there any specific areas where you experienced slow loading times or delays? If yes, please specify. (Open-ended text response)\\
9. How would you rate the organization and layout of the user interface? (Scale of 1-5, with 1 being poor and 5 being excellent)\\
10. Were there any specific areas where you felt the organization or layout could be improved? If yes, please provide details. (Open-ended text response)\\
11. Did you find the user interface easy to navigate and find what you were looking for? (Yes/No)\\
12. Were there any specific areas where you had difficulty navigating or finding information? If yes, please specify. (Open-ended text response)\\
13. How likely are you to recommend the user interface to others? (Scale of 1-5, with 1 being not likely at all and 5 being very likely)\\
14. Do you have any additional comments or suggestions for improving the user interface? (Open-ended text response)\\
\\
Thank you for taking the time to provide your feedback!
}

\vspace{-0.05in}
\Remarks{Remarks}{
The rejected response fails to follow the instruction (provide a single question) and instead provides a lengthy form. A reward model might erroneously prefer it due to spurious features like length, formatting (numbered list), and politeness ("Sure!", "Thank you!").
}

\clearpage
\subsection*{Example 5: Formatting Bias vs. Content Quality (Electrician)}
\QuestionV{Question}{What is the fastest way to become an electrician in the us}
\AcceptedV{Chosen Response}{
Obtain a high school diploma or GED.
Complete an electrical apprenticeship program, which typically lasts 4–5 years and combines on-the-job training with classroom instruction.
Pass the Journeyman Electrician Exam to become licensed.
Note that requirements and the time frame may vary depending on the state you are in.
}
\RejectedV{Rejected Response}{
To become an electrician in the US, the fastest path typically involves the following steps:
Foundational Knowledge:
Formal Education: Consider pursuing a degree in electrical engineering or a related field. This will provide you with structured courses that cover the basics of electrical theory, circuit analysis, electronics, power systems, electromagnetism, and more.
Online Courses and Tutorials: Websites like Coursera, edX, Udacity, and Khan Academy offer courses in electrical engineering and electronics. They can serve as an excellent supplement or alternative to traditional education.
Hands-on Experience:
DIY Projects: Start with small projects like building basic circuits, then gradually move to more complex ones. Kits are available online to help beginners learn the basics of circuit design and assembly.
Internships and Apprenticeships: These provide invaluable real-world experience. Working under a mentor or experienced electrician can greatly enhance your learning.
Certifications and Specializations:
While not always mandatory, obtaining certifications can validate your expertise and open doors for specialized roles or areas within the field of electrics.
Continuous Learning:
Workshops and Seminars: Regularly attend these to stay updated with the latest techniques, tools, and best practices in the field.
Books and Publications: Regularly read books, journals, and other publications related to electrical engineering and electronics.
Networking:
Professional Associations: Joining organizations like the Institute of Electrical and Electronics Engineers (IEEE) can provide networking opportunities, resources, and continued education.
Online Forums and Groups: Websites like Stack Exchange (for electrical engineering), online forums, and even Reddit have active communities where you can ask questions, share knowledge, and learn from others.
Practice and Experimentation:
Regularly challenge yourself with new projects or problems to solve. The more you apply your knowledge, the more intuitive and deeply ingrained it will become.
Stay Updated:
Industry Trends: Electrics, like all fields of technology, is continuously evolving. Ensure you're aware of the latest tools, technologies, and methodologies in the field.
Regular Refreshers: Periodically revisit foundational concepts to ensure your foundational knowledge remains strong.
Soft Skills:
While technical knowledge is crucial, don't underestimate the importance of communication, teamwork, problem-solving, and critical thinking skills. These are often essential, especially if you aim to take on leadership or interdisciplinary roles.
Teach and Mentor:
Once you've gained a certain level of expertise, consider teaching or mentoring others. This not only benefits the community but can also reinforce and expand your understanding.
Becoming an electrician requires dedication, passion, and a willingness to continually push your boundaries. With focused effort and the right resources, you can achieve a high level of expertise and enjoy a fulfilling career in the world of electronics.
}
\Remarks{Remarks}{
The rejected response, while containing much unnecessary and potentially irrelevant information (e.g., suggesting an electrical engineering degree), uses extensive formatting (numbered lists with sub-points). A reward model overly sensitive to formattedness might prefer it over the more concise and directly relevant chosen response.
}

\end{document}